\documentclass[10pt,twocolumn,letterpaper]{article}

\usepackage[pagenumbers]{templates/wacv/wacv}
\usepackage{censor}
\usepackage[utf8]{inputenc}
\usepackage[T1]{fontenc}
\usepackage{url}            
\usepackage{amsfonts}       
\usepackage{nicefrac}       
\usepackage{microtype}     
\usepackage{amsmath}
\usepackage{soul}
\usepackage{amssymb}
\usepackage{enumitem}
\usepackage{xfrac}
\usepackage[export]{adjustbox}
\usepackage{svg}
\usepackage{style}
\usepackage{pifont}
\usepackage{multirow}
\usepackage{mwe}
\usepackage{contour}
\usepackage{transparent}
\usepackage{arydshln}
\usepackage{tikz}
\usepackage{tabularx}
\usepackage{colortbl}
\usepackage{pdflscape}
\usepackage{fontawesome5} 
\usepackage{pgfplots}
\pgfplotsset{compat=1.18}
\usetikzlibrary{shapes,patterns}
\usetikzlibrary{arrows.meta}
\usetikzlibrary{fadings, shadows}
\usetikzlibrary{arrows, positioning, fit, backgrounds, calc}
\usetikzlibrary{matrix, fit}
\usepackage{subcaption}

\definecolor{cmarkColor}{RGB}{108, 142, 191}
\definecolor{xmarkColor}{RGB}{184, 84, 80}
\renewcommand{\cmark}{{\color{cmarkColor}\ding{51}}}%
\renewcommand{\xmark}{{\color{xmarkColor}\ding{55}}}%

\renewcommand{\paragraph}[1]{\smallskip\noindent{\bf#1}}
\NewDocumentCommand{\rot}{O{45} O{1em} m}{\makebox[#2][l]{\rotatebox{#1}{#3}}}%

\definecolor{wacvblue}{rgb}{0.21,0.49,0.74}
\usepackage[pagebackref,breaklinks,colorlinks,allcolors=wacvblue]{hyperref}
\usepackage[capitalize]{cleveref}

\title{Remote Sensing Change Detection via Weak Temporal Supervision}

\author{%
    Xavier Bou\textsuperscript{1,4 *} \quad
    Elliot Vincent\textsuperscript{2 *} \quad
    Gabriele Facciolo\textsuperscript{1,5} \\
    Rafael Grompone von Gioi\textsuperscript{1} \quad
    Jean-Michel Morel\textsuperscript{3} \quad
    Thibaud Ehret\textsuperscript{4} \\
    \textsuperscript{1}Université Paris-Saclay, CNRS, ENS Paris-Saclay, Centre Borelli, France \\
    \textsuperscript{2}Université Gustave Eiffel, Géodata Paris, IGN, LASTIG, France \\
    \textsuperscript{3}Lingnan University, Hong Kong\; \textsuperscript{4}AMIAD, Pole Recherche, France \; \textsuperscript{5}Institut Universitaire de France \\
    {\small\url{https://xavibou.github.io/CDviaWTS/}}
}

\begin{document}
\twocolumn[{%
\renewcommand\twocolumn[1][]{#1}%
\maketitle%
\begin{center}
    \begin{center}

\captionsetup{type=figure}
\centering  
\resizebox{0.9\linewidth}{!}{
\begin{tabular}{wl{0.33\linewidth}wl{0.33\linewidth}l}
         Ground truth & Ours & FSC-180k~\cite{benidir2025change} \\
         \multicolumn{3}{c}{\includegraphics[width=\linewidth,trim={0 4.75cm 0 0}, clip]{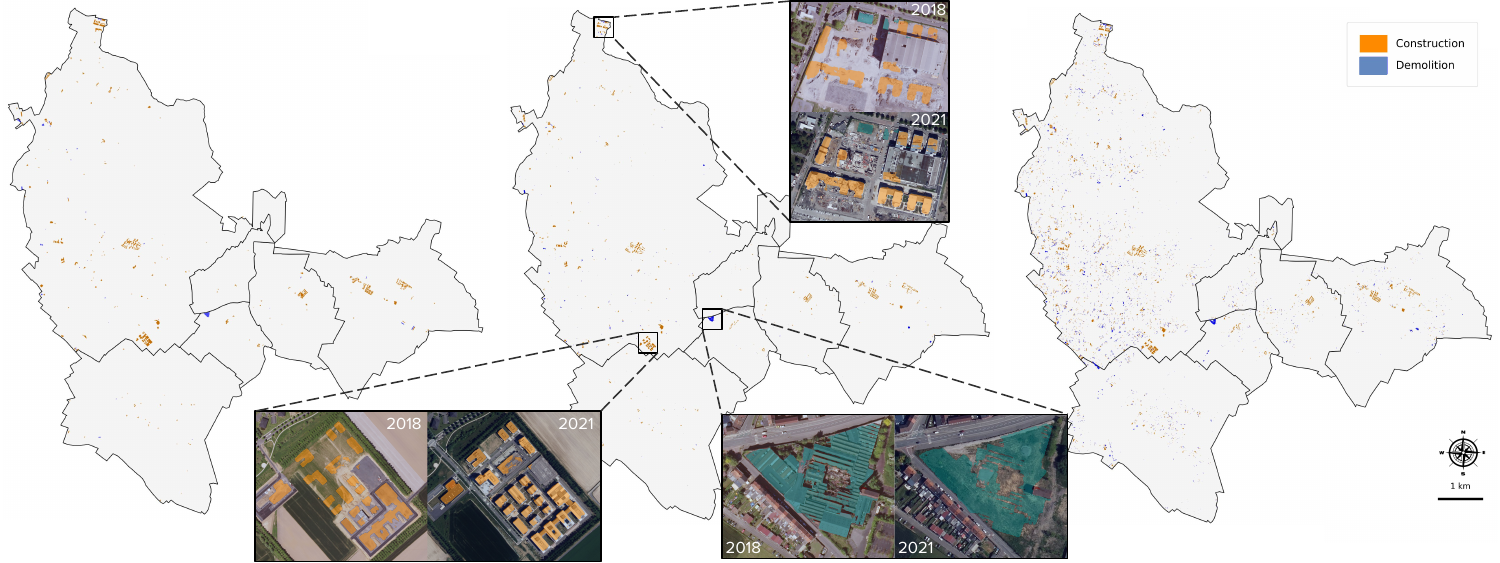}}
    \end{tabular}
}
\vspace{-.5em}
\captionof{figure}{
\textbf{Large-scale building change detection comparison on BD ORTHO~\cite{ign2025bdortho}.} 
    Availability of annotated datasets has always been a challenge for semantic change detection. 
    We propose a weak temporal supervision strategy leveraging additional temporal observations 
    of existing single-date annotated data. 
    Left: reference building changes derived from IGN's OCS GE data~\cite{ign2025ocsge}. 
    Middle: Dual UNet trained with our methodology. 
    Right: Dual UNet trained on FSC-180k~\cite{benidir2025change}.
    Results for additional methods are shown in the Supplementary Material.}
\label{fig:teaser}
\end{center}

\end{center}%
}]
\vspace{-6mm}

\nnfootnote{\textsuperscript{*}~These authors contributed equally}

\begin{abstract}
  Semantic change detection in remote sensing aims to identify land cover changes between bi-temporal image pairs. Progress in this area has been limited by the scarcity of annotated datasets, as pixel-level annotation is costly and time-consuming. To address this, recent methods leverage synthetic data or generate artificial change pairs, but transferring change detection models across datasets and acquisition conditions remains challenging. In this work, we introduce a weak temporal supervision strategy that leverages additional temporal observations of existing single-temporal datasets, without requiring any new annotations. Specifically, we extend single-date remote sensing datasets with new observations acquired at different times and train a change detection model by assuming that real bi-temporal pairs mostly contain no change, while pairing images from different locations to generate change examples. To handle the inherent noise in these weak labels, we employ an object-aware change map generation and an iterative refinement process. We validate our approach on extended versions of the FLAIR and IAILD aerial datasets, showing competitive zero-shot and low-data regime performance. Lastly, we showcase results over large areas in France, highlighting the scalability potential of our method.
\end{abstract}
\section{Introduction}

Detection and monitoring of changes in remote sensing data remain a central topic within the Earth observation community over recent years~\cite{cheng2024change}. A wide range of applications emerge from remote sensing change detection, including natural disaster damage assessment~\cite{brunner2010change, chuvieco2020satellite, bou2024portraying}, urban development monitoring~\cite{hegazy2015monitoring, wen2016extraction}, or deforestation detection~\cite{khan2017forest, watanabe2018early}. In particular, the task of bi-temporal change detection consists in finding the semantic differences between a pair of co-registered satellite or aerial images acquired at different dates, given a set of land cover classes of interest. 

State-of-the-art methods for this task use deep neural networks~\cite{chen2024changemamba, ding2024joint, paranjape2025mamba, tan2025triples} trained on datasets of image pairs annotated with pixel-level change maps~\cite{ji2018fully, chen2020spatial, yang2021asymmetric, shen2021s2looking} or even semantic maps at both dates~\cite{daudt2018urban, tian2022large}. Nevertheless, gathering such data is both expensive and time consuming. As a result, change detection data is often spatially or temporally clustered, which can limit the transfer of models to new locations and acquisition conditions. To address the scarcity of labeled data, several works have attempted to train change detection models in an unsupervised~\cite{celik2009unsupervised, noh2022unsupervised} or weakly supervised fashion~\cite{andermatt2020weakly, kalita2021land, wang2023cs, zheng2025csnet}, learning from much cheaper image-level annotations. However, these methods still lag behind fully supervised approaches, particularly in the spatial accuracy of predicted changes.

To benefit from quality semantic annotations at a larger scale, several works attempt to leverage single-temporal annotated datasets in order to train change detection model. This is the case of synthetic change datasets generated from such existing data~\cite{benidir2025change}, 
but also of learning frameworks considering non-overlapping images as training pairs~\cite{zheng2021change, seo2023self}, pairing images from different locations and thus generating \textit{fake change pairs}. However, such attempts do not take advantage of the growing availability of remote sensing data. Satellite and aerial image acquisition campaigns are indeed often programmed to occur regularly by national GIS institutes~\cite{ign2025bdortho, usgs2022how} or to last a long period of time by spatial agencies~\cite{esa2017sentinel, nasa2024landsat}. 

In this paper, we propose to expand single-date semantic segmentation datasets into bi-temporal collections, leveraging easily accessible imagery. We introduce a new weakly supervised paradigm, where only single-temporal semantic maps are available to train a change detection network on bi-temporal pairs. 
Introducing this temporal variability exposes the model to naturally occurring variations that do not constitute semantically meaningful changes, with the aim of improving robustness to temporal and acquisition variability.
Because this weak temporal supervision is by nature noisy, we propose three key training ideas. First, we balance real bi-temporal pairs and fake non-overlapping pairs during training. Secondly, inspired by~\cite{zheng2021change}, we supervised fake pairs with an sIoU-based~\cite{rottmann2020prediction, chan2021segment} change map. Third, we iteratively clean the extended dataset, filtering out real pairs exhibiting land-cover changes, so that such pairs can be considered as unchanged during training. A general diagram of our method is provided in the Supplementary Material.

We validate our methodology by expanding two existing datasets, FLAIR~\cite{garioud2023flair} and IAILD~\cite{maggiori2017dataset}. Through several experiments, we demonstrate that one can use additional temporal acquisitions in addition to semantic segmentation datasets to improve change detection with zero labeling cost. Our contributions are summarized as follows:
\begin{itemize}
    \item We extend FLAIR and IAILD datasets from single-date to bi-temporal and release them for anyone to use. We additionally curate and release a test set for in-domain evaluation of building change detection methods trained on our extension of FLAIR.
    \item We develop an approach to leverage the additional, non-annotated images for training a change detection network.
    \item Through extensive experiments on several datasets, we show that our method improves the performance of change detection models. Thus, we demonstrate competitive zero-shot and low-data regime performance, as well as compelling large-scale qualitative results, shown in Fig.~\ref{fig:teaser}.
\end{itemize}
\section{Related Works}
\label{sec:rw}
In this paper, we tackle detection of land cover changes in bi-temporal remote sensing image pairs, referring to this task as semantic change detection (SCD).

\paragraph{Semantic change detection.} Over the past few years, SCD in remote sensing images has gained significant interest, leading to a large number of publications on the topic and several field surveys~\cite{asokan2019change, khelifi2020deep, shafique2022deep, cheng2024change}. Most state-of-the-art methods rely on deep learning in order to train 3-branch neural networks. First introduced by Daudt \textit{et al.}~\cite{daudt2019multitask}, such architectures output, for a given bi-temporal image pair, a triplet consisting of two semantic maps and a binary change map. Multiple variants~\cite{yang2021asymmetric, ding2022bi, zhao2022spatially, xia2022deep, zheng2022changemask, li2023lightweight, jiang2023ttnet, cui2023mtscd, ding2024joint, liu2024tbscd, long2025bgsnet} improve different aspects of the procedure, including the multitask objective~\cite{zheng2022changemask, cui2023mtscd}, the fusion mechanisms~\cite{xia2022deep, jiang2023ttnet}, the consistency between the three outputs~\cite{ding2024joint}, the quality of the predicted change boundaries~\cite{liu2024tbscd, long2025bgsnet}, or the computational cost~\cite{li2023lightweight}. All these methods require pixel-level annotated image pairs in order to train a model in a fully supervised manner. Because gathering such labeled data is costly and time-consuming, we instead propose a weakly supervised framework based on single-temporal annotations, leveraging new temporal images without additional annotation cost.

\paragraph{Weakly supervised change detection.} Weakly supervised learning encompasses any training algorithm that enables performing different or more complex tasks than typically possible with the available data. However, in the context of SCD, it specifically describes methods that infer pixel-level change predictions from weak image-level labels~\cite{andermatt2020weakly, kalita2021land, wang2023cs, zheng2025csnet}. For instance, Andermatt~\textit{et al.}~\cite{andermatt2020weakly} train a SCD model with image-level labels such as ``forests to agricultural surfaces'' for a given image pair. Between image- and pixel-level supervision, other weak signals such as bounding boxes or coarse masks~\cite{daudt2023weakly, wang2023sdcdnet, liu2025box2change}, noisy labels~\cite{cao2023full}, and low-resolution annotations~\cite{li2022outcome} have been used to train SCD networks. Toker~\textit{et al.} presented DynamicEarthNet~\cite{toker2022dynamicearthnet}, a dataset of daily satellite image time series for which only the first image of each month is annotated, with a focus on semi- rather than weak supervision. Beyond this work, and to the best of our knowledge, weak settings in which semantic information is completely missing for one image of the pair are often overlooked in the literature, despite corresponding to cases that commonly occur in practice due to the recurrent acquisition of imagery by space and GIS agencies. A unique work that deserves mention is Segment Any Change~\cite{NEURIPS2024_94154162}, which leverages foundational model SAM~\cite{sam} to detect bi-temporal, class-agnostic changes in remote sensing images in a zero-shot manner.

\begin{figure*}[t]
    \centering
    \setlength{\tabcolsep}{0.7pt}
    \scriptsize
    \resizebox{.9\linewidth}{!}{
    \begin{tabular}{@{\hspace{-6mm}}ccccccccc}
    \multicolumn{6}{@{\hspace{-6mm}}l}{\includegraphics[height=7pt]{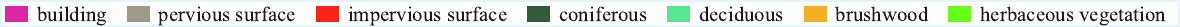}} &
    \multicolumn{3}{c}{\includegraphics[height=7pt]{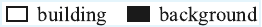}} \\

    \sampletriplet{b-FLAIR}{014567} &
    \sampletriplet{b-FLAIR-spot}{014567} 
    \sampletriplet{b-IAILD}{austin10_46} \\

    \sampletriplet{b-FLAIR}{017891} &
    \sampletriplet{b-FLAIR-spot}{017891} 
    \sampletriplet{b-IAILD}{chicago36_0} \\

    \multicolumn{3}{@{\hspace{-7mm}}c}{(a) b-FLAIR} &
    \multicolumn{3}{@{\hspace{-15mm}}c}{(b) b-FLAIR-spot} &
    \multicolumn{3}{@{\hspace{-5mm}}c}{(c) b-IAILD} \\
    \end{tabular}
    }
    \vspace{-0.5em}
    \caption{\textbf{Visual examples.} For each dataset, we show example triplets ($S_{t}$, $I_{t}$, $I_{t'}$) from left to right: annotation mask, original image, and additional acquisition. Top row: significant land cover changes; middle and bottom: irrelevant changes (shadows, seasonal variations).}
    \vspace{-2em}
    \label{fig:example_triplets}
\end{figure*}

\paragraph{Leveraging single-temporal annotated data.} Due to the high cost of pixel-level change annotations, several works have explored leveraging single-temporal semantic segmentation datasets to enable change detection without requiring ground truth change labels. A common approach is post-classification change detection, often referred to as “the most obvious method of change detection”~\cite{singh1989review}. It compares the outputs of a single-temporal segmentation model on each image of a bi-temporal pair. In this case, multi-temporal acquisitions may serve as data augmentation at training time~\cite{ayala2022multi}. However, this method suffers from prediction error accumulation and lacks temporal modeling~\cite{tewkesbury2015critical}. Other methods construct artificial change pairs by randomly pairing images from different locations and computing change maps based on label differences, \eg STAR~\cite{zheng2021change} or \cite{seo2023self}. Another line of work consists in creating synthetic SCD datasets based on a single-temporal land cover real-world dataset, using techniques such as GANs~\cite{zheng2023scalable, seo2023self}, or diffusion~\cite{tang2024changeanywhere, zheng2024changen2, benidir2025change}. Building on the observation that annotation is significantly more expensive than data acquisition itself, we instead propose to extend existing single-temporal datasets with new, aligned temporal acquisitions, using their existing annotations as weak supervision for bi-temporal change detection.
\section{Methodology}
\label{sec_method}
We focus on learning to detect changes from a single-temporal satellite or aerial image dataset annotated for the semantic segmentation task. Based on the observation that accessing new acquisitions is less costly than obtaining change labels, our pipeline starts by extending such datasets temporally (Sec.~\ref{sec:data}). Then we leverage these new bi-temporal pairs, along with the single-temporal annotations in order to train a semantic change detection model (Sec.~\ref{sec:training}). We detail our implementation in Sec.~\ref{sec:implementation}.

\paragraph{Problem Definition.} Let $\mathcal{D}$ be a single-temporal dataset of $N$ aerial or satellite annotated images $I_{t_i}^i$ in $\mathbb{R}^{C\times H\times W}$ acquired at time $t_i$, for $i \in \{1, \ldots, N\}$. $C$, $H$, and $W$ respectively refer to the number of channels, height, and width of the images. Each $I_{t_i}^i$ is annotated with a semantic mask  $S_{t_i}^i$ in $\{1, ...,K\}^{H\times W}$, with $K$ the number of semantic classes. Now, let $f_\theta$ be a 
neural network with learnable parameters $\theta$, returning, for a given input pair $(I_{t}, I_{t'})$ of bi-temporal images, a change map of pixels whose semantic class changes between the acquisition timestamps $t$ and $t'$. Our goal is to train such a model $f_\theta$, using the dataset $\mathcal{D}$ extended with new, easily-available, non-annotated temporal acquisitions $I_{t'_i}$.
\subsection{Data} 
\label{sec:data}
To verify our methodology, we apply it to two existing aerial datasets, FLAIR~\cite{garioud2023flair} and the Inria
aerial image labeling dataset (IAILD)~\cite{maggiori2017dataset}, spanning areas in three countries (France, USA and Austria). Since we extend these datasets so they contain \textit{bi-temporal} pairs, we prepend their name with ``b-'' to distinguish them from the original datasets.

\paragraph{b-FLAIR:} FLAIR is based on aerial images from the French National Institute of Geographical and Forest Information (IGN)'s BD ORTHO~\cite{ign2025bdortho}: a database of aerial orthophotographies covering the French territory at 0.2 meters per pixel (m/px). All $512\times512$ FLAIR image patches are annotated with semantic maps that assign each pixel to one of 19 land cover classes. BD ORTHO is renewed on average every three years, making it possible to collect aerial images of the same locations at different dates. We thus extract, for every FLAIR image, an additional observation from the BD ORTHO database at around three years before or after the original FLAIR acquisition.
b-FLAIR is challenging for weakly supervised change detection with single-date annotations for two reasons. First, FLAIR areas were selected for land cover diversity with no particular temporal consideration, probably resulting in a low proportion of pairs with relevant semantic change. Second, approximately 30\% of pairs have acquisition dates differing by two or more months, introducing seasonal variations that, combined with varying acquisition times within the day, may introduce significant radiometric and shadow changes.

\paragraph{b-IAILD:} IAILD covers urban areas in the USA and Austria. We only extended IAILD's training set, as its test set is not publicly available. It consists of 180 patches of size $5000\times5000$ at a resolution of 0.3 m/px, annotated with binary building footprint masks. For each location, we gathered a patch from the most recent available orthoimage acquisition campaign released by the corresponding agency (USGS for the USA, and the respective GIS agencies for Austrian provinces). The added images have acquisition dates spanning 2022 to 2024, while IAILD's original images were acquired before 2017. As spatial resolution of new acquisitions vary from 0.15 to 0.6 m/px, we standardized b-IAILD to 0.6 m/px.

\paragraph{b-FLAIR-spot:} To verify our methodology beyond very high resolution aerial imagery, we also created a satellite variant of b-FLAIR. For each bi-temporal pair, we downloaded the corresponding SPOT-6/7 images from IGN's ORTHO-SAT database~\cite{ign2025orthosat} for the same acquisition year and location. The images have a spatial resolution of 1.5 m/px, which we resampled to 1.6 m/px to obtain $64\times64$ patches. The corresponding single-date semantic maps were also resampled to the same resolution via nearest neighbor interpolation. We refer to this satellite dataset as b-FLAIR-spot.

b-FLAIR, b-IAILD and b-FLAIR-spot thus contain triplets ($S_{t}$, $I_{t}$, $I_{t'}$). We show example of such triplets in Fig.~\ref{fig:example_triplets}. Further details on these datasets, and additional example images can be found in the Supplementary.

\paragraph{Evaluation Datasets.} We evaluate competing methods on 8 datasets. For the in-domain evaluation of methods trained on b-FLAIR and b-FLAIR-spot, we release FLAIR-BCD and FLAIR-BCD-spot (for building change detection), two evaluation sets of 1730 image pairs annotated with a binary building change mask. The images are in the same format as FLAIR images and were acquired, processed, and formatted following the procedure described in~\cite{garioud2022flair}. The images were extracted from 9 different French administrative departments and do not intersect the FLAIR dataset, which allows for a sound evaluation of methods trained on FLAIR or on our bi-temporal extensions of FLAIR. The pairs were annotated by photointerpretation experts and verified by a non-expert assessor. The pairs either show new building constructions, or no building change at all ($\thicksim$30\% of the pairs), and no pair exhibits building destruction. For the out-of-domain evaluation of all methods, we selected LEVIR-CD~\cite{chen2020spatial}, WHU-CD~\cite{ji2018fully}, and S2Looking~\cite{shen2021s2looking}, three datasets commonly used in building change detection benchmarks. Moreover, we conduct experiments on the Hi-UCD mini~\cite{hiucd} and Hi-UCD~\cite{tian2022large} datasets, evaluating the results of our approach in a multi-class SCD setup. Hyperparameter studies are performed using a subset of b-FLAIR for training, and FLAIR-BCD for evaluation. To provide an untouched evaluation, we additionally release FLAIR-BCD-heldout: 400 aerial pairs spatially disjoint from FLAIR, b-FLAIR and FLAIR-BCD. More details on the evaluation datasets can be found in the Supplementary.

\subsection{Training with single date labels}
\label{sec:training}
\input{tables/main_results}

Given a set of triplets $(S_{t_i}^i, I_{t_i}^i, I_{t_i'}^i)_{i=1,...,N}$, our goal is to train a bi-temporal change detection neural network $f_\theta$. Our methodology can be applied to any model predicting, for an input image pair ($I_{t}$, $I_{t'}$), a triplet ($\hat{S}_{t}$, $\hat{S}_{t'}$, $\hat{M}$) respectively consisting of a semantic mask for each frame and a binary change mask.
This corresponds to a widely accepted family of architectures. Indeed, they usually consist of a triple-branch network~\cite{daudt2019multitask, yang2021asymmetric, ding2022bi, ding2024joint, chang2024triple} with two semantic segmentation branches and a binary change branch. We now detail the three main components of our method: batch sampling, change map generation, and iterative refinement.

\paragraph{Balanced batch sampling.} Our extended datasets typically contain a small fraction of pairs exhibiting actual change events. Zheng \textit{et al.}~\cite{zheng2021change} compensate the lack of annotated change datasets in the literature by artificially pairing images from different locations, using as change labels a difference of their respective semantic masks. Building on this idea, we choose to mix in a single batch such \textit{fake} pairs with \textit{real} bi-temporal pairs. We thus introduce a parameter $p_{\text{real}}$ in $[0,1]$, corresponding to the proportion of real pairs in each batch.  Specifically, for a batch of size $B$ of pairs ($I_{t_i}^i, I_{t_i'}^i)_{i=1,...,B}$ sampled from $\mathcal{D}$, the method splits it into $N_{\text{real}}$ bi-temporal examples and $N_{\text{fake}}$ unaligned examples, following:
\begin{equation}
N_{\text{real}} = \lfloor B \times p_{\text{real}} \rfloor~, \quad N_{\text{fake}} = B - N_{\text{real}}~.
\end{equation} 

We apply a random permutation to the $N_{\text{fake}}$ samples, such that the image $I_{t_i}^i$ is paired with  $I_{t_j'}^j$ with $j \neq i$.

\begin{figure}[t]
    \centering
    \setlength{\tabcolsep}{1pt}
    \scriptsize
    \resizebox{0.9\linewidth}{!}{
    \begin{tabular}{ccccc}
    \includegraphics[width=0.2\linewidth,trim={.55cm 0 20.1cm 0.75cm}, clip]{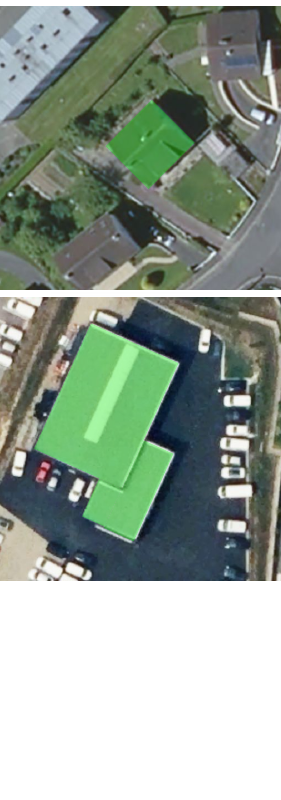} &
    \includegraphics[width=0.2\linewidth,trim={5.45cm 0 15.2cm 0.75cm}, clip]{images/siou_vs_xor.pdf} &
    \includegraphics[width=0.2\linewidth,trim={10.4cm 0 10.25cm 0.75cm}, clip]{images/siou_vs_xor.pdf} &
    \includegraphics[width=0.2\linewidth,trim={15.275cm 0 5.375cm 0.75cm}, clip]{images/siou_vs_xor.pdf} &
    \includegraphics[width=0.2\linewidth,trim={20.3cm 0 0.35cm 0.75cm}, clip]{images/siou_vs_xor.pdf}\vspace{-5em}\\
    (a) $I_t$ & (b) $I_{t'}$ & (c) Overlapped & (d) XOR & (e) sIoU \\  
    & & masks & change map & change map
    \end{tabular}
    }\vspace{-.5em}
    \caption{\textbf{XOR vs. sIoU for change map generation} from image pairs, for the building change detection binary task. Top: a real pair with slight viewpoint variation—XOR falsely detects changes, while sIoU correctly shows none. Bottom: a \textit{fake} pair with different buildings—XOR misses overlapping changes, sIoU correctly marks both. Only one building per image is labeled for clarity.
    \vspace{-0.5em}} 
    \label{fig:siou_vs_xor}
\end{figure}

\paragraph{Change map generation.} The models trained with our method usually supervise their predicted maps ($\hat{S}_t$, $\hat{S}_{t'}$, $\hat{M}$) with ground-truth maps ($S_t$, $S_{t'}$, $M$). In our weakly supervised setting, we only have access to $S_t$. We create missing semantic and change maps distinguishing two cases, depending on if the pair is real or fake. For real pairs, we deliberately (though incorrectly in general) assume no semantically relevant change between the two acquisitions, and define:
\begin{equation}
S_{t'} = S_t~, \quad M = 0~.
\end{equation} 
For fake pairs ($I_{t_i}^i$, $I_{t_j'}^j$), we use the  semantic mask of $I_{t_j}^j$ as proxy of the  unavailable mask $I_{t_j'}^j$, again assuming no change between the two acquisitions. In order to generate a change mask $M$ from this pair of non-spatially aligned maps $S_{t_i}$ and $S_{t_j}$, Zheng \textit{et al.}~\cite{zheng2021change} apply an XOR operation, which only consider the semantic temporal differences at the pixel level. Instead, we choose to use an object-aware metric that considers changes at the object level. To this end, we generate a change mask by thresholding the sIoU~\cite{rottmann2020prediction, chan2021segment} between connected components of $S_{t_i}$ and $S_{t_j}$. The sIoU is an 
object-level variation of the traditional intersection over union (IoU). Unlike the conventional IoU, which penalizes fragmented ground truth regions by assigning each prediction a moderate IoU score, the sIoU does not penalize predictions for a segmentation if other predicted segmentations sufficiently cover the remaining ground truth. This enables a robust evaluation of changes at the object level, especially when accounting for shadow occlusions, deformations, or label inconsistencies. Let $\mathcal{C}_{t_i}$ and $\mathcal{C}_{t_j}$ denote the sets of connected components in $S_{t_i}$ and $S_{t_j}$ respectively, where each component corresponds to a semantic class. We compute the sIoU between each component $c^k$ in $\mathcal{C}_{t_i}$, characterized by its semantic category $k$, with respect to $\mathcal{C}_{t_j}$ as
\begin{equation}
        \text{sIoU}_{\mathcal{C}_{t_j}}(c^k) := \frac{\left| c^k \cap \mathcal{C}_{t_j}(c^k) \right|}{\left| \left(c^k \cup \mathcal{C}_{t_j}(c^k) \right) \setminus \mathcal{A}(c^k) \right|}~, 
\end{equation}
with
\begin{equation}
        \mathcal{C}_{t_j}(c^k) = \bigcup_{\substack{\bar{c}^l \in \mathcal{C}_{ t_j}\\ \bar{c}^l \cap c^k \neq \varnothing,\ l=k}}\bar{c}^l \text{ and } 
        \mathcal{A}(c^k) = \bigcup_{\substack{\bar{c}^l \in \mathcal{C}_{t_i} \setminus\{c^k\}\\l=k}} \bar{c}^l.
\end{equation}
The notation $\left|\cdot\right|$ denotes the size of a set in terms of number of pixels. The term \(\mathcal{A}(c^k)\) excludes from the denominator any pixels belonging to other $\mathcal{C}_{t_i}$ components that overlap with \(c^k\) and share the same semantic class $k$, preventing interference from nearby objects. During training, we apply this object-level comparison to each fake pair in the batch. For each pair of semantic maps $(S_{t_i}, S_{t_j})$, we compute the sIoU for all components in both directions: $\mathcal{C}_{t_i}$ with respect to $\mathcal{C}_{t_j}$, and vice versa. A pixel location ($x,y$) of the binary change mask $M$ is labeled as a change if it belongs to any connected component in either $\mathcal{C}_{t_i}$ or $\mathcal{C}_{t_j}$ for which the sIoU is below a predefined threshold $\tau$:
\begin{equation}
    M(x,y) = \begin{cases}
    1 & \text{if } (x,y)\in \displaystyle\bigcup_{\substack{c\in\mathcal{C}_{t_i}\\ \text{sIoU}_{\mathcal{C}_{t_j}(c)<\tau}}} c~, \\
    1 & \text{if } (x,y)\in \displaystyle\bigcup_{\substack{c\in\mathcal{C}_{t_j}\\ \text{sIoU}_{\mathcal{C}_{t_i}(c)<\tau}}} c~, \\
    0 & \text{otherwise}.
\end{cases}
\end{equation}
Fig.~\ref{fig:siou_vs_xor} provides a visual explanation of the motivations and effects of the object-level change map generation based on the sIoU. In particular, our method is consistent with labeling co-registered real pairs with zero change maps, whereas XOR change maps are sensitive to viewpoint changes.

\paragraph{Iterative refinement.} The rough assumption that real bi-temporal pairs can be labeled with a zero change mask introduces noise in the supervision signal. Indeed, genuine semantic changes may have occurred during the several years separating the original and new acquisitions added in our extended datasets. This leads to mislabeled training samples where changes are incorrectly treated as unchanged, which can degrade model performance. To mitigate this, we adopt an iterative refinement strategy. First, we train a model on the entire dataset. We then use its predictions to identify and filter out image pairs likely to contain changes. In practice, we remove all image pairs for which the model predicts more than 2\% of changed pixels. In the next iteration, we train a model from scratch using this cleaned dataset, thus improving training stability and overall prediction quality. We repeat this for $N_\text{iter}$ iterations, where $N_\text{iter}$ is a hyperparameter of the method. Ignoring a subset of the dataset by removing noisy or incorrect samples, also known as data cleaning, has been proven successful in other works~\cite{guyon1994discovering, john1995robust, jeatrakul2010data, cote2024data}.

\subsection{Implementation}
\label{sec:implementation}
Recent analyses show that simple baselines like UNet remain highly competitive for SCD despite the proliferation of complex task-specific architectures~\cite{cd_reality_check}. As architectural design is beyond the scope of this paper, we use a 3-branch Dual UNet, following Benidir \textit{et al}~\cite{benidir2025change}. Note that our methodology is architecture agnostic and thus can also be applied to any 3-branch models (see Sec.~\ref{sec:architectures}).

We train our models in a multi-task setting, supervising the semantic segmentation at both dates as well as the binary change detection with corresponding focal losses. All models are trained using the AdamW optimizer~\cite{loshchilov2018decoupled} with a 1e-2 weight decay over 100 epochs and a learning rate of 1e-4. We keep the model with the best validation IoU on the segmentation task over the original sample $t$. We only use the RGB bands of the images in the datasets, and focus on building change detection when not explicitly stated otherwise. For example, with FLAIR-based datasets, we ignore infrared and elevation bands, merging all non-building classes into a ``background'' class. Our hyperparameters are set to $p_\text{real} = 0.25$, $\tau = 0.25$, and $N_\text{iter} = 3$.

\section{Experiments}
\paragraph{Baselines.} We compare our methodology with four frameworks that use single-temporal semantic segmentation annotations for change detection: post-classifica\-tion with and without multi-temporal data augmentation, synthetic dataset learning, and STAR~\cite{zheng2021change}, which generates \textit{fake change pairs} by pairing images from different locations in single-temporal datasets. For post-classification, we train a UNet~\cite{ronneberger2015u} on the semantic segmentation task and detect building changes by comparing bi-temporal predictions. Bi-temporal data augmentation is achieved by using our extended data during training. For the other frameworks, we train a Dual UNet~\cite{benidir2025change} with a multi-task objective. We consider three synthetic datasets: SyntheWorld~\cite{song2024syntheworld}, FSC-180k~\cite{benidir2025change}, and a FLAIR-based dataset built with the Changen2~\cite{zheng2024changen2} pipeline and which we refer to as Changen2-S19-180k following the naming conventions in~\cite{zheng2024changen2}.
Segment Any Change~\cite{NEURIPS2024_94154162} is excluded from our quantitative evaluation because it predicts class-agnostic changes and requires a $(x,y)$ point prompt from one of the input images to specify the target semantics.
Additional details on the creation process of Changen2-S19-180k and a qualitative comparison with Segment Any Change are provided in the Supplementary.

\paragraph{Metrics.} We adopt F1-score (F1) and intersection over union (IoU) as evaluation metrics for building change detection. These scores are reported as percentages. We also report the false positive rates (FPR), and the number and size of connected components to evaluate false alarms.

\begin{figure*}[t]
    \centering
    \resizebox{0.9\linewidth}{!}{
    \setlength{\tabcolsep}{1pt}   
    \renewcommand{\arraystretch}{0.1}
    \scriptsize
    \begin{tabular}{
        *{9}{>{\centering\arraybackslash}p{0.11\textwidth}}
    }

    \qualitative{693} \\
    \qualitative{1162} \\
    \qualitative{1579} \\
    & & \\
    \tiny Image $I_t$ & \tiny Image $I_{t'}$ & \tiny Ground &  \tiny Post-classif. &
    \tiny Post-classif. & \tiny Changen2 & \tiny FSC-180k & \tiny STAR & \tiny Ours \\
     & & \tiny truth & & \tiny +temp. aug. & \tiny pretraining & \tiny pretraining & & \\

    \end{tabular}
    }
    \vspace{-.5em}
    \caption{\textbf{Qualitative results on FLAIR-BCD.}
    We compare the building change maps predicted by baseline methods and ours.
    }
    \label{fig:qualitative_results}
    \vspace{-1em}
\end{figure*}
\paragraph{In-domain results.}
Tab.~\ref{tab:comparison_results} reports in-domain performance on the FLAIR-BCD-heldout and FLAIR-BCD-spot datasets (described in Sec.~\ref{sec:data}), with qualitative examples in Fig.~\ref{fig:qualitative_results}. Our method achieves the second-best score on FLAIR-BCD-heldout, while performance slightly drops on FLAIR-BCD-spot. We attribute this to two main factors.
(1)~Our weak supervision relies on temporal augmentation, which can introduce minor misalignments or imperfect change annotations. This produces slightly \textit{dilated} predictions, smoother boundaries and less defined corners, affecting IoU and F1 scores.
(2)~The model assumes most regions remain unchanged between acquisitions, biasing predictions toward the no-change class. Given that change detection datasets are dominated by the background class, false negatives penalize IoU and F1 more than false positives.

Beyond quantitative results, Figs.~\ref{fig:teaser} and \ref{fig:qualitative_results} confirm that our model is notably more robust in no-change regions, producing fewer false positives. We argue that this behavior is desirable: in large-scale deployments, minimizing false alarms is often more critical than achieving perfectly sharp change boundaries. Therefore, we complement our evaluation with an analysis of the FPR and connected components, provided in the Supplementary. 

\paragraph{Out-of-domain results.}
We further evaluate zero-shot building change detection on LEVIR-CD, WHU-CD, and S2Looking (Tab.~\ref{tab:comparison_results}). Comparisons are made across models trained on the same data source (FLAIR, SPOT, and IAILD), with the best results highlighted for each setting. On WHU-CD and S2Looking, our method achieves the best scores within each of the three training domains, and the b-IAILD variant is the best overall on both datasets. On S2Looking, our b-FLAIR variant leads Changen2-S19-180k only by a small margin. On LEVIR-CD, only the b-FLAIR-spot variant outperforms its counterparts. We conduct in the Supplementary, an in-depth analysis of the results on LEVIR-CD for the b-FLAIR model variants.

\begin{table}[t!]
  \caption{\textbf{Low data regimes.} IoU scores on LEVIR-CD and S2Looking datasets when fine-tuning a Dual UNet on limited target data (1\%, 10\% and 30\%). Each result is averaged over 10 runs. Bold denote the best result for each dataset. $^\dagger$Results from~\cite{benidir2025change}.}
  \setlength{\tabcolsep}{4pt}
  \vspace{-2mm}
  \centering
  \renewcommand{\arraystretch}{0.85}
  \scriptsize
  \resizebox{0.9\linewidth}{!}{
  \begin{tabular}{lccc}
    \toprule
    Pretraining & \multicolumn{3}{c}{Proportion of target data used} \\
    \cmidrule(lr){2-4}
    dataset & 1\% & 10\% & 30\% \\
    \midrule
    & \multicolumn{3}{c}{LEVIR-CD} \\
    \cmidrule(lr){2-4}
    No pre-training $^\dagger$ & 0.36 & 0.69 & 0.75 \\
    SyntheWorld$^\dagger$ & 0.41 (+14\%) & 0.71 (+3\%) & 0.77 (+3\%) \\
    FSC-180k$^\dagger$ & 0.55 (+53\%) & 0.73 (+6\%) & 0.79 (+5\%) \\
    \rowcolor{blue!10}Ours (b-FLAIR) & 0.60 (+67\%) & 0.77 (+12\%) & 0.81 (+8\%) \\
    \rowcolor{blue!10}Ours (b-IAILD) & \textbf{0.67} (+86\%) & \textbf{0.79} (+14\%) & \textbf{0.82} (+9\%) \\
    \midrule
    & \multicolumn{3}{c}{S2Looking} \\
    \cmidrule(lr){2-4}
    No pre-training$^\dagger$ & 0.10 & 0.27 & 0.38 \\
    SyntheWorld$^\dagger$ & 0.08 (-20\%) & 0.34 (+26\%) & 0.41 (+8\%) \\
    FSC-180k$^\dagger$ & 0.15 (+50\%) & 0.36 (+33\%) & \textbf{0.43} (+13\%) \\
    \rowcolor{blue!10}Ours (b-FLAIR) & 0.24 (+140\%) & 0.31 (+15\%) & 0.41 (+8\%) \\
    \rowcolor{blue!10}Ours (b-IAILD) & \textbf{0.27} (+170\%) & \textbf{0.37} (+37\%) & 0.41 (+8\%) \\
    \bottomrule
  \end{tabular}
  }
  \label{tab:low_data_regime}
\end{table}

\paragraph{Results in low-data regime.}  We finetune a Dual UNet, pretrained with our methodology, on small subsets of 1\%, 10\%, and 30\% of training data for LEVIR-CD and S2Looking.  This evaluates the quality of the learned representations when training with very few samples, a common practical scenario. Following Benidir \textit{et al.}~\cite{benidir2025change}, we randomly sample training sets from the target dataset and average results over 10 runs. Tab.~\ref{tab:low_data_regime} reports results for our b-FLAIR and b-IAILD models, compared with FSC-180k and SyntheWorld from~\cite{benidir2025change}. Our approach achieves substantial improvements when only minimal annotations are available (e.g., 1\%), while performance converges as more data is available.

\paragraph{Multi-class change detection results. }
In the absence of a test set enabling multi-class semantic change detection evaluation for methods trained on FLAIR-based datasets, we evaluate our method on the multi-class datasets Hi-UCD mini~\cite{hiucd} and Hi-UCD~\cite{tian2022large}, which respectively contain 1200 and 12,000 training images of size $512 \times 512$. Both include the semantic classes \textit{water}, \textit{grass}, \textit{building}, \textit{greenhouse}, \textit{road}, \textit{bridge}, \textit{bare land}, \textit{woodland}, and \textit{others}, and provide a semantic mask for all images. 
In our experiments, we discard the semantic labels at $t'$ and apply our method in a multi-class setting. Results on the Hi-UCD mini test set are reported in Tab.~\ref{tab:multiclass_eval}, where we compare our approach with a post-classification change detection baseline using temporal augmentation and STAR. Our method consistently outperforms the baselines in binary change detection (IoU), while STAR achieves the best multi-class segmentation (mIoU).

\begin{table}[t!]
  \caption{\textbf{Results on the multi-class Hi-UCD mini dataset.} Comparison between the single-temporal baseline, STAR and our method. Models are trained either on the Hi-UCD mini training set or on the full Hi-UCD training split.}
  \centering  
  \renewcommand{\arraystretch}{0.85}
  \vspace{-2mm}
  \scriptsize
  \resizebox{\linewidth}{!}{
  \begin{tabular}{l cc cc}
  \toprule
    & \multicolumn{2}{c}{Hi-UCD mini} & \multicolumn{2}{c}{Hi-UCD} \\
    \cmidrule(lr){2-3}\cmidrule(lr){4-5}    
    Method & Ch. IoU & Seg. mIoU & Ch. IoU & Seg. mIoU \\
    \midrule
    Post-classification + temp. aug. (UNet) & 14.4 & 37.7 & 15.0 & 38.2 \\
    STAR~\cite{zheng2021change} (Dual UNet) & 45.1 & 26.3 & 45.3 & \textbf{48.3} \\
    \rowcolor{blue!10}Ours (Dual UNet) & \textbf{47.2} & \textbf{39.7} & \textbf{47.2} & 46.7 \\
    \bottomrule
  \end{tabular}
  }
  \label{tab:multiclass_eval}
\end{table}

\subsection{Different architectures}
\label{sec:architectures}
We report performances using two different architectures: Dual UNet~\cite{benidir2025change} (65.05\,M parameters) and SCanNet~\cite{ding2024joint} (27.9\,M parameters).
Fig.~\ref{fig:model_comparison} (left) presents the F1 scores on FLAIR-BCD for the FSC-180k~\cite{benidir2025change} approach, the STAR~\cite{zheng2021change} baseline and our method. Fig.~\ref{fig:model_comparison} (right) reports the same evaluation in a zero-shot setting on the WHU-CD dataset.
Our method achieves competitive results when using both Dual UNet and SCanNet backbones across both datasets.

\begin{figure}[t]
\centering
\begin{minipage}{0.49\linewidth}
\centering
\resizebox{0.97\linewidth}{!}{
\definecolor{fsc}{RGB}{136, 204, 238}    
\definecolor{ours}{RGB}{51, 34, 136}   
\definecolor{star}{RGB}{204, 102, 119} 
\begin{tabular}{ccc}
\begin{minipage}[t]{0.3\textwidth}
\centering
\setlength{\tabcolsep}{2pt}
\begin{tabular}{@{\hspace{5mm}}rl@{\hspace{10mm}}rl@{\hspace{10mm}}rl}
\tikz{\fill[fsc, draw=fsc, very thick] (0,0) rectangle (0.5,0.2);} & \small FSC-180k~\cite{benidir2025change} &
\tikz{\fill[star, draw=star, very thick] (0,0) rectangle (0.5,0.2);} & \small STAR~\cite{zheng2021change} &
\tikz{\fill[ours, draw=ours, very thick] (0,0) rectangle (0.5,0.2);} & \small Ours 
\end{tabular}
\end{minipage}
\\
\hspace{-6mm}
\begin{minipage}[t]{0.49\linewidth}
\begin{tikzpicture}
    \begin{axis}[
        title style={align=center, font=\footnotesize},
        width=90pt,
        height=100pt,
        title={},
        ylabel={F1},
        ylabel style={rotate=-90, at={(axis description cs:0.7,1)}, font=\footnotesize},
        ybar,
        bar width=9pt,
        enlarge y limits=0.15,
        enlarge x limits=0.3,
        symbolic x coords={fsc, star, ours},
        xtick={fsc, star, ours},
        xticklabels={,,},
        x tick label style={rotate=45, anchor=east, font=\footnotesize},
        y tick label style={font=\footnotesize},
        ytick={40, 60, 80},
        ymin=35,
        ymax=85,
        bar shift=3pt,
        axis y line*=left,
        axis x line*=bottom,
        xmajorticks=false,
    ]     
    \addplot[fill=star,draw=star, very thick] coordinates {(star, 75.9)};
    
    \addplot[fill=fsc, draw=fsc, very thick] coordinates {(fsc,83.1)};

    \addplot[fill=ours,draw=ours, very thick] coordinates {(ours, 79.0)};
    
    \end{axis}
    \node[font=\footnotesize, align=center] at (0.7,-0.45) {Dual UNet~\cite{benidir2025change}};
\end{tikzpicture}
\end{minipage}
&
\begin{minipage}[t]{0.49\linewidth}
\begin{tikzpicture}
    \begin{axis}[
        title style={align=center, font=\footnotesize},
        width=90pt,
        height=100pt,
        title={},
        ylabel={F1},
        ylabel style={rotate=-90, at={(axis description cs:0.7,1)}, font=\footnotesize},
        ybar,
        bar width=9pt,
        enlarge y limits=0.15,
        enlarge x limits=0.3,
        symbolic x coords={fsc, star, ours},
        xtick={fsc, star, ours},
        xticklabels={,,},
        x tick label style={rotate=45, anchor=east, font=\footnotesize},
        y tick label style={font=\footnotesize},
        ytick={40, 60, 80},
        ymin=35,
        ymax=85,
        bar shift=3pt,
        axis y line*=left,
        axis x line*=bottom,
        xmajorticks=false,
    ]     
    \addplot[fill=star,draw=star, very thick] coordinates {(star, 66.2)};
    
    \addplot[fill=fsc, draw=fsc, very thick] coordinates {(fsc,43.3)};

    \addplot[fill=ours,draw=ours, very thick] coordinates {(ours, 81.8)};
    
    \end{axis}
    \node[font=\footnotesize, align=center] at (0.7,-0.45) {SCanNet~\cite{ding2024joint}};
\end{tikzpicture}
\end{minipage}
\end{tabular}}
\subcaption[a]{Results on FLAIR-BCD}
\end{minipage}
\hfill
\begin{minipage}{0.49\linewidth}
\centering
\resizebox{0.97\linewidth}{!}{
\definecolor{fsc}{RGB}{136, 204, 238}     
\definecolor{ours}{RGB}{51, 34, 136}     
\definecolor{star}{RGB}{204, 102, 119}   

\begin{tabular}{ccc}
\begin{minipage}[t]{0.3\textwidth}
\centering
\setlength{\tabcolsep}{2pt}
\end{minipage}
\\
\hspace{-6mm}
\begin{minipage}[t][2.2cm][t]{.49\linewidth}
\begin{tikzpicture}
    \begin{axis}[
        title style={align=center, font=\footnotesize},
        width=90pt,
        height=100pt,
        title={},
        ylabel={F1},
        ylabel style={rotate=-90, at={(axis description cs:0.7,1)}, font=\footnotesize},
        ybar,
        bar width=9pt,
        enlarge y limits=0.15,
        enlarge x limits=0.3,
        symbolic x coords={fsc, star, ours},
        xtick={fsc, star, ours},
        xticklabels={,,},
        x tick label style={rotate=45, anchor=east, font=\footnotesize},
        y tick label style={font=\footnotesize},
        ytick={60, 70, 80},
        ymin=60,
        ymax=80,
        bar shift=3pt,
        axis y line*=left,
        axis x line*=bottom,
        xmajorticks=false,
    ]     
    \addplot[fill=star,draw=star, very thick] coordinates {(star, 70.6)};
    
    \addplot[fill=fsc, draw=fsc, very thick] coordinates {(fsc,63.3)};

    \addplot[fill=ours,draw=ours, very thick] coordinates {(ours, 77.3)};
    
    \end{axis}
    \node[font=\footnotesize, align=center] at (0.7,-0.45) {Dual UNet~\cite{benidir2025change}};
\end{tikzpicture}
\end{minipage}
&
\begin{minipage}[t][2.2cm][t]{.49\linewidth}
\begin{tikzpicture}
    \begin{axis}[
        title style={align=center, font=\footnotesize},
        width=90pt,
        height=100pt,
        title={},
        ylabel={F1},
        ylabel style={rotate=-90, at={(axis description cs:0.7,1)}, font=\footnotesize},
        ybar,
        bar width=9pt,
        enlarge y limits=0.15,
        enlarge x limits=0.3,
        symbolic x coords={fsc, star, ours},
        xtick={fsc, star, ours},
        xticklabels={,,},
        x tick label style={rotate=45, anchor=east, font=\footnotesize},
        y tick label style={font=\footnotesize},
        ytick={60, 70, 80},
        ymin=60,
        ymax=80,
        bar shift=3pt,
        axis y line*=left,
        axis x line*=bottom,
        xmajorticks=false,
    ]     
    \addplot[fill=star,draw=star, very thick] coordinates {(star, 68.7)};
    
    \addplot[fill=fsc, draw=fsc, very thick] coordinates {(fsc,66.8)};

    \addplot[fill=ours,draw=ours, very thick] coordinates {(ours, 72.4)};
    
    \end{axis}
    \node[font=\footnotesize, align=center] at (0.7,-0.45) {SCanNet~\cite{ding2024joint}};
\end{tikzpicture}
\end{minipage}
\end{tabular}}
\subcaption[b]{Results on WHU-CD}
\end{minipage}
\vspace{-2mm}
\caption{\textbf{Ablation on different backbones.} F1 scores of our method compared to FSC-180k~\cite{benidir2025change} and STAR~\cite{zheng2021change}, for the DualUnet~\cite{benidir2025change} and SCanNet~\cite{ding2024joint} architectures.}
\label{fig:model_comparison}
\end{figure}

\subsection{Comparison with STAR} 

We provide a fine-grained comparison of our method with the STAR~\cite{zheng2021change} baseline. The differences between our method and STAR include the loss function, the use of real pairs in the batch, sIoU-based change masks, and iterative refinement. Regarding the loss function, we opted for a Focal loss over the objective described in Eqs. (4--6) in~\cite{zheng2021change}. We isolate the respective contributions of the key differences in Tab.~\ref{tab:STARcomparison}.
\begin{table}[t]
    \centering
    \caption{\textbf{Comparison with STAR.} Training on b-FLAIR subset and evaluating on FLAIR-BCD.}
    \resizebox{\linewidth}{!}{
    \begin{tabular}{lccccccc}
        \toprule
        Method & Focal loss & XOR & sIoU & Real pairs & $F1\uparrow$ & $IoU\uparrow$ & $FPR\downarrow$ \\   
        \midrule
        STAR~\cite{zheng2021change} & \xmark & \cmark & \xmark & \xmark & 66.3 & 49.6 & 2.51\\
        \midrule
         & \cmark & \cmark & \xmark & \xmark & 67.4 & 50.8 & 2.42 \\ 
         & \cmark & \xmark & \cmark & \xmark & 72.6 & 56.6 & 1.37 \\ 
         & \cmark & \cmark & \xmark & \cmark & 76.0 & 61.3 & 0.60 \\ 
        \rowcolor{blue!10} Ours (1 Iter.) & \cmark & \xmark & \cmark & \cmark & \textbf{77.0} & \textbf{62.6} & \textbf{0.45} \\ 
        \bottomrule
    \end{tabular}
    }
    \label{tab:STARcomparison}
\end{table}

\subsection{Hyperparameters analysis}
For $\tau$ and $p_\text{real}$, we report results from models trained on a subset of b-FLAIR.
More details and visualizations discussing hyperparameters can be found in the Supp. Mat.

\begin{table}[t]
\caption{\textbf{Impact of $p_\text{real}$ and $\tau$.} Left: Comparison of different values for the proportion $p_\text{real}$ of real bi-temporal pairs. Right: Comparison between XOR change maps and our sIoU-based maps with different threshold $\tau$. Scores reported on FLAIR-BCD.}
\label{tab:preal_siou_comparison_bflair}

\centering
\small
\setlength{\tabcolsep}{4pt}
\vspace{-2mm}
\renewcommand{\arraystretch}{0.85}
\scriptsize
\begin{subtable}{0.48\linewidth}
\centering
\begin{tabular}{lcc}
\toprule
$p_{\text{real}}$ & F1 & IoU \\
\midrule
0.00 & 72.3 & 56.6 \\
0.10 & 76.1 & 61.4 \\
0.25 & \textbf{77.0} & \textbf{62.6} \\
0.35 & 73.8 & 58.5 \\
0.50 & 73.6 & 58.2 \\
\bottomrule
\end{tabular}
\end{subtable}
\hfill
\begin{subtable}{0.48\linewidth}
\centering
\begin{tabular}{ccccc}
\toprule
XOR & sIoU & $\tau$ & F1 & IoU \\
\midrule
\cmark &  & ---  & 76.0 & 61.3  \\
 & \cmark & 0.75 & 74.8 & 59.7 \\
 & \cmark & 0.50 & 73.1 & 57.8  \\
 & \cmark & 0.25 & \textbf{77.0} & \textbf{62.6} \\
\bottomrule
\end{tabular}
\end{subtable}
\end{table}

\paragraph{Impact of $p_\text{real}$.} Tab.~\ref{tab:preal_siou_comparison_bflair} shows adding real pairs in a batch improves performance, with $p_\text{real}=0$ yielding the lowest scores. Performance peaks at $p_\text{real}=0.25$, while higher values degrade it, likely because overrepresented real pairs with no change bias the model toward predicting no change.

\paragraph{Impact of $\tau$.} As shown in Tab.~\ref{tab:preal_siou_comparison_bflair}, a threshold of $\tau=0.25$ for sIoU-based change map generation outperforms the logical XOR operation. On the contrary, higher thresholds ($\tau \in \{0.5, 0.75\}$) tend to classify components as changed despite significant overlap between building footprints across the two masks, resulting in noisy supervision. Note that $\tau=1$ reduces our sIoU-based maps to the OR operation, which Zheng \textit{et al.}~\cite{zheng2024single} demonstrated to be inferior to XOR-based change maps for training change detection models.

\begin{table}[t!]
  \caption{\textbf{Performance across training iterations.} F1 Score (\%) and IoU (\%) of a Dual UNet trained with our method over three training iterations, evaluated on the corresponding in-domain test set. Best results per dataset are shown in bold.}
  \centering
  \scriptsize
  \setlength{\tabcolsep}{4pt} 
  \renewcommand{\arraystretch}{0.85}
  \begin{tabular}{lcccc}
    \toprule
    & \multicolumn{2}{c}{FLAIR-BCD} 
    & \multicolumn{2}{c}{FLAIR-BCD-spot}\\
    \cmidrule(lr){2-3}\cmidrule(lr){4-5}    
    & F1 & IoU & F1 & IoU \\
    \midrule
    Iteration 1 & 67.3 & 50.7 & 22.6 & 12.7 \\
    Iteration 2 & 78.1 & 64.1 & 21.7 & 12.2 \\
    Iteration 3 & \textbf{79.0} & \textbf{65.2} & \textbf{22.9} & \textbf{12.9}\\  
    \bottomrule
  \end{tabular}
  \label{tab:iteration_comparison_bflairspot}
\end{table}

\begin{figure}[t]
    \centering
    \setlength{\tabcolsep}{1pt}
    \begin{tabular}{cccccccc}
    \parbox{4mm}{\rotatebox[origin=c]{90}{$I_{t}$}} &
    \raisebox{-.4\height}{\includegraphics[width=0.13\linewidth]{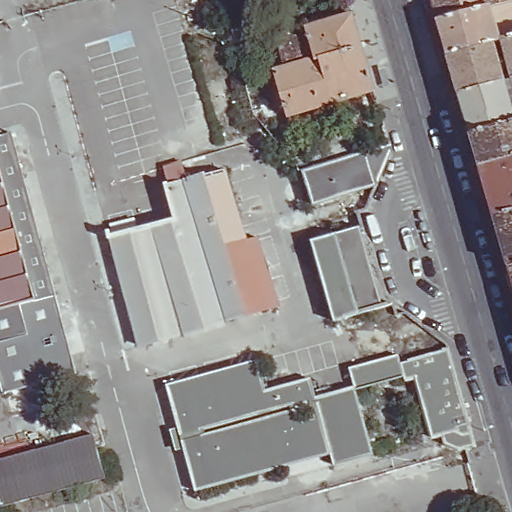}} &
    \raisebox{-.4\height}{\includegraphics[width=0.13\linewidth]{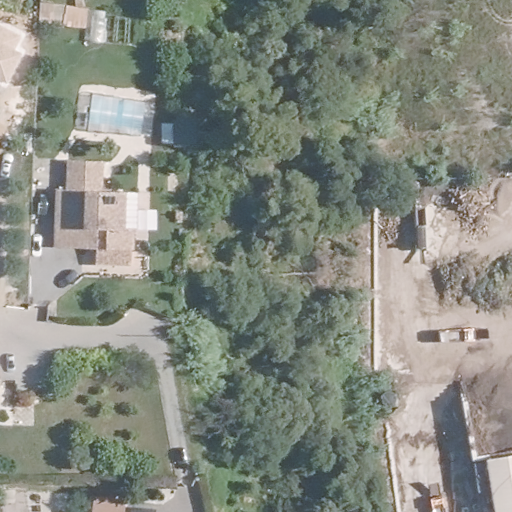}} &
    \raisebox{-.4\height}{\includegraphics[width=0.13\linewidth]{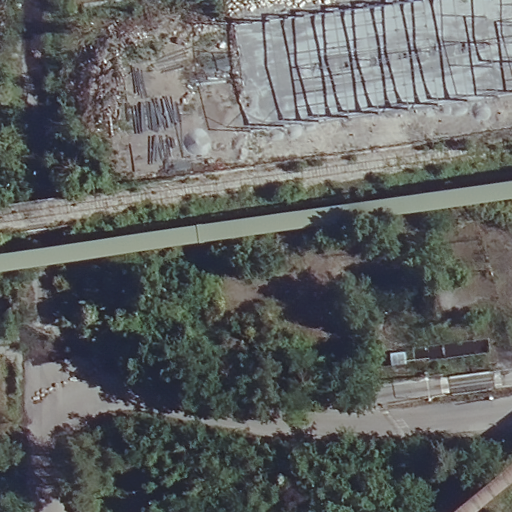}} &
    \raisebox{-.4\height}{\includegraphics[width=0.13\linewidth]{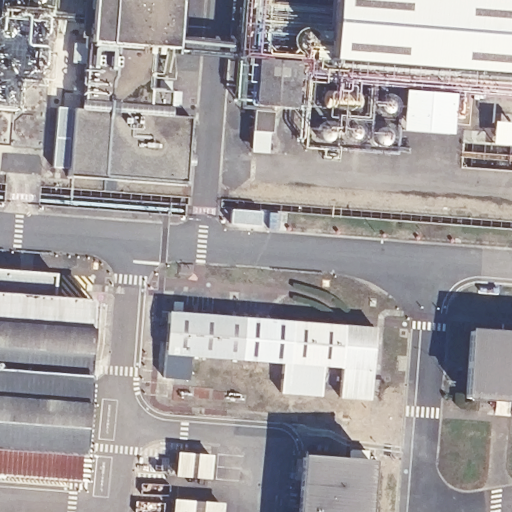}} &
    \raisebox{-.4\height}{\includegraphics[width=0.13\linewidth]{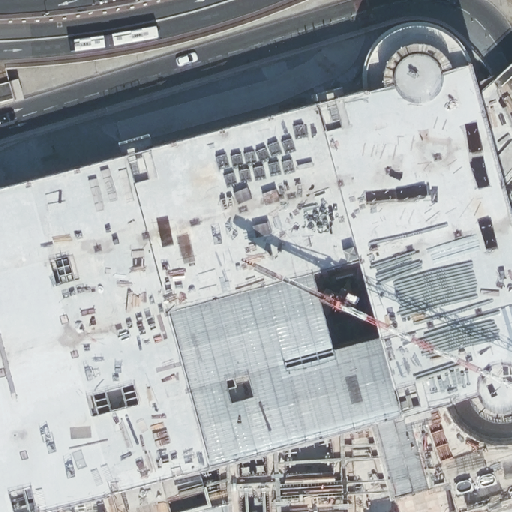}} &
    \raisebox{-.4\height}{\includegraphics[width=0.13\linewidth]{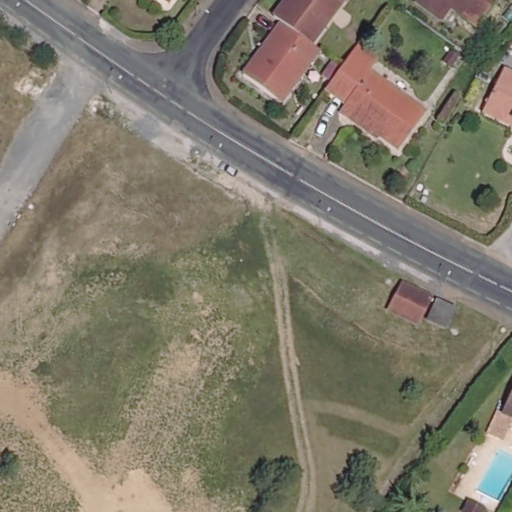}}  &
    \raisebox{-.4\height}{\includegraphics[width=0.13\linewidth]{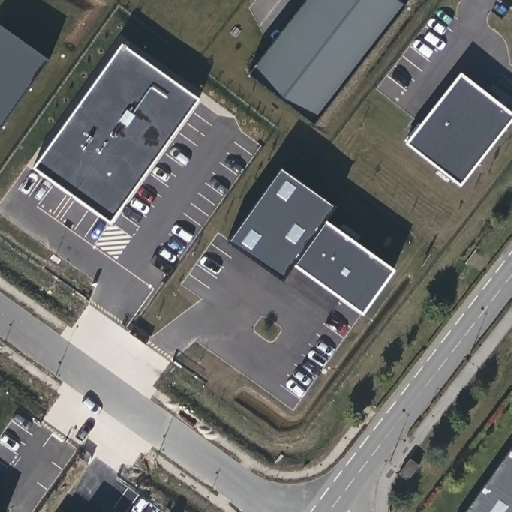}} \vspace{2pt}\\
    \parbox{4mm}{\rotatebox[origin=c]{90}{$I_{t'}$}} & 
    \raisebox{-.4\height}{\includegraphics[width=0.13\linewidth]{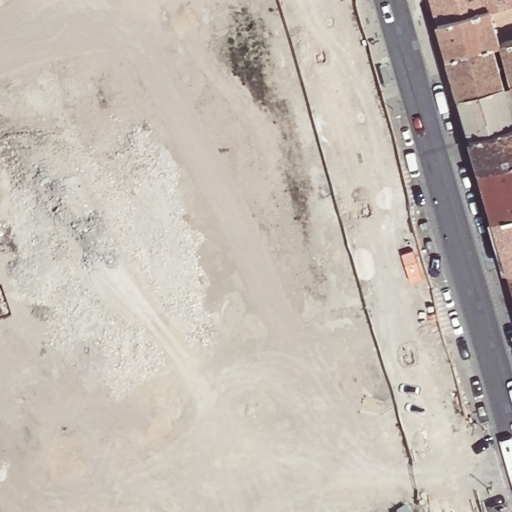}} &
    \raisebox{-.4\height}{\includegraphics[width=0.13\linewidth]{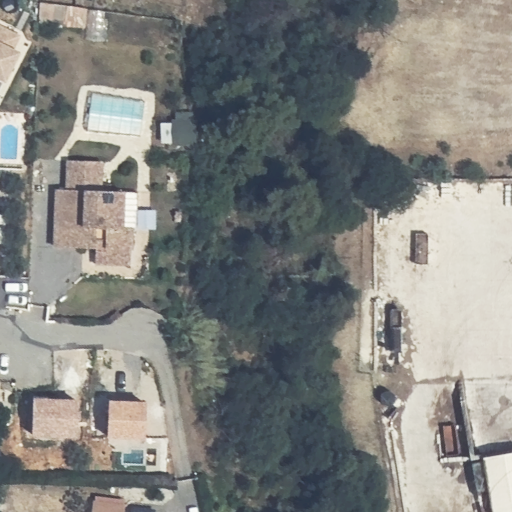}} &
    \raisebox{-.4\height}{\includegraphics[width=0.13\linewidth]{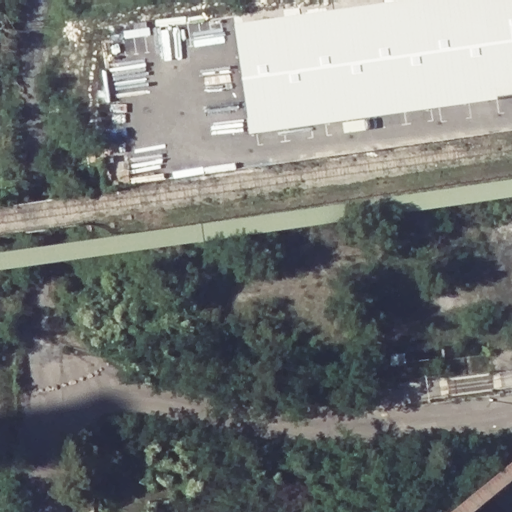}} &
    \raisebox{-.4\height}{\includegraphics[width=0.13\linewidth]{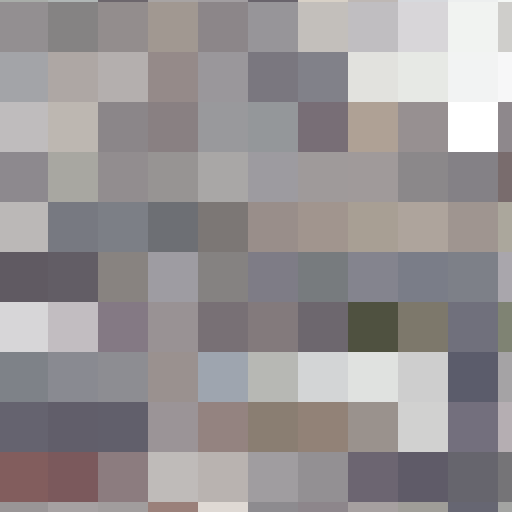}} &
    \raisebox{-.4\height}{\includegraphics[width=0.13\linewidth]{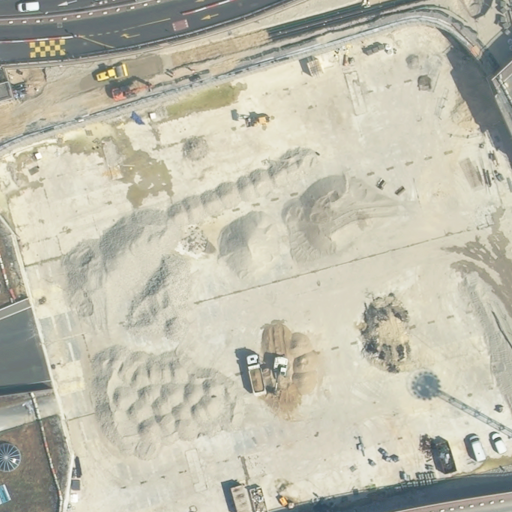}} &
    \raisebox{-.4\height}{\includegraphics[width=0.13\linewidth]{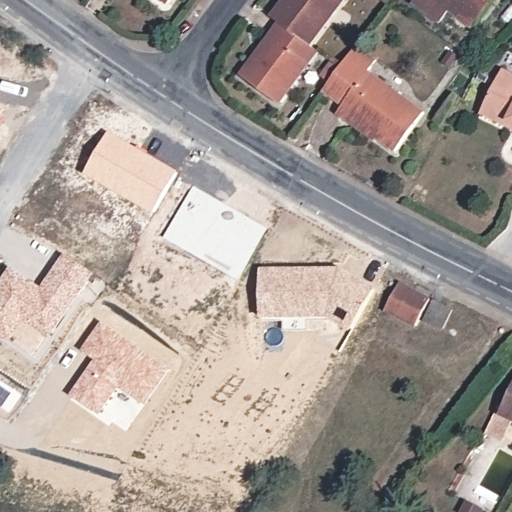}} &
    \raisebox{-.4\height}{\includegraphics[width=0.13\linewidth]{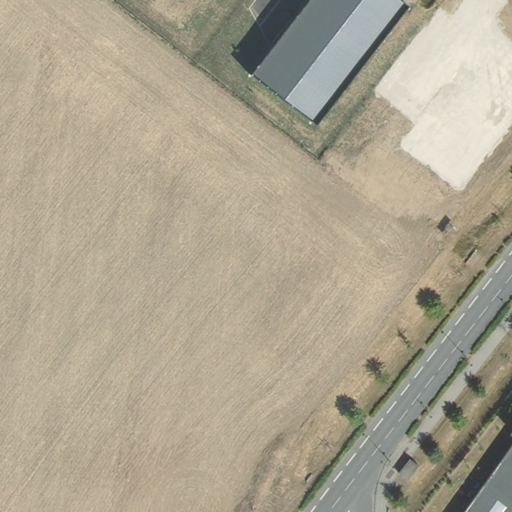}} 
    \vspace{2pt}\\
    \parbox{4mm}{\rotatebox[origin=c]{90}{$\hat{M}$}} & 
    \raisebox{-.4\height}{\includegraphics[width=0.13\linewidth]{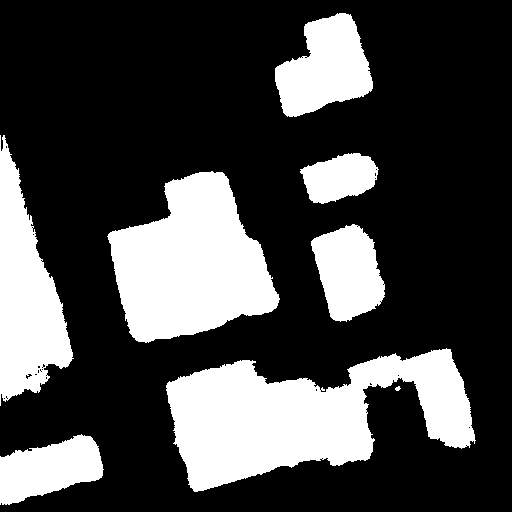}} &
    \raisebox{-.4\height}{\includegraphics[width=0.13\linewidth]{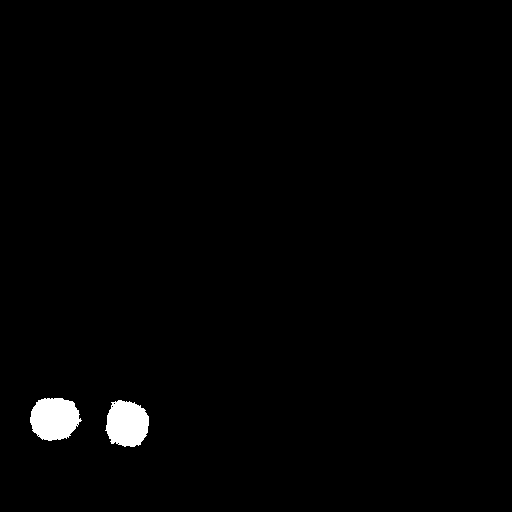}} &
    \raisebox{-.4\height}{\includegraphics[width=0.13\linewidth]{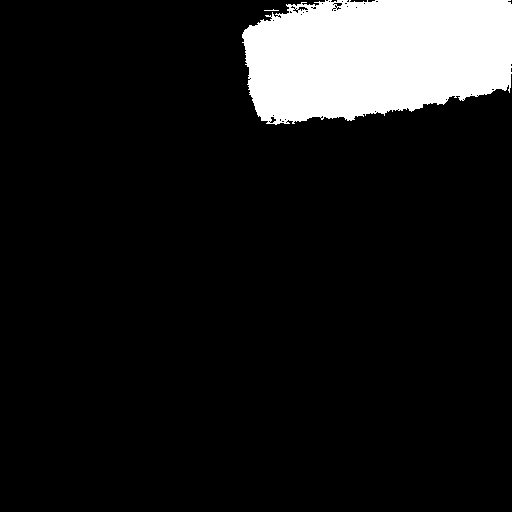}} &
    \raisebox{-.4\height}{\includegraphics[width=0.13\linewidth]{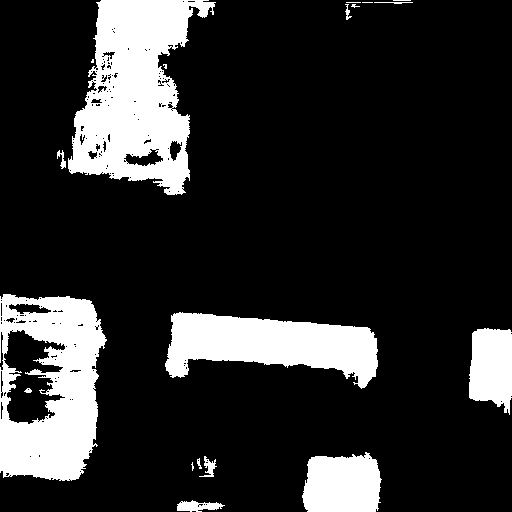}} &
    \raisebox{-.4\height}{\includegraphics[width=0.13\linewidth]{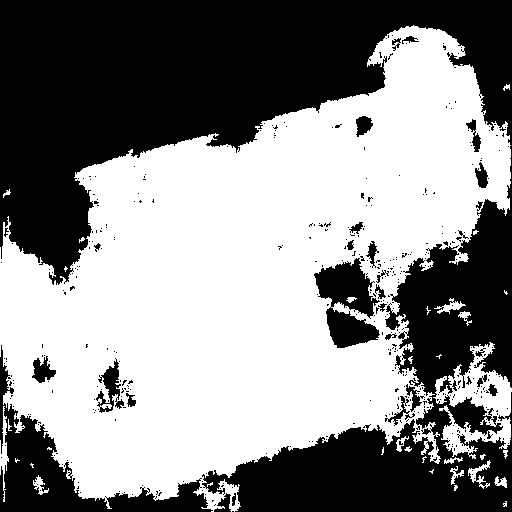}} &
    \raisebox{-.4\height}{\includegraphics[width=0.13\linewidth]{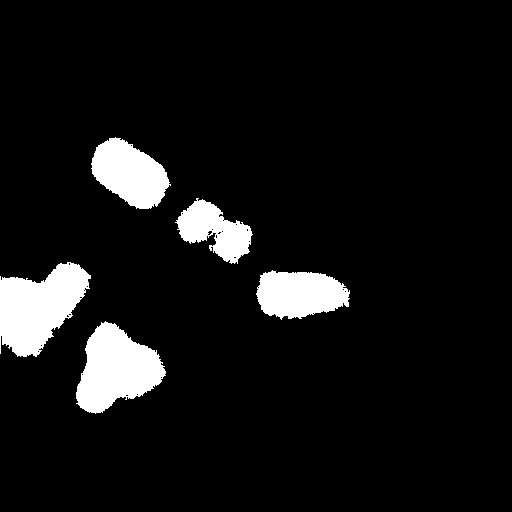}} &
    \raisebox{-.4\height}{\includegraphics[width=0.13\linewidth]{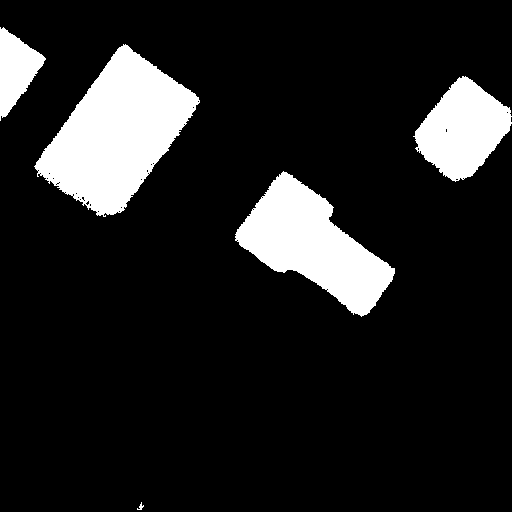}} \\
    \end{tabular}
    \caption{\textbf{Samples removed during iterative refinement.} Triplets ($I_{t}$, $I_{t'}$, $\hat{M}$) removed during refinement exhibit real building changes or blurred sensitive areas, where the assumption $M=0$ does not hold. Such samples are correctly identified by the model and excluded from the training set on subsequent iterations.
    }   
    \label{fig:found_changes}
\end{figure}

\paragraph{Impact of the iterative refinement.} Our methodology relies on mixing bi-temporal image pairs with fake pairs of unaligned images, while assuming by default that all real pairs contain no change. This assumption is incorrect in practice, particularly since our extension of existing single-temporal datasets is based on new acquisitions often captured several years apart. Tab.~\ref{tab:iteration_comparison_bflairspot} therefore demonstrates the benefit of iterative cleaning of the extended datasets, through filtering of image pairs that would contain significant semantic changes. For b-FLAIR, the second iteration increases F1 by over 10pt and IoU by nearly 14pt on FLAIR-BCD, while gains at the third iteration are smaller. We therefore set $N_\text{iter} = 3$ by default. Fig.~\ref{fig:found_changes} illustrates filtered pairs, including building changes and blurred sensitive areas. On b-FLAIR-spot, successive iterations have little effect.

\section{Conclusion}
We introduced a novel approach for training remote sensing change detection models by extending single-date datasets and supervising changes through weak temporal supervision, without additional annotations. We assume no change occurs between two observations of the same location while providing change examples by pairing images from different locations. The results across multiple datasets show that our methodology can transfer to unseen datasets and acquisition conditions, although performance varies depending on the target domain. The resulting models can also be fine-tuned effectively with limited annotations and reduce false positives in large-scale \textit{in-the-wild} scenarios. Qualitative results show good alignment with ground truth building construction and demolition maps, supporting the applicability of the proposed framework to large-scale change monitoring. Overall, our framework provides a practical, annotation-efficient approach to scalable change detection for both aerial and satellite observations.
\newpage
\section*{Acknowledgement}

This work was funded by AID-DGA (l’Agence de l’Innovation de D\'{e}fense \`{a} la Direction G\'{e}n\'{e}rale de l’Armement—Minist\`{e}re des Arm\'{e}es), and was performed using HPC resources from GENCI-IDRIS (grants 2023-AD011011801R3, 2023-AD011012453R2,
2023-AD011012458R2) and from the “M\'{e}socentre” computing center of CentraleSup\'{e}lec and ENS Paris-Saclay supported by CNRS and R\'{e}gion Île-de-France (\href{https://mesocentre.universite-paris-saclay.fr}{http://mesocentre.universite-paris-saclay.fr}). Centre Borelli is also with Universit\'{e} Paris Cit\'{e}, SSA and INSERM. This work is additionally supported by RGC-GRF project 11309925, Mathematical Formalization of GIS. We thank Etienne Bourgeat, Fabien Poilane and Floryne Roche for their valuable assistance in data curation and annotation.

{
\small
\bibliographystyle{templates/wacv/ieeenat_fullname}
\bibliography{biblio}
}

\vspace{1em}
{\noindent \Large \bf Supplementary Material \par}
\vspace{1em}
We first explicit the motivations behind our Weak Temporal Supervision (Sec.~\ref{sec:motivations}) and provide additional details on the datasets used in this work (Sec.~\ref{sec:suppmat_data}). Then we detail our implementation (Sec.~\ref{sec:suppmat_implt_details}). Finally, we report supplementary quantitative and qualitative results (Sec.~\ref{sec:suppmat_additionnal_results}). See \url{https://xavibou.github.io/CDviaWTS/} for our code and datasets 

\setcounter{table}{0}
\renewcommand{\thetable}{A\arabic{table}}
\setcounter{equation}{0}
\renewcommand{\theequation}{A\arabic{equation}}
\setcounter{figure}{0}
\renewcommand{\thefigure}{A\arabic{figure}}
\setcounter{figure}{2}
\renewcommand{\thefigure}{A\arabic{figure}}
\renewcommand{\thesection}{A}
\section{Motivations}\label{sec:motivations}

The main objective of this work is to address the scarcity of annotations for the task of change detection in remote sensing. This scarcity is mainly due to the cost of annotating image pairs with pixel-precise change masks. While prior works rely on artificially constructed pairs, we leverage opportunistic multi-temporal acquisitions of the same geographic location, together with a single semantic annotation per location, to derive supervision for bi-temporal change detection. Thus, the novelty lies in the supervision paradigm: we demonstrate that reliable change detection can be learned from real repeated observations under weak supervision. This formulation enables a practical trade-off between annotation cost and performance, as shown in Tab. 5 (main paper, details below). Moreover, the word \textit{opportunistic} here is key: as shown with the cases of USGS or IGN, national geographic institutes have often already captured such data for their processes. In turn, our weak temporal supervision framework increases its value. While our proposed method is a combination of existing techniques (sIoU, batch sampling, iterative training), we hope it encourage the development of more sophisticated weak temporal supervision methods in the future.

\paragraph{Equal data budget evaluation:} We evaluate methods under an equal data budget (Tab.~\ref{tab:data_budget_eval}). Via reference costs for image acquisition ($Ac$), annotation ($An$) and generation ($Gen$), we estimate the budget of single-temporal methods (post-classification change detection and STAR) on Hi-UCD mini (i.e., acquisition and annotation at $t_1$) to be \$1214. We then derive the number of images other approaches (Changen2 pre-training, fully supervised change detection, and our method) can use under the same budget, accounting for $t_2$ acquisition, annotation, and generation costs. We use publicly available estimates for aerial acquisition~\cite{nc_orthoimagery} (\$100/km$^2$), annotation~\cite{mindkosh_estimator} (\$3/image$\approx$\$287/km$^2$), and generative model pricing~\cite{gemini_pricing} (\$10/km$^2$). The budget of a method per km$^2$ is given by $100(Ac_{t_1} + Ac_{t_2})+287(An_{t_1} + An_{t_2})+10Gen$. Under this fixed data budget, our method achieves the best performance. 

\begin{table}[t]
\centering
\scriptsize
\setlength{\tabcolsep}{3pt}
\resizebox{\linewidth}{!}{
\begin{tabular}{lcccccccccc}
\toprule
Method & \% & \# Im. & km$^2$ & Ac$_{t1}$ &  An$_{t1}$ & Ac$_{t2}$ &  An$_{t2}$ & Gen & Ch. IoU & Seg. mIoU \\
\midrule
Post-classif. & 100 & 300 & 3.14 & \cmark & \cmark & \xmark & \xmark & \xmark & 14.4 & 37.7 \\
STAR & 100 & 300 & 3.14 & \cmark & \cmark & \xmark & \xmark & \xmark & 45.1 & 26.3 \\
Changen2 & 90 & 270 & 2.92 & \cmark & \cmark & \xmark & \xmark & \cmark & 22.7 & 35.2 \\
\rowcolor{blue!10} Ours & 80 & 240 & 2.50 & \cmark & \cmark & \cmark & \xmark & \xmark & \textbf{47.2} & \textbf{40.0} \\
Sup. CD & 50 & 150 & 1.57 & \cmark & \cmark & \cmark & \cmark & \xmark & 44.0 & 34.9 \\
\bottomrule
\end{tabular}
}
\caption{\textbf{Evaluation under an equal data budget on Hi-UCD mini.} Reference budget is acquisition + annotation at $t_1$ (i.e. Post-classif. and STAR).
Synth adds cost of $t_2$ image generation (Gen). 
Ours adds budget to acquire $t_2$, while supervised CD annotates $t_2$.
\vspace{-1em}
}
\label{tab:data_budget_eval}
\end{table}

\setcounter{section}{0}
\setcounter{table}{0}
\renewcommand{\thetable}{B\arabic{table}}
\setcounter{equation}{0}
\renewcommand{\theequation}{B\arabic{equation}}
\setcounter{figure}{0}
\renewcommand{\thefigure}{B\arabic{figure}}
\setcounter{figure}{0}
\renewcommand{\thefigure}{B\arabic{figure}}
\renewcommand{\thesection}{B}
\begin{figure*}
\centering   
    \setlength{\tabcolsep}{1pt}
    \scriptsize
    \resizebox{1\linewidth}{!}{
    \begin{tabular}{ccccccccc}
    \multicolumn{6}{c}{\includegraphics[height=5pt]{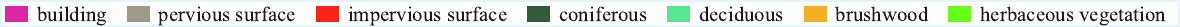}} & \multicolumn{3}{c}{\includegraphics[height=5pt]{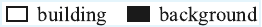}}\\
    \sampletriplet{b-FLAIR}{19} &
    \sampletriplet{b-FLAIR-spot}{028966} &
    \sampletriplet{b-IAILD}{kitsap20_85}\\
    \sampletriplet{b-FLAIR}{27} &
    \sampletriplet{b-FLAIR-spot}{041359} &
    \sampletriplet{b-IAILD}{chicago6_55}\\
    \sampletriplet{b-FLAIR}{38} &
    \sampletriplet{b-FLAIR-spot}{058483} &
    \sampletriplet{b-IAILD}{vienna33_33}\\  
    \sampletriplet{b-FLAIR}{056028} &
    \sampletriplet{b-FLAIR-spot}{056028} &
    \sampletriplet{b-IAILD}{tyrol-w12_0} \\
    \multicolumn{3}{c}{(a) b-FLAIR} &
    \multicolumn{3}{c}{(b) b-FLAIR-spot} & 
    \multicolumn{3}{c}{(c) b-IAILD} \\
    \end{tabular}
    }
\caption{\textbf{Additional example triplets from our extended datasets}, b-FLAIR, b-FLAIR-spot, and b-IAILD, to complement the visual examples already displayed in Fig. 2 of the main paper. For each dataset, we show triplets ($S_{t}$, $I_{t}$, $I_{t'}$) in this order from left to right.}
\label{fig:add_visual_examples}
\end{figure*}

\section{Data}\label{sec:suppmat_data}

Fig.~\ref{fig:add_visual_examples} shows further examples of triplets of bi-temporal images and their corresponding single-temporal semantic masks, for our three extended datasets, b-FLAIR, b-FLAIR-spot and b-IAILD. We describe them below, and also give a brief description of the benchmark datasets we used (LEVIR-CD, WHU-CD and S2Looking).

\begin{figure*}[t]
    \centering
    \includegraphics[width=1\linewidth,trim={0 0.1cm 0 0}, clip]{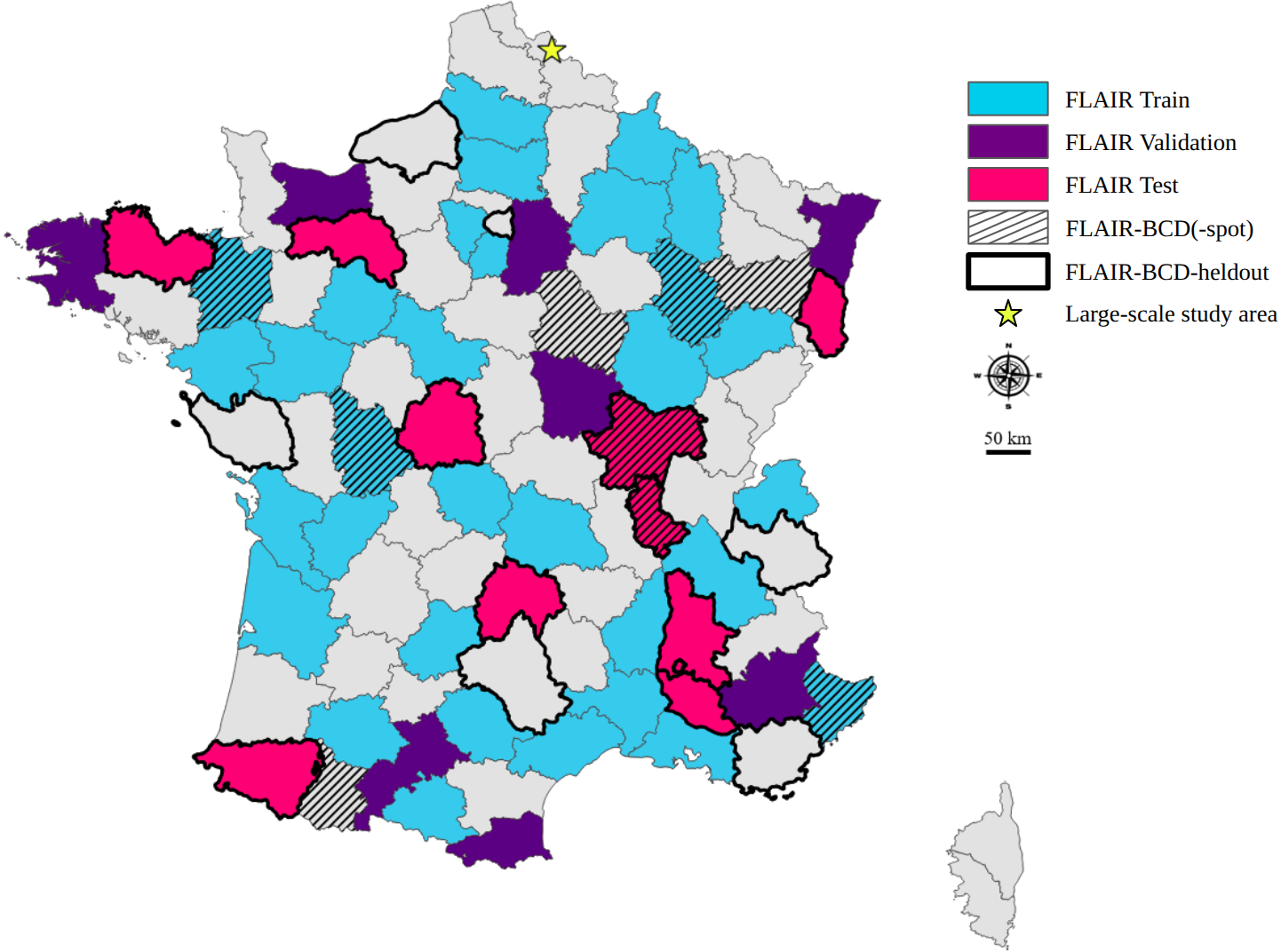}
    \caption{\textbf{Spatial repartition of FLAIR-BCD(-spot/-heldout) data w.r.t. FLAIR}, within Metropolitan France. Note that even if pairs from FLAIR-BCD(-spot) are from departments already covered by FLAIR Train/Val, we made sure that there is no intersection between the two datasets. We also indicate with a \protect\contour{black}{\color{yellow}\Large$\star$} the location of the large-scale zone used for Fig. 1 of the main paper.}
    \label{fig:data_splits}
\end{figure*}

\subsection{b-FLAIR}

The original FLAIR dataset is based on the French IGN's BD ORTHO product~\cite{ign2025bdortho} and its associated elevation data. For details on these data products, their pre-processing and their formatting into the FLAIR dataset, please refer to the FLAIR datapaper~\cite{garioud2022flair}. The BD ORTHO product is updated every three years on average at each location with new aerial acquisitions. This allowed us to extend FLAIR dataset with a new temporal acquisition for each of its 77,762 patches. For each patch, we extracted a 5-band (RGB, infrared, elevation) patch from the closest temporal acquisition of the BD ORTHO product and its associated elevation data, following the same process as described in~\cite{garioud2022flair}. We use the same training splits as in the original paper~\cite{garioud2023flair} (see Fig.~\ref{fig:data_splits}), except for the hyperparameter studies for which a subset of the data is used (departments 13, 21, 44, 63, 80 for training, and 77 for validation). The average number of days between two acquisitions is 1097 days ($\thicksim$3 years). 82.9\% of our added images correspond to images acquired after the corresponding original FLAIR patches, while the remaining were acquired before. The average number of days between two acquisitions within the same calendar year is 44 ($\thicksim$1.5 months), with 18\% of the pairs having more than a 3-month difference within the same calendar year, indicating possible significant seasonal variations for a given image pair. In addition, the differences in the time of the acquisitions within the day---which is 153 min on average---and in the angle of acquisition, introduce significant radiometric, geometric, and shadow-related variations.

\begin{figure*}[t]
    \centering
    \scriptsize
    \vspace{-1.5em}
    \begin{tabular}{cc}
        \includegraphics[width=0.45\linewidth]{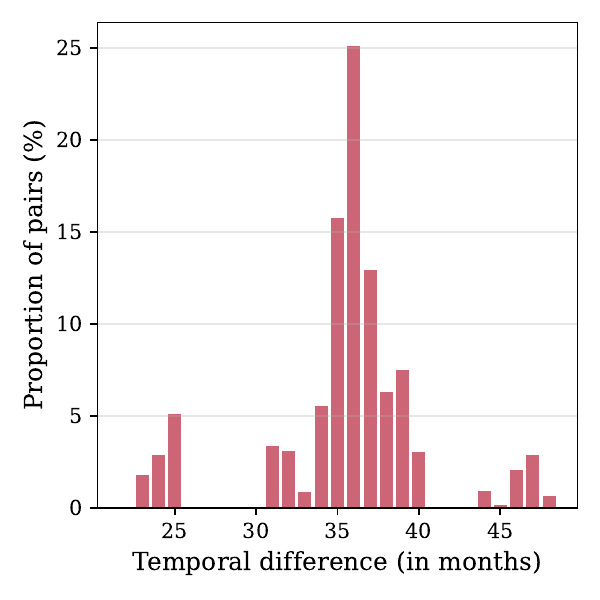}    &  
        \includegraphics[width=0.45\linewidth]{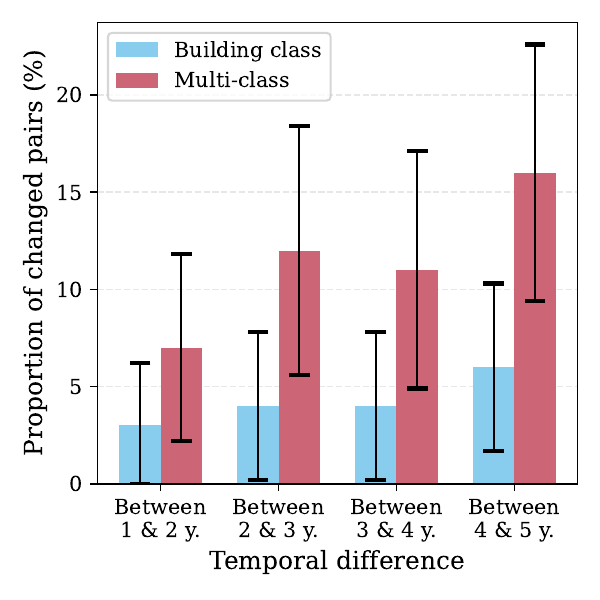}\vspace{-1.6em}
    \end{tabular}
    \caption{Temporal difference for b-FLAIR pairs (left) and estimated variation of the change rate with temporal distance (right).}
    \label{fig:month_diff_flair_distribution}
\end{figure*}

\begin{figure}[t]
    \centering
    \includegraphics[width=0.85\linewidth]{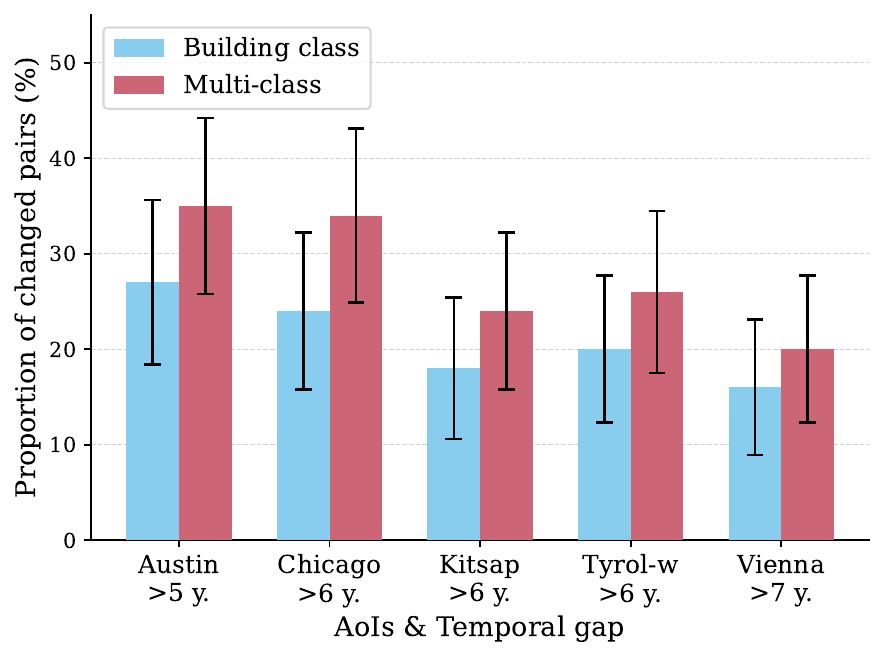}\vspace{-1.2em}
    \caption{Estimated variation of the change rate per AoIs/temporal distance on b-IAILD.}
    \label{fig:change_rate_iaild}
\end{figure}

\paragraph{Estimation of change events in b-FLAIR.}
Fig.~\ref{fig:month_diff_flair_distribution} (left) depicts the distribution of temporal gaps (in months) between image pairs in b-FLAIR. As observed, most pairs are approximately three years apart. A smaller portion (about 5\%) corresponds to gaps around two years, and a slightly smaller fraction around four years. Intuitively, larger temporal gaps are more likely to contain land-cover changes.
Although the exact number of change events in b-FLAIR is unknown, we estimate the change rate through manual inspection. To this end, we randomly sample 100 image pairs from each of the following temporal intervals: 1–2 years, 2–3 years, 3–4 years, and 4–5 years. For each subset, we manually identify land-cover changes and report the estimated change rate with 95\% confidence intervals, for both (i) multi-class changes (based on the original FLAIR semantic classes) and (ii) building changes only. 
Fig.~\ref{fig:month_diff_flair_distribution} (right) presents the estimated change proportions. The results indicate a gradual increase in change probability with larger temporal gaps. The two central intervals (2–3 years and 3–4 years), which account for the majority of the dataset, exhibit approximately 4\% building changes and about 12\% multi-class changes. A similar figure is provided for the b-IAILD dataset in \ref{fig:change_rate_iaild}, where we see that the observed changes likelihood in real pairs is greater, mostly due to the larger temporal window between observations and the urban focus of the original IAILD dataset, but remains lower than \~30\%.

\subsection{b-FLAIR-spot}

For all patch location and acquisition date of b-FLAIR, we download a corresponding SPOT-6/7 patch from the ORTHO-SAT database~\cite{ign2025orthosat}. These are RGB images in 8 bits, at a spatial resolution of 1.5 m/px. We resized the patches at the shape 64$\times$64, resulting in an effective spatial resolution of 1.6 m/px. The ORTHO-SAT is an annual product, and we align the acquisition year of the images of b-FLAIR-spot on the corresponding images in b-FLAIR. This allows us to reuse FLAIR annotations for b-FLAIR-spot, provided a nearest-neighbor resampling from 512$\times$512 to 64$\times$64.

\subsection{b-IAILD}

The original IAILD~\cite{maggiori2017dataset} public training set is composed of images from 5 cities from the USA (Austin, Chicago, Kitsap) and Austria (Tyrol, Vienna). For each city, the dataset contains 36 tiles of size 5000$\times$5000 px at the spatial resolution of 30 cm/px. We downloaded new acquisitions over the same locations from the USGS website\footnote{\url{https://earthexplorer.usgs.gov/}} and the respective website of the Austrian provinces of Tyrol\footnote{\url{https://data-tiris.opendata.arcgis.com/}} and Vienna\footnote{\url{https://www.wien.gv.at/ma41datenviewer/public/start.aspx}}. Years of acquisition of the new added orthophotographies are the following : 2024 for Vienna, 2023 for Chicago, Kitsap and Tyrol, and 2022 for Austin. Because the spatial resolution varies across the areas of interest (15 cm/px for Vienna, 20 cm/px for Tyrol, and 60 cm/px for Austin, Chicago and Kitsap), we resample all the downloaded tiles---as well as the original IAILD tiles---at the common resolution of 60 cm/px. As in~\cite{maggiori2017dataset}, we keep the first 5 tiles of each zone for validation, and use the other 31 for training. The resulting 2500$\times$2500 images are split in 100 patches of size 256$\times$256, over a grid with an overlap of 6 pixels between adjacent cells. This results in a dataset containing 15500 training pairs (3100 per area) and 2500 validation pairs (500 per area). 

\begin{figure*}[t]
    \centering
    \setlength{\tabcolsep}{1pt}
    \resizebox{\linewidth}{!}{
    \begin{tabular}{cccc@{~~~~}cccc}\\
    \flaircd{t1}
    \flaircd{t2}
    \flaircd{annot}
    \multicolumn{4}{c}{(a) Samples from FLAIR-BCD} & \multicolumn{4}{c}{(b) Samples from FLAIR-BCD-spot} \\    
    \flaircdtwo{t1}
    \flaircdtwo{t2}
    \flaircdtwo{annot}
    \multicolumn{4}{c}{(c) Samples from FLAIR-BCD-heldout (same res.)} & \multicolumn{4}{c}{(d) Samples from FLAIR-BCD-heldout (diff. res)} \\
    \end{tabular}
    }
    \caption{\textbf{Example triplets of our test sets for building change detection.} We show additional example triplets ($I_t$, $I_{t'}$, $M$) in this order from top to bottom, for FLAIR-BCD (a), FLAIR-BCD-spot (b), and FLAIR-BCD-heldout (for (c) same res. and (d) different res. pairs).}
    \label{fig:b_flair_test}
\end{figure*}
\renewcommand{\thefigure}{C\arabic{figure}}
\setcounter{figure}{0}

\subsection{FLAIR-BCD \& FLAIR-BCD-spot}

FLAIR-BCD is composed of 1730 image pairs annotated with a binary building change mask. The images are in the same 5-band 512$\times$512 format as FLAIR images and where acquired, processed, and formatted following the procedure described in~\cite{garioud2022flair}. The images were extracted from 9 different French administrative departments and do not intersect the FLAIR dataset, which allows for a sound evaluation of methods trained on FLAIR or on our bi-temporal extension of FLAIR (see Fig.~\ref{fig:data_splits}). The pairs were annotated by photointerpretation experts and verified by a non-expert assessor. The pairs either show new building constructions, or no building change at all ($\thicksim$30\% of the pairs), but no pair exhibits building destruction. FLAIR-BCD-spot is built from FLAIR-BCD similarly as how we obtained b-FLAIR-spot based on b-FLAIR, \textit{i.e.} by downloading acquisition at the same date and location for each patch, and downsampling the annotation masks. Example triplets for these two datasets can be visualized in Fig.~\ref{fig:b_flair_test}.

\begin{table*}[t]
\centering
\caption{\textbf{Overview of benchmark datasets.} We report the main characteristics of the 6 evaluation datasets on which we compare our approach to the baselines.}
\resizebox{\linewidth}{!}{
\begin{tabular}{llllll}
\toprule
Dataset & Modality & Resolution (cm/px) & Location & Image Size & Num. test pairs \\
\midrule
LEVIR-CD~\cite{chen2020spatial} & Aerial & 50 & Texas, US & $1024\times1024$ & 128 \\
WHU-CD~\cite{ji2018fully} & Aerial & 7.5 & Christchurch, New Zealand & $256\times256$ & 762 \\
S2Looking~\cite{shen2021s2looking} & Satellite & 50-80 & Global & $1024\times1024$ & 1000 \\
Hi-UCD mini~\cite{hiucd} & Aerial & 10 & Estonia & $1024\times1024$ & 386 \\
\rowcolor{blue!10}FLAIR-BCD (ours) & Aerial & 20 & France & $512\times512$ & 1730 \\
\rowcolor{blue!10}FLAIR-BCD-spot (ours) & Satellite & 160 & France & $64 \times 64$ & 1730 \\
\rowcolor{blue!10}FLAIR-BCD-heldout (ours) & Aerial & 20-50 & France & $512 \times 512$ & 400 \\
\bottomrule
\end{tabular}
}
\label{tab:datasets}
\end{table*}

\subsection{FLAIR-BCD-heldout}

FLAIR-BCD-heldout is composed of 400 image pairs annotated with binary building change masks. This dataset follows the same construction process as FLAIR-BCD. FLAIR-BCD-heldout spans the French départements 12, 15, 22, 26, 36, 61, 64, 68, 69, 71, 73, 76, 83, 84, 85, 92, 93, 94 and therefore does not overlap with b-FLAIR train and validation splits. We also made sure there was no spatial overlap between FLAIR-BCD-heldout and FLAIR-BCD. A key difference between the two being that, where b-FLAIR contains orthophotographies acquired between 2018 and 2023, FLAIR-BCD-heldout spans the years 2011 to 2024. As a result, some pairs mix recent and older generation aerial images, introducing resolution  mismatches (0.2cm/px VS 0.5cm/px) for 31.8\% of the pairs. All 0.5cm/px images are upsampled to the size 512$\times$512. No image pair exhibits building destruction and 33.3\% of the pairs exhibits no change at all. We show example FLAIR-BCD-heldout images in Fig.~\ref{fig:b_flair_test}. Because our hyperparameter choices are made w.r.t. the performance of our method on FLAIR-BCD, results on FLAIR-BCD-heldout are to be seen as an evaluation on untouched, held out data.

\subsection{LEVIR-CD, WHU-CD, S2Looking and Hi-UCD mini}

\paragraph{LEVIR-CD~\cite{chen2020spatial}} is composed of 637 pairs of aerial images of size 1024$\times$1024 at the spatial resolution of 50 cm/px. We keep the original data splits (445 images for training, 64 for validation, and 128 for testing). LEVIR-CD's images are from 20 different regions in the state of Texas, US, and were acquired between 2002 and 2018. The temporal gap for two acquisitions of the same location in the dataset range from less than a year to 15 years.

\paragraph{WHU-CD~\cite{ji2018fully}} is composed of 7620 pairs of aerial images of size 256$\times$256 at the spatial resolution of 7.5 cm/px. We keep the original data splits (6096 images for training, 762 for validation, and 762 for testing). WHU-CD's images are of the area of Christchurch, New Zealand, and were acquired in 2012 (pre-change) and 2016 (post-change). 

\paragraph{S2Looking~\cite{shen2021s2looking}} is composed of 5000 pairs of satellite images of size 1024$\times$1024 at a spatial resolution between 50 cm/px and 80 cm/px. We keep the original data splits (3500 images for training, 500 for validation, and 1000 for testing). S2Looking's images are from 15 areas of interest spread across Europe, Asia, Africa, and North and South America. They were acquired between 2017 and 2020, with a temporal gap ranging from 1 to 3 years between bi-temporal acquisitions.

\paragraph{Hi-UCD mini~\cite{hiucd}} is composed of 745 pairs of aerial images of size 1024$\times$1024 at the resolution of 10 cm/px. We keep the original data splits (300 images for training, 59 for validation and 386 for testing). Hi-UCD mini's images are from Estonia and were acquired between 2017 and 2019. Each image is annotated with a 9-class semantic mask (water, grassland, woodland, bare land, building, greenhouse, road, bridge, other). We also evaluate on Hi-UCD mini's test set methods trained on Hi-UCD's training set~\cite{tian2022large}, which contain 12,000 pairs of size 512$\times$512.\\

Tab.~\ref{tab:datasets} summarizes the main characteristics of these four datasets, as well as those of our three datasets, FLAIR-BCD, FLAIR-BCD-spot, and FLAIR-BCD-heldout

\setcounter{table}{0}
\renewcommand{\thetable}{C\arabic{table}}
\setcounter{equation}{0}
\renewcommand{\theequation}{C\arabic{equation}}
\setcounter{figure}{0}
\renewcommand{\thefigure}{C\arabic{figure}}
\setcounter{figure}{0}
\renewcommand{\thefigure}{C\arabic{figure}}
\renewcommand{\thesection}{C}
\section{Implementation details}\label{sec:suppmat_implt_details}
\begin{figure*}[ht]
    \centering
    \includegraphics[width=\linewidth,trim={0.5cm 5.25cm 2.5cm 0}, clip]{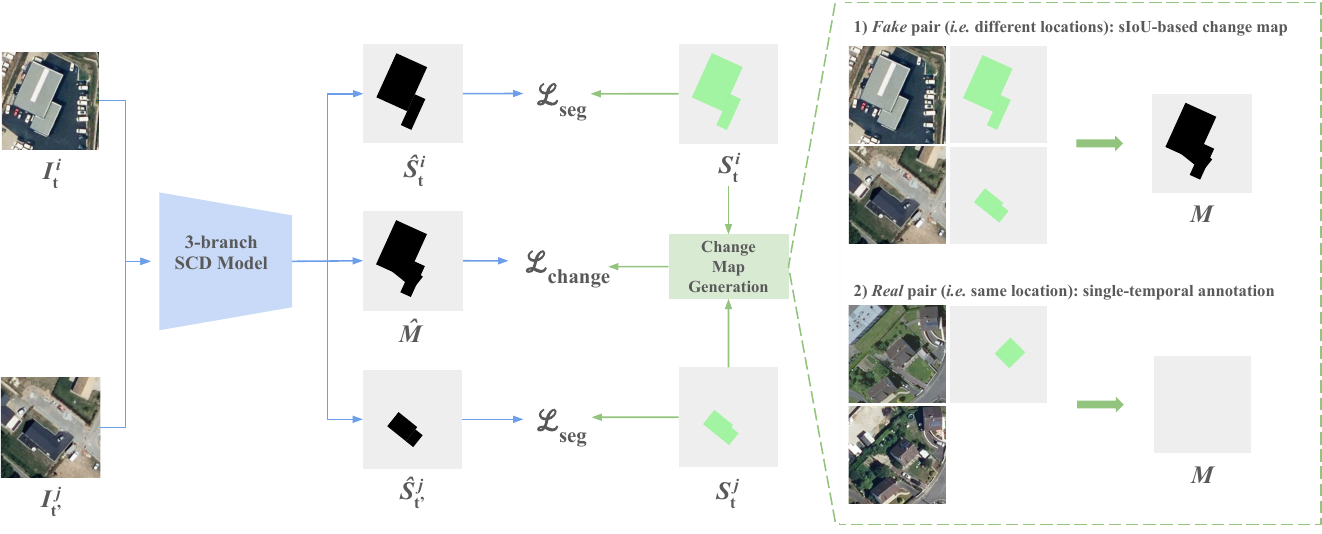}
    \caption{\textbf{General diagram of the proposed methodology.} Our methodology can be applied to any 3-branch SCD model, provided it can take two images at different dates as input and produce a semantic segmentation mask for each date, and a change mask. The semantic segmentation task is supervised through the single-temporal available ground truth, implying a temporal augmentation on the additional non-annotated image. The change detection task is supervised via weak temporal augmentation, creating \textit{fake} change examples by pairing images of different locations, and no-change examples by assuming no changes occurred between \textit{real} pairs.}
    \label{fig:general_diagram}
\end{figure*}

A general diagram of the proposed weak temporal supervision approach is shown in Fig.~\ref{fig:general_diagram}. We trained the b-IAILD and b-FLAIR-spot models on a single NVIDIA RTX 6000 GPU with batch sizes of 64 and 256, respectively. Training the b-IAILD model required approximately 26 hours, while b-FLAIR-spot completed in about 5.5 hours. The b-FLAIR dataset contains a significantly larger number of pixels than the other datasets, resulting in substantially higher computational complexity. Consequently, we trained the b-FLAIR models on a multi-GPU setup with 32 NVIDIA V100 GPUs and a batch size of 32, where a complete training took roughly 4 hours. For all models, we use image crops of $256 \times 256$ pixels (except b-FLAIR-spot, for which images are of size $64 \times 64$) and normalize input images using the per-channel mean and standard deviation computed over the entire training dataset, following Benidir \textit{et al.}~\cite{benidir2025change}. During inference and evaluation, target images are normalized using the same statistics as the training set. 

In Fig.~\ref{fig:example_change_maps}, we show example of change maps generated with our implemented sIoU-based methods. We visually compare them to change maps computed with logical operations (OR and XOR). While XOR-based masks leave residual change pixels around almost-overlapping buildings, our masks ignore such pixels, though they may correspond to an actual difference in land cover. We believe these ``cleaner'', object-based change masks constitute a better supervision in order to train models that are less prone to false alarms. 

\begin{figure*}[ht]
    \centering
    \setlength{\tabcolsep}{1pt}
    \resizebox{\linewidth}{!}{
    \begin{tabular}{ccccccccc}
    \xorsiousample{003357_013156}
    \xorsiousample{046186_011460}
    \xorsiousample{058357_029042}
    \xorsiousample{060091_002942}
    Image $I_{t_i}^i$ & Image $I_{t'_j}^j$ & Ground & Ground & XOR & sIoU & sIoU & sIoU & OR\\
     & & truth i & truth j &  & ($\tau=0.25$) & ($\tau=0.50$) & ($\tau=0.75$) & 
    \end{tabular}
    }
    \caption{\textbf{Example of generated change maps} for pairs of images of b-FLAIR from different locations, with a focus on the ``building class''. We show maps resulting from logical operations (XOR or OR) as well as sIoU-based maps computed for various values of the threshold $\tau$. We highlight with {\color{red}red circles}, areas for which differences in building footprints between the two images are variably considered as changed or not depending on the method.}    
    \label{fig:example_change_maps}
\end{figure*}

\begin{figure*}[t]
    \centering
    \setlength{\tabcolsep}{1pt}
    \resizebox{\linewidth}{!}{
    \begin{tabular}{cccccc}\\
    \changen{img1}
    \changen{label1}
    \changen{img2}
    \changen{label2}
    \multicolumn{6}{c}{\includegraphics[width=0.97\linewidth]{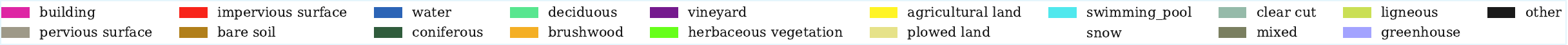}}\vspace{-1em}
    \end{tabular}
    }
    \caption{\textbf{Example synthetic change pairs generated with our implementation of Changen2 on FLAIR.} We show FLAIR images, their corresponding semantic mask, our generated change images, and the new semantic mask, in this order from top to bottom.}
    \label{fig:changen2sample}
\end{figure*}

\paragraph{Changen2 implementation.} In order to compare Changen2-based synthetic data pre-training~\cite{zheng2024changen2} with other methods on an equal footing, we generate our own synthetic dataset using their approach on FLAIR data. We train their RSDiT diffusion architecture on FLAIR for 100 epochs and use the resulting model to generate a synthetic bitemporal semantic change detection dataset. Following~\cite{benidir2025change}, we generate 3 images for each of the 60,000 first images in FLAIR's training set, resulting in 180,000 change pairs in total. Inspired by the naming conventions in~\cite{zheng2024changen2}, we name the resulting dataset Changen2-S19-180k, where the `S19' refers to the number of semantic classes in the dataset (i.e.\ FLAIR classes). We follow the masked change diffusion inference described in~\cite{zheng2024changen2}, with $T=50$ iterations and a pre-image guidance ratio of $\lambda=0.5$. We adapt their change event simulation tool to the class distribution of FLAIR. More specifically, we use the "Attribute Edit" rule-based function, where a segment is randomly assigned a new class when sampling a new semantic mask. We show in Fig.~\ref{fig:changen2sample} some examples of the images we generated. We also report in Tab.~\ref{tab:imagequality} the FID~\cite{heusel2017gans} and the IS~\cite{salimans2016improved} of the generated images on the validation set. Both quantitative and qualitative results are on par with those of the Changen2 datasets reported in~\cite{zheng2024changen2}, confirming the validity of this FLAIR-based synthetic dataset.

\begin{table}[t]
\centering
\caption{\textbf{Image quality evaluation.} We report image quality metrics for Changen2-based synthetic datasets (including ours). $^\dagger$Results reported in~\cite{zheng2024changen2}.}
\begin{tabular}{lll}
\toprule
Dataset & FID $\downarrow$ & IS $\uparrow$ \\
\midrule
Changen-90k~\cite{zheng2023scalable} & 34.74$^\dagger$ & 5.41$^\dagger$ \\
Changen2-S1-15k~\cite{zheng2024changen2} & 32.44$^\dagger$ & 4.64$^\dagger$ \\
Changen2-S19-180k (Ours) & 35.69 & 5.69  \\
\bottomrule
\end{tabular}
\label{tab:imagequality}
\end{table}

\renewcommand{\thefigure}{C\arabic{figure}}
\setcounter{figure}{0}

\setcounter{table}{0}
\renewcommand{\thetable}{D\arabic{table}}
\setcounter{equation}{0}
\renewcommand{\theequation}{D\arabic{equation}}
\setcounter{figure}{0}
\renewcommand{\thefigure}{D\arabic{figure}}
\setcounter{figure}{2}
\renewcommand{\thefigure}{D\arabic{figure}}
\renewcommand{\thesection}{D}
\section{Additional Results}\label{sec:suppmat_additionnal_results}
This section provides additional quantitative and qualitative results that complement those of the main paper.

\subsection{Results on FLAIR-BCD} We provide in Table~\ref{tab:bFLAIRtest2_suppmat} the complete results on the FLAIR-BCD evaluation set, introduced in Section~\ref{sec:suppmat_data} and used to select the values for the hyperparameters $p_\text{real}$, $\tau$, and $N_\text{iter}$. We can observe that all methods achieve higher performance than on FLAIR-BCD-heldout, as the temporal, spatial and geographical distributions are different closer to the one of b-FLAIR, and the gap between FSC-180k and our method reduces. Nevertheless, the relative ranking of methods remains unchanged.

\begin{table}[t]
    \centering
    \caption{\textbf{FLAIR-BCD results} for methods trained on b-FLAIR. The FLAIR-BCD is introduced in Section~\ref{sec:suppmat_data}.}
    \resizebox{0.85\linewidth}{!}{
    \begin{tabular}{lccc}
        \toprule
        Method & $F1\uparrow$ & $IoU\uparrow$ & $FPR\downarrow$ \\
        \midrule
        Post-classif & 64.2 & 47.3 & 2.87\\ 
        Post-classif + Temp. aug. & 59.0 & 41.8 & 2.46\\ 
        Changen2-S19-180k pretraining & 58.5 & 41.4 & 3.63\\ 
        FSC-180k pretraining & \textbf{83.1} & \textbf{71.1} & \underline{0.73}\\
        STAR & 75.1 & 60.1 & 1.58 \\
        \rowcolor{blue!10} Ours & \underline{79.0} & \underline{65.3} & \textbf{0.35} \\
        \bottomrule
    \end{tabular}
    }
    \label{tab:bFLAIRtest2_suppmat}
\end{table}

\paragraph{False positive rate comparison.}Tab.~\ref{tab:fpr_and_connected_components} reports FPR, number of change connected components per image pair, and average component size for our predictions, ground truth, and those of Benidir \textit{et al.}~\cite{benidir2025change}.
Component-level evaluation treats small isolated pixels equally to large false detections.  While such small artifacts can be easily removed through simple post-processing, larger false detections are more challenging to handle. Following prior work~\cite{parks2008evaluation, postprocessing_evaluation, Bou2025}, we report results both before and after applying a 5×5 median filter on the change maps. Across both metrics, our model produces substantially fewer false positives, consistent with the qualitative large-scale results in Fig. 1, where our predictions closely align with ground truth while FSC-180k shows numerous false alarms.

\begin{table}[t]
    \caption{\textbf{Comparison between our method and a pretraining on FSC-180k~\cite{benidir2025change}} on FLAIR-BCD. We report the false positive rate (FPR) of both methods, as well as object-related indicators: average number of change objects per pair, and average size of change objects. An object is defined as a pixel-connected component in the binary building change map.}
    \vspace{-2mm}
    \centering
    \renewcommand{\arraystretch}{0.85}
    \scriptsize
    \resizebox{0.9\linewidth}{!}{
    \begin{tabular}{lcccc}
        \toprule
        & 5x5 & \multirow{1}{*}{FPR ($\downarrow$)} & Num. & Obj. size \\
        & median filter & & obj. (avg) & (avg, m²)\\
        \midrule
        Ground truth   & ---  & ---  & 1.7 & 187 \\
        \midrule
        FSC-180k pretraining~\cite{benidir2025change} & \xmark & 0.73 & 12.1 & 30  \\
        \rowcolor{blue!10}Ours (b-FLAIR)  & \xmark & 0.35 & 3.0 & 87 \\
        \midrule
        FSC-180k pretraining~\cite{benidir2025change} & \cmark & 0.71 & 3.6 & 98  \\
        \rowcolor{blue!10}Ours (b-FLAIR)   & \cmark    & 0.28 & 1.8 & 145 \\
        \bottomrule
    \end{tabular}
    }
    \vspace{-.9em}
    \label{tab:fpr_and_connected_components}
\end{table}

\subsection{Ablation on filtering threshold}
\begin{figure*}[t]
    \centering
    \includegraphics[width=\linewidth]{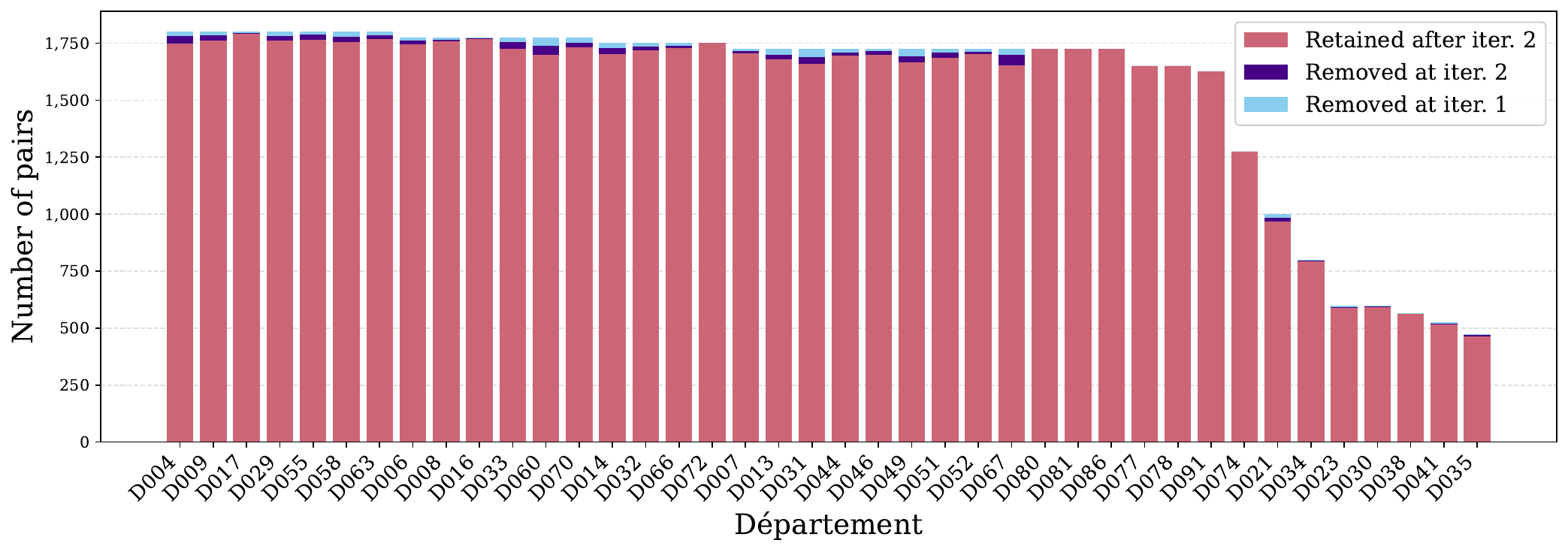}\vspace{-1.2em}
    \caption{Geographic distribution of bitemporal pairs in b-FLAIR and filtering across refinement iterations.}
    \label{fig:geo_distrib_per_iter}
\end{figure*}

During training, we filter potential change events using a pixel-change threshold of 2\%. Table~\ref{tab:change_filtering} reports an ablation study on this threshold for the third training iteration of the b-FLAIR model. As expected, increasing the threshold significantly reduces the number of filtered samples, since only pairs with a larger proportion of predicted change pixels are removed. With the selected 2\% threshold, 1.8\% of the training samples are discarded. This value remains well below the estimated building change rate reported in Fig.~\ref{fig:month_diff_flair_distribution}, while still removing a considerable number of noisy samples from the training set. Lastly, we provide information about the b-FLAIR filtered pairs during training: Tab.~\ref{tab:metadata_per_iter} provides the temporal statistics of remaining pairs before each iteration during training, and Figure~\ref{fig:geo_distrib_per_iter} reports the geographic distribution of pairs across refinement iterations. We observe that the refinement procedure does not introduce any significant bias in the training data, with the temporal and geographical distributions of the data remaining substantially the same.

\begin{table} [t]
    \centering
    \caption{\textbf{Temporal statistics.} Temporal statistics of remaining pairs before each training iteration.}
    \resizebox{0.8\linewidth}{!}{
        \begin{tabular}{lccc}
        \toprule
        Statistic
          & Iter.~1
          & Iter.~2
          & Iter.~3 \\
        \midrule
        Total bitemporal pairs              & 61,712 & 61,228 & 60,680\\
        \midrule
        \multicolumn{4}{l}{\textit{Acquisition-gap distribution (\%)}} \\
        \quad $< 24$ months                 & 2.2 & 2.3 & 2.3\\
        \quad $24$--$30$ months             & 7.0 & 7.0 & 7.0\\
        \quad $30$--$36$ months             & 35.1 & 35.1 & 35.1\\
        \quad $36$--$42$ months             & 47.3 & 47.2 & 47.2\\
        \quad $> 42$ months                 & 8.4 & 8.4 & 8.5\\
        Median gap (months)                 & 36.3 & 36.3 & 36.3\\
        Mean gap (months)                   & 36.1 & 36.1 & 36.1\\
        \bottomrule
        \end{tabular}
    }
    \label{tab:metadata_per_iter}
\end{table}

\begin{table}[t]
\vspace{0.8em}
\centering
\caption{\textbf{Filtering threshold.} Removed samples for different pixel change thresholds for the third training iteration of the b-FLAIR model.}
\label{tab:change_filtering}
\begin{tabular}{cccc}
\toprule
Threshold (\%) & Removed & Removed (\%) & Acc (\%) \\
\midrule
1  & 1121 & 2.39 & 70.1 \\
2  & 842  & 1.80 & 78.7 \\
5  & 478  & 1.02 & 89.7 \\
10 & 315  & 0.67 & 95.2 \\
20 & 161  & 0.34 & 96.9 \\
\bottomrule
\end{tabular}
\end{table}

\subsection{Detecting demolitions.} Our method naturally handles building demolitions via random pairing of \textit{fake} image pairs, learning building–background transitions in both temporal directions. We quantitatively evaluate this by inverting the temporal order of FLAIR-BCD input image to simulate demolition events. 
We report in Table~\ref{tab:inversed_flair_test} the results on this dataset variant for the DualUnet model, which we name \textit{FLAIR-BCD-inv}. As shown, the model achieves robust results on the demolition simulations as well.

\begin{table}[t]
\centering
\setlength{\tabcolsep}{4pt}
\caption{\textbf{Results on FLAIR-BCD-inv.} Inverted FLAIR-BCD input images to simulate building demolitions.}
\label{tab:inversed_flair_test}
\begin{tabular}{lcc}
\toprule
Dataset & F1 & IoU \\
\midrule
FLAIR-BCD & 79.0 & 65.3 \\
FLAIR-BCD-Inv & 81.2 & 68.3 \\
\bottomrule
\end{tabular}
\end{table}

\subsection{Large-scale results}
In Fig. 1 of the main paper, we present large-scale results over the metropolitan area of Lille, France (approximately 55.3\,km\textsuperscript{2}, see Fig.~\ref{fig:data_splits} for location on France map). This section provides additional details on the inference setup and extends the analysis to SPOT-6/7 satellite imagery.

\begin{figure*}[t]
    \centering
    \setlength{\tabcolsep}{1pt}
    \vspace{-0.8em}
    \includegraphics[width=\linewidth,trim={0cm 9.25cm 1.5cm 0cm}, clip]{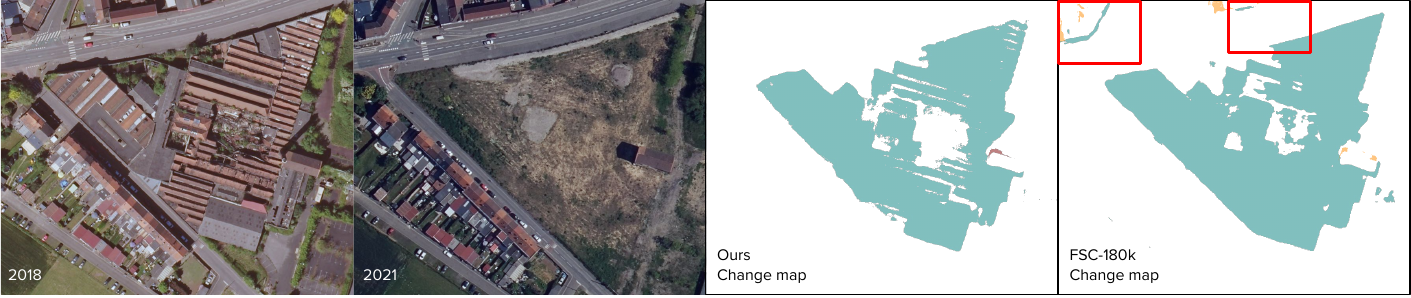}\vspace{-0.4em}
    \caption{\textbf{Large scale qualitative comparison of FSC-180k and our method for an example of the Lille metropolitan area}. Demolitions (blue) and constructions (orange) are shown in each case. As seen, FSC-180k predicts some false positives seemingly due to shadow changes and temporal variations.
    }    
    \label{fig:qualitative_fsc_FP}
\end{figure*}

\begin{figure*}[t]
    \centering
    \setlength{\tabcolsep}{1pt}
    \vspace{-0.8em}
    \begin{tabular}{wl{0.33\linewidth}wl{0.33\linewidth}l}
         Ground truth & Ours & Changen2~\cite{zheng2024changen2} \\
         \multicolumn{3}{c}{\includegraphics[width=\linewidth,trim={0cm 5.25cm 0cm 0cm}, clip]{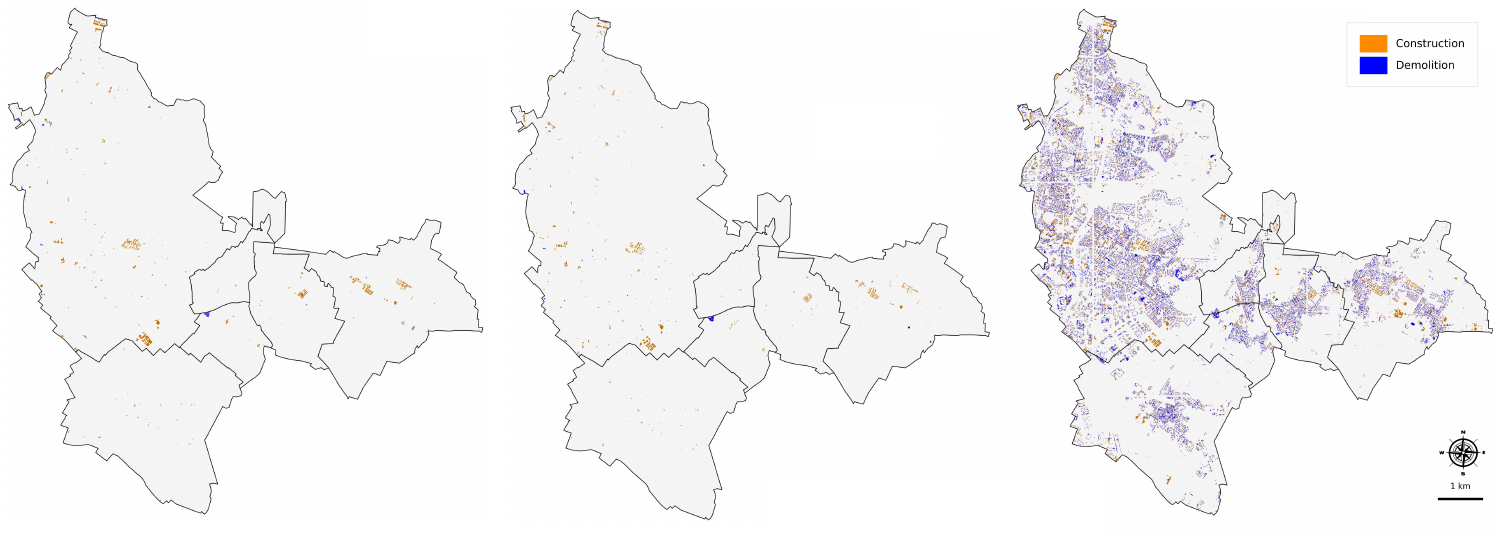}}
    \end{tabular}
    \vspace{-0.4em}
    \caption{\textbf{Large scale qualitative comparison of Changen2 and our method for the Lille metropolitan area}. Demolitions (blue) and constructions (orange) are shown in each case. Changen2 seems to predict actual demolitions and construction events but it also generates an overwhelming number of false positives.
    }    
    \label{fig:large_scale_changen2}
\end{figure*}

The large-scale inference on Lille uses BD ORTHO~\cite{ign2025bdortho} bi-temporal aerial images from 2018 and 2021. We extract $512 \times 512$ crops with a 6-pixel overlap and feed them to our model pre-trained on b-FLAIR, as well as to the FSC-180K model of Benidir \textit{et al.}~\cite{benidir2025change}. A $5 \times 5$ spatial median filter is applied to the model outputs. We provide in Figure~\ref{fig:qualitative_fsc_FP} a qualitative example of our large scale experiments over the Lille metropolitan area, comparing the results of our b-FLAIR method and FSC-180k. As observed, the latter predicts false positives linked to shadow changes and temporal variations. In addition, we show in Figure~\ref{fig:large_scale_changen2} the comparative large scale results between our method and Changen2 applied to FLAIR data. As seen, Changen2 generates a large amount of false positives, seemingly due to the lack of temporal diversity and light building misalignments.

\begin{figure*}[ht]
    \centering
    \begin{tabular}{wl{0.33\linewidth}wl{0.33\linewidth}l}
         Ground truth & Ours & STAR~\cite{zheng2021change} \\
         \multicolumn{3}{c}{\includegraphics[width=\linewidth,trim={0 4.75cm 0 0}, clip]{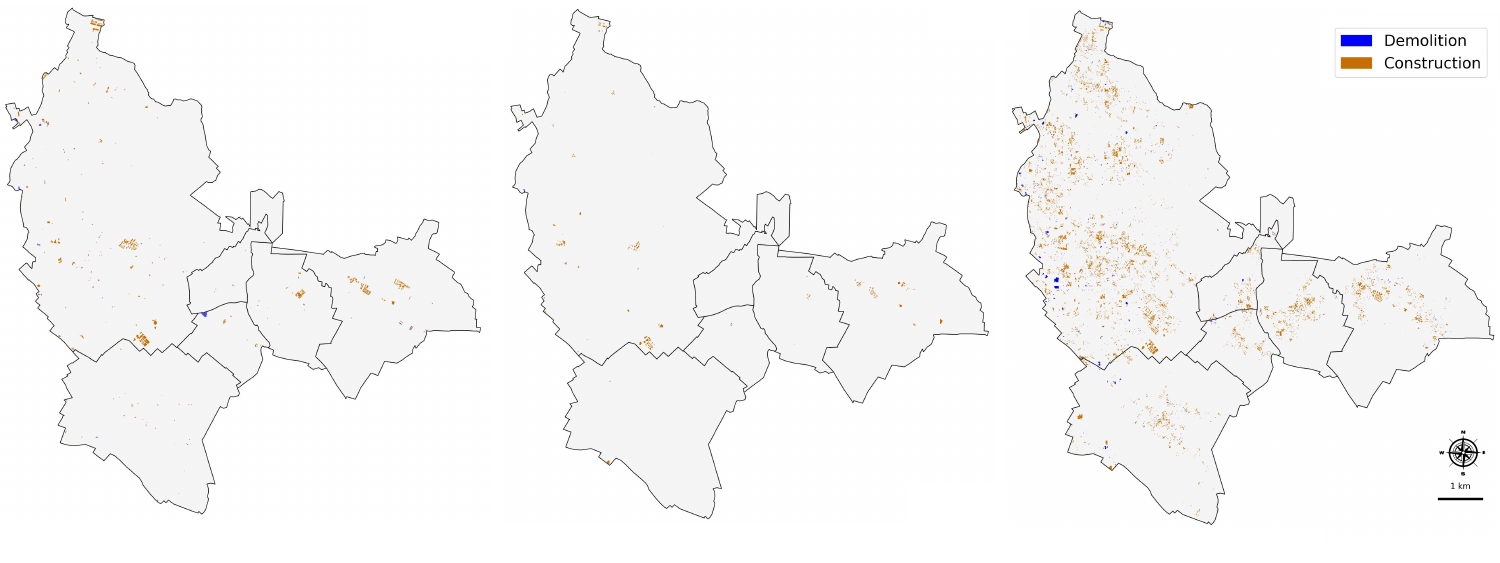}}
    \end{tabular}
    \caption{\textbf{Large-scale SPOT-6/7 change detection results.} Comparison between ground truth, our b-FLAIR-spot model, and the STAR approach over the metropolitan region region of Lille, France. While our model predicts a considerable number of false negatives, it detects most areas where larger changes were produced, while the STAR version detects an overwhelming amount of false positives.}
    \label{fig:spot_large_scale}
\end{figure*}

Motivated by the strong performance on very high-resolution aerial data, we also evaluate our method at scale on more widely accessible satellite imagery. We extract $64 \times 64$ SPOT-6/7 crops from IGN’s ORTHO-SAT database~\cite{ign2025orthosat} at 1.5\,m/pixel resolution with a 5\% overlap. We process these crops with the same inference pipeline and report results for both our b-FLAIR-spot model and its STAR counterpart in Fig.~\ref{fig:spot_large_scale}. As shown, the STAR model produces many false positives, similar to FSC-180K on the BD ORTHO experiment. In contrast, our model is substantially more robust to false alarms and successfully highlights most large change regions, although it remains more susceptible to false negatives.

\input{appendix/main_table_FPR}

\begin{table*}[t!]
  \renewcommand{\arraystretch}{0.92}
  \caption{\textbf{Performance across training iterations.} F1 Score (\%) and IoU (\%) of our method over three training iterations. Best score for each dataset within a training data source is highlighted in bold. The number of filtered pairs with respect to the training dataset is indicated for each iteration. $^\dagger$Scores for LEVIR-CD with FLAIR data source where computed with a crop size of 256 without resampling (see Sec.~\ref{sec:resultslevirana}).}
  \centering
  \resizebox{0.7\linewidth}{!}{
  \begin{tabular}{ccc *{10}{c}}
    \toprule
    \multirow{4}{*}{\rot[70]{Data source}} & \multirow{4}{*}{Iteration} & \multirow{4}{*}{\shortstack{Filtered\\samples}} &&&&&&&&&&\\
    & & & \multicolumn{4}{c}{In-domain} & \multicolumn{6}{c}{Out-of-Domain}\\
    \cmidrule(lr){4-7}\cmidrule(lr){8-13}
    & & & \multicolumn{2}{c}{FLAIR-BCD} & \multicolumn{2}{c}{FLAIR-BCD-spot} & \multicolumn{2}{c}{LEVIR-CD} & \multicolumn{2}{c}{WHU-CD} & \multicolumn{2}{c}{S2Looking}\\
    \cmidrule(lr){4-5}\cmidrule(lr){6-7}\cmidrule(lr){8-9}\cmidrule(lr){10-11}\cmidrule(lr){12-13}
    & & & $F1\uparrow$ & $IoU\uparrow$ & $F1\uparrow$ & $IoU\uparrow$ & $F1\uparrow$ & $IoU\uparrow$ & $F1\uparrow$ & $IoU\uparrow$ & $F1\uparrow$ & $IoU\uparrow$ \\
    \midrule
    \multirow{3}{*}{\rot[90]{\small{FLAIR}}}
    & 1 & 0   & 67.3 & 50.7 & --- & --- & 5.2$^\dagger$ & 2.7$^\dagger$ & 51.8 & 34.9 & 10.5 & 5.6 \\
    & 2 & 484 & 78.1 & 64.1 & --- & --- & \textbf{20.3}$^\dagger$ & \textbf{11.3}$^\dagger$ & 76.1 & 61.4 & \textbf{15.4} & \textbf{8.3} \\
    \rowcolor{blue!10} \global\let\CT@@do@color\relax & \global\let\CT@@do@color\oriCT@@do@color 3 & 842 & \textbf{79.0} & \textbf{65.2} & --- & --- & 17.8$^\dagger$ & 9.3$^\dagger$ & \textbf{77.3} & \textbf{63.0} & 13.6 & 7.3 \\
    \midrule
    \addlinespace[3mm]
    \multirow{3}{*}{\rot[90]{\small{FLAIR-spot}}}
    & 1 & 0   & --- & --- & 22.6 & 12.7 & 34.3 & 20.7 & 48.6 & 32.1 & 6.9 & 3.2 \\
    & 2 & 148 & --- & --- & 21.7 & 12.2 & \textbf{35.4} & \textbf{21.5} & 47.2 & 30.9 & 5.5 & 2.8 \\
    \rowcolor{blue!10} \global\let\CT@@do@color\relax & \global\let\CT@@do@color\oriCT@@do@color 3 & 287 & --- & --- & \textbf{22.9} & \textbf{12.9} & 34.1 & 20.6 & \textbf{49.3} & \textbf{32.7} & \textbf{7.1} & \textbf{3.7} \\
    \addlinespace[4mm]
    \midrule
    \multirow{3}{*}{\rot[90]{\small{IAILD}}}
    & 1 & 0   & --- & --- & --- & --- & \textbf{57.2} & \textbf{40.1} & 66.9 & 50.2 & 14.5 & 7.8 \\
    & 2 & 49  & --- & --- & --- & --- & 34.4 & 20.8 & \textbf{67.9} & \textbf{51.4} & 14.7 & 8.0 \\
    \rowcolor{blue!10} \global\let\CT@@do@color\relax & \global\let\CT@@do@color\oriCT@@do@color 3 & 41  & --- & --- & --- & --- & 35.9 & 21.9 & 63.3 & 46.3 & \textbf{17.6} & \textbf{9.6} \\
    \bottomrule
  \end{tabular}}
  \label{tab:iteration_comparison_bflairspot_full}
\end{table*}

\subsection{Additional qualitative results}
Tab.~\ref{tab:comparison_results_supp_mat} provides additional FPR results for all evaluated methods as an extension of the results presented in the main paper. Our method consistently leads to less false positive than baselines, though performing on par with them w.r.t other metrics on most datasets. This is especially the case on in-domain datasets and on WHU-CD. In Fig.~\ref{fig:qualitative_results_spot}, we can clearly see that, on FLAIR-BCD-spot, our method mostly predict no change, whereas its change predictions are accurate (see first row). On the contrary, the other methods show a significant number of false alarms, making them unusable in practical applications. This behavior can also be observed with the zero-shot qualitative results on LEVIR-CD (Fig.~\ref{fig:qualitative_levir}) and WHU-CD (Fig.~\ref{fig:qualitative_whu}). On S2Looking (Fig.~\ref{fig:qualitative_s2l}), all methods globally perform poorly due to the important domain gap with the training data. 

\subsection{Detailed impact of the iterative refinement}

Tab.~\ref{tab:iteration_comparison_bflairspot_full} reports the results for each iteration of our models trained on b-FLAIR, b-FLAIR-spot, and b-IAILD, respectively. Although we set the hyperparameter $N_\text{iter}=3$ based solely on in-domain dataset results, we also report the impact of iterative refinement on out-of-domain zero-shot performance. With the exception of the b-IAILD model on LEVIR-CD, where the first iteration significantly outperforms subsequent ones, removing detected change pairs from the dataset generally benefits out-of-domain performance. The number of filtered pairs at each iteration suggests that additional iterations could further improve model performance. However, given the marginal improvement between the second and third iterations on in-domain evaluation and considering computational cost constraints, we did not explore values of $N_\text{iter}$ greater than 3.

\begin{figure*}[ht]
    \centering
    \setlength{\tabcolsep}{1pt}
    \begin{tabular}{ccccccc}
    \qualitativespot{593}
    \qualitativespot{1062}
    \qualitativespot{1679}
    Image $I_t$ & Image $I_{t'}$ & Ground truth & Post-classif. & Post-classif. & STAR & Ours \\
    & & & & + temp. aug. & & \\
    \end{tabular}
    \caption{\textbf{Qualitative results on FLAIR-BCD-spot.} We compare the building change maps predicted by baseline methods and ours.}    
    \label{fig:qualitative_results_spot}
\end{figure*}

\begin{figure*}[ht]
    \centering
    \scriptsize
    \setlength{\tabcolsep}{1pt}
    \resizebox{\linewidth}{!}{
    \begin{tabular}{cccccccc}
    \qualitativeoutdomainone{levir}{test_33_768_256}
    \qualitativeoutdomainone{levir}{test_63_512_512}
    \qualitativeoutdomainone{levir}{test_86_768_0}
    Image $I_t$ & Image $I_{t'}$ & Ground truth & PC (FLAIR) & PC (FLAIR-spot) & PC (IAILD) & PC (b-FLAIR) & PC (b-FLAIR-spot)\\
    \qualitativeoutdomaintwo{levir}{test_33_768_256}    
    \qualitativeoutdomaintwo{levir}{test_63_512_512}
    \qualitativeoutdomaintwo{levir}{test_86_768_0}
    PC (b-IAILD) & FSC-180k p.t. & STAR (FLAIR) & STAR (FLAIR-spot) & STAR (IAILD) & Ours (b-FLAIR) & Ours (b-FLAIR-spot) & Ours (b-IAILD)
    \end{tabular}
    }
    \caption{\textbf{Qualitative zero-shot results on LEVIR-CD} for 3 randomly selected input pairs. ``PC'' stands for ``post-classification'' and ``p.t.'' for ``pre-training''.}  
    \label{fig:qualitative_levir}
\end{figure*}

\begin{figure*}[ht]
    \centering
    \scriptsize
    \setlength{\tabcolsep}{1pt}
    \resizebox{\linewidth}{!}{
    \begin{tabular}{cccccccc}
    \qualitativeoutdomainone{whu}{132}
    \qualitativeoutdomainone{whu}{369}
    \qualitativeoutdomainone{whu}{657}
    Image $I_t$ & Image $I_{t'}$ & Ground truth & PC (FLAIR) & PC (FLAIR-spot) & PC (IAILD) & PC (b-FLAIR) & PC (b-FLAIR-spot)\\
    \qualitativeoutdomaintwo{whu}{132}    
    \qualitativeoutdomaintwo{whu}{369}
    \qualitativeoutdomaintwo{whu}{657}
    PC (b-IAILD) & FSC-180k p.t. & STAR (FLAIR) & STAR (FLAIR-spot) & STAR (IAILD) & Ours (b-FLAIR) & Ours (b-FLAIR-spot) & Ours (b-IAILD)
    \end{tabular}
    }
    \caption{\textbf{Qualitative zero-shot results on WHU-CD} for 3 randomly selected input pairs. ``PC'' stands for ``post-classification'' and ``p.t.'' for ``pre-training''.}  
    \label{fig:qualitative_whu}
\end{figure*}

\begin{figure*}[ht]
    \centering
    \scriptsize
    \setlength{\tabcolsep}{1pt}
    \resizebox{\linewidth}{!}{
    \begin{tabular}{cccccccc}
    \qualitativeoutdomainone{s2l}{2_512_768}
    \qualitativeoutdomainone{s2l}{270_768_0}
    \qualitativeoutdomainone{s2l}{680_0_512}
    Image $I_t$ & Image $I_{t'}$ & Ground truth & PC (FLAIR) & PC (FLAIR-spot) & PC (IAILD) & PC (b-FLAIR) & PC (b-FLAIR-spot)\\
    \qualitativeoutdomaintwo{s2l}{2_512_768}    
    \qualitativeoutdomaintwo{s2l}{270_768_0}
    \qualitativeoutdomaintwo{s2l}{680_0_512}
    PC (b-IAILD) & FSC-180k p.t. & STAR (FLAIR) & STAR (FLAIR-spot) & STAR (IAILD) & Ours (b-FLAIR) & Ours (b-FLAIR-spot) & Ours (b-IAILD)
    \end{tabular}
    }
    \caption{\textbf{Qualitative zero-shot results on S2Looking} for 3 randomly selected input pairs. ``PC'' stands for ``post-classification'' and ``p.t.'' for ``pre-training''.}  
    \label{fig:qualitative_s2l}
\end{figure*}

\subsection{Results on LEVIR-CD}
\label{sec:resultslevirana}
Our quantitative results on out-of-domain datasets demonstrate that the proposed approach achieves competitive performance on WHU-CD and S2Looking. However, its performance on LEVIR-CD falls below that of baseline methods. The overall results on LEVIR-CD are, in fact, quite lower compared to those on WHU-CD. Through extensive analysis, we identify two key factors of this phenomenon.

\paragraph{Resolution.}
We observe that the LEVIR-CD dataset presents a significant distribution shift for models trained on FLAIR-based data, primarily due to resolution differences: the test resolution is less than half that of the training resolution. Addressing this, we find that applying a simple bilinear interpolation to double the image size before model inference substantially improves performance on this dataset. Tab.~\ref{fig:levir_tesults_multisize} illustrates the variability in LEVIR-CD results for models pre-trained on FLAIR-based datasets, evaluated across crop sizes of $256 \times 256$ and $512 \times 512$, and resampling image sizes of $256 \times 256$ and $512 \times 512$ (i.e., zooming in, zooming out, or maintaining the original crop size).

\paragraph{Temporal variation and dataset difficulty.}
Our results indicate that approaches employing temporal augmentation via the b-FLAIR dataset, i.e. our method and the post-classification approaches, yield lower performance on LEVIR-CD. Furthermore, our low-data regime experiments reveal that our b-FLAIR pre-trained model surpasses previous baselines with only 1\% of the training data. This suggests that LEVIR-CD benefits minimally from temporal augmentation, as the dataset exhibits limited temporal and appearance variability. Consequently, minimal fine-tuning is enough to adapt the pre-trained model effectively to LEVIR-CD.

\begin{table}[t]
    \centering
    \caption{\textbf{F1 scores on LEVIR-CD for different image sizes and resampling scales}. As seen, simple interpolation to zoom into the images drastically increases performance.}
    \begin{tabular}{lcccc}
    \toprule
     & \multicolumn{2}{c}{Crop size 256} & \multicolumn{2}{c}{Crop size 512} \\
    \cmidrule(lr){2-3} \cmidrule(lr){4-5}
    Resampling size & 256 & 512 & 256 & 512 \\
    \midrule
    Post-classif. & 37.3 & 51.4 & 10.4 & 37.4 \\
    Post-classif.+temp. aug. & 32.3 & 40.3 & 7.9 & 30.4 \\
    Changen2-S19-180k & 45.4 & 51.8 & 17.3 & 44.8 \\
    FSC-180k (our eval) & 57.0 & 65.3 & 11.2 & 56.3 \\
    STAR & 43.6 & 53.3 & 12.0 & 44.5 \\
    Ours & 17.8 & 44.7 & 4.1 & 25.2 \\
    \bottomrule
    \end{tabular}
    \label{fig:levir_tesults_multisize}
\end{table}

\subsection{Comparison with Segment Any Change}

\paragraph{Segment Any Change for class-agnostic changes.}
As discussed in the main text, Segment Any Change~\cite{chan2021segment} builds upon Segment Anything Model~\cite{sam} to detect class-agnostic changes. Consequently, the predicted masks often correspond to changes that are not associated with specific semantic categories such as buildings. In many cases, no clear semantic interpretation can be assigned to the detected regions. This is reflected by the authors in their paper, as they note that pixel-wise metrics such as F1 score and IoU are used only as a reference when evaluating on building change detection datasets (e.g., LEVIR-CD). For this reason, we exclude Segment Any Change from the quantitative comparison in the main text.

\begin{figure*}[t]
\centering
\scriptsize
\setlength{\tabcolsep}{2pt}

\resizebox{\linewidth}{!}{
\begin{tabular}{cccc}
Input t1 & Input t2 & Prediction & Binary Change Map \\
\includegraphics[width=0.22\textwidth]{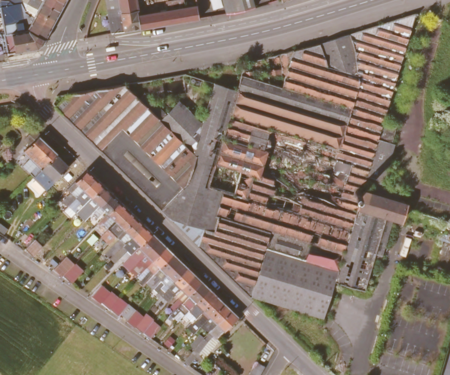} &
\includegraphics[width=0.22\textwidth]{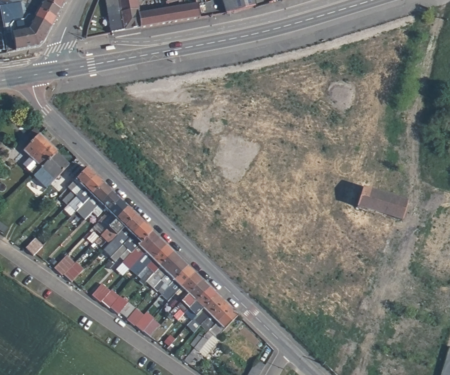} &
\includegraphics[width=0.22\textwidth]{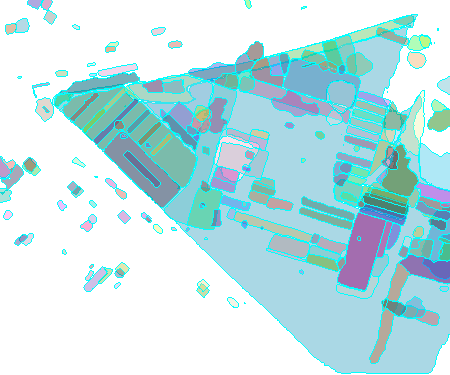} &
\includegraphics[width=0.22\textwidth]{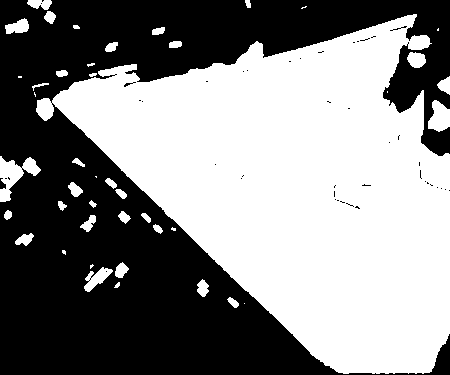} \\
\includegraphics[width=0.22\textwidth]{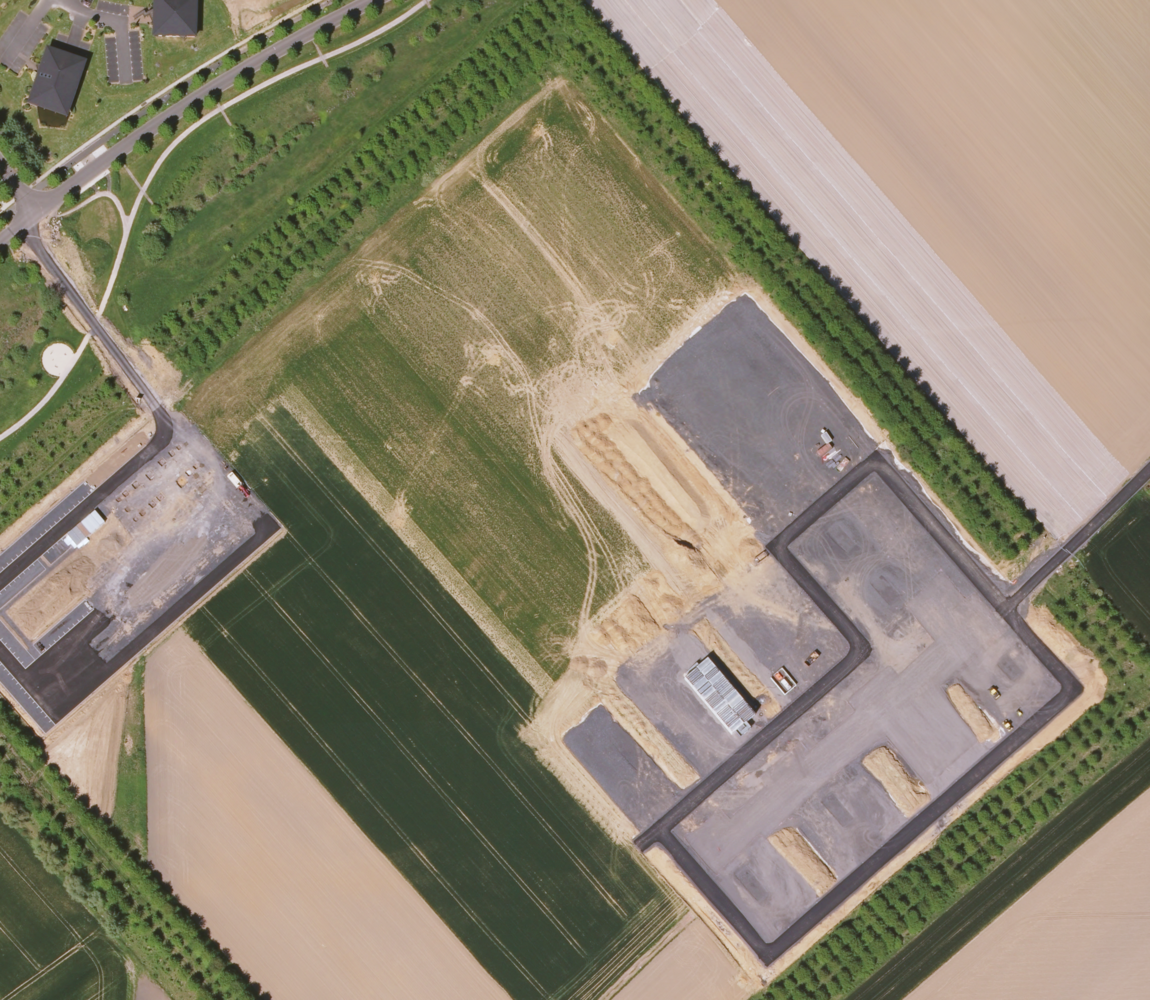} &
\includegraphics[width=0.22\textwidth]{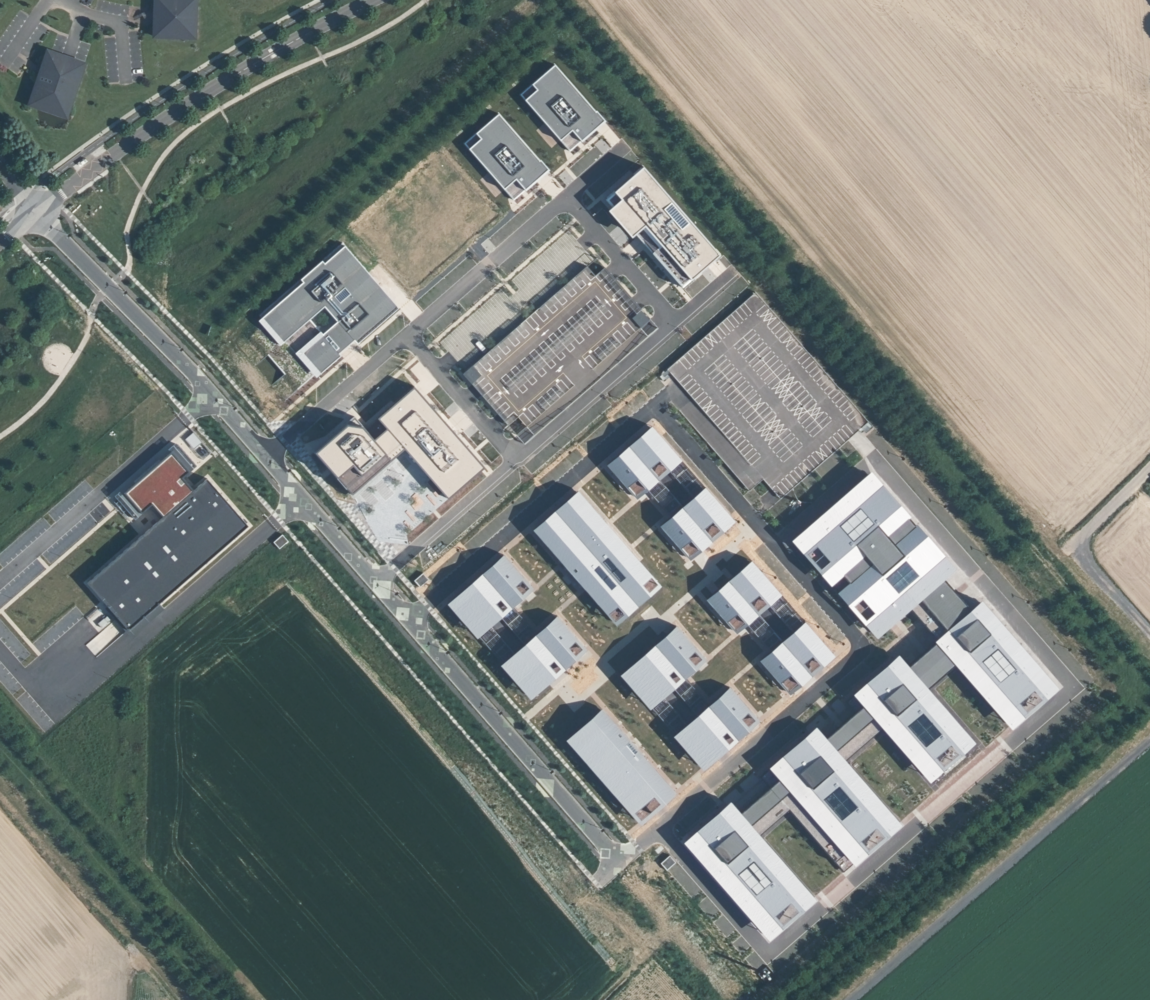} &
\includegraphics[width=0.22\textwidth]{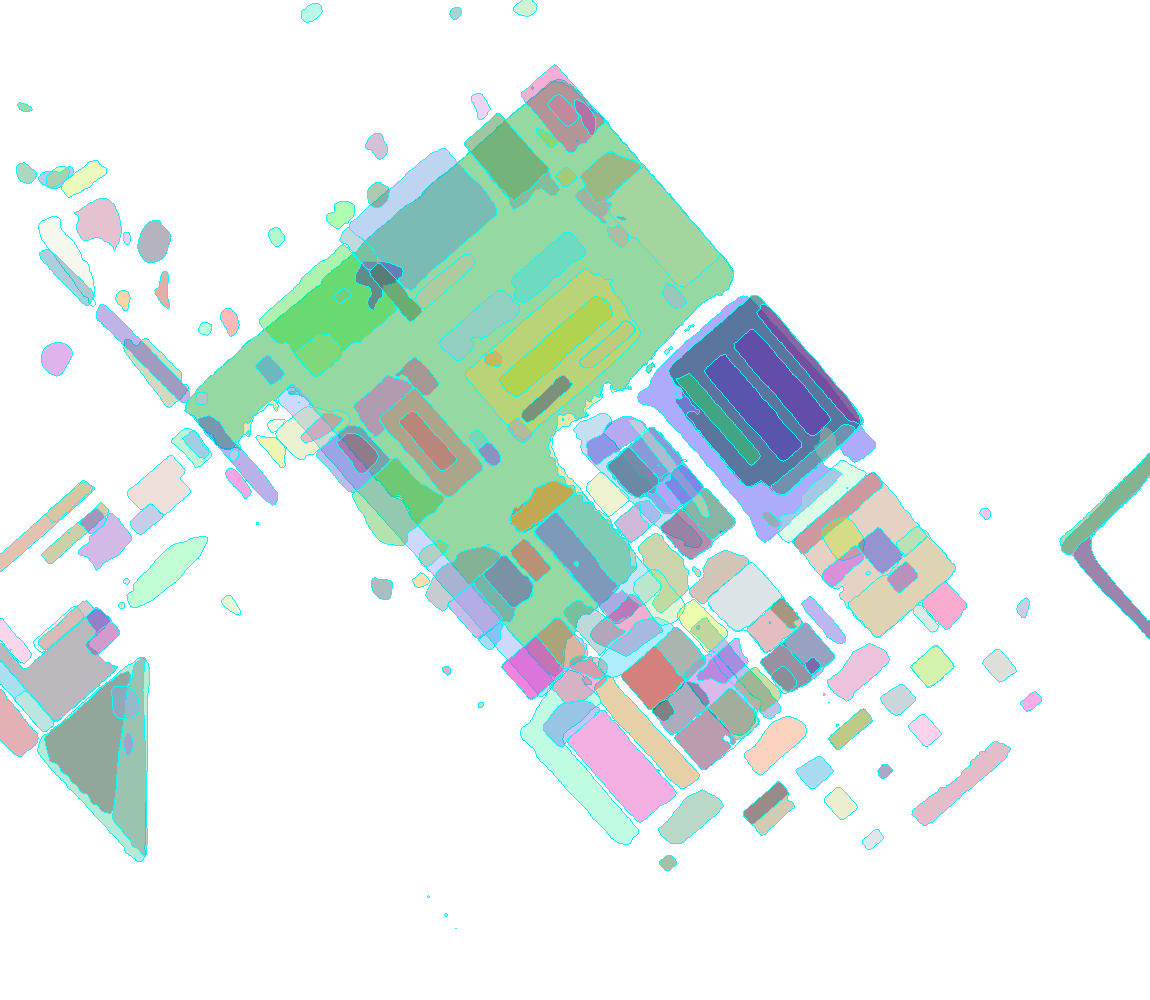} &
\includegraphics[width=0.22\textwidth]{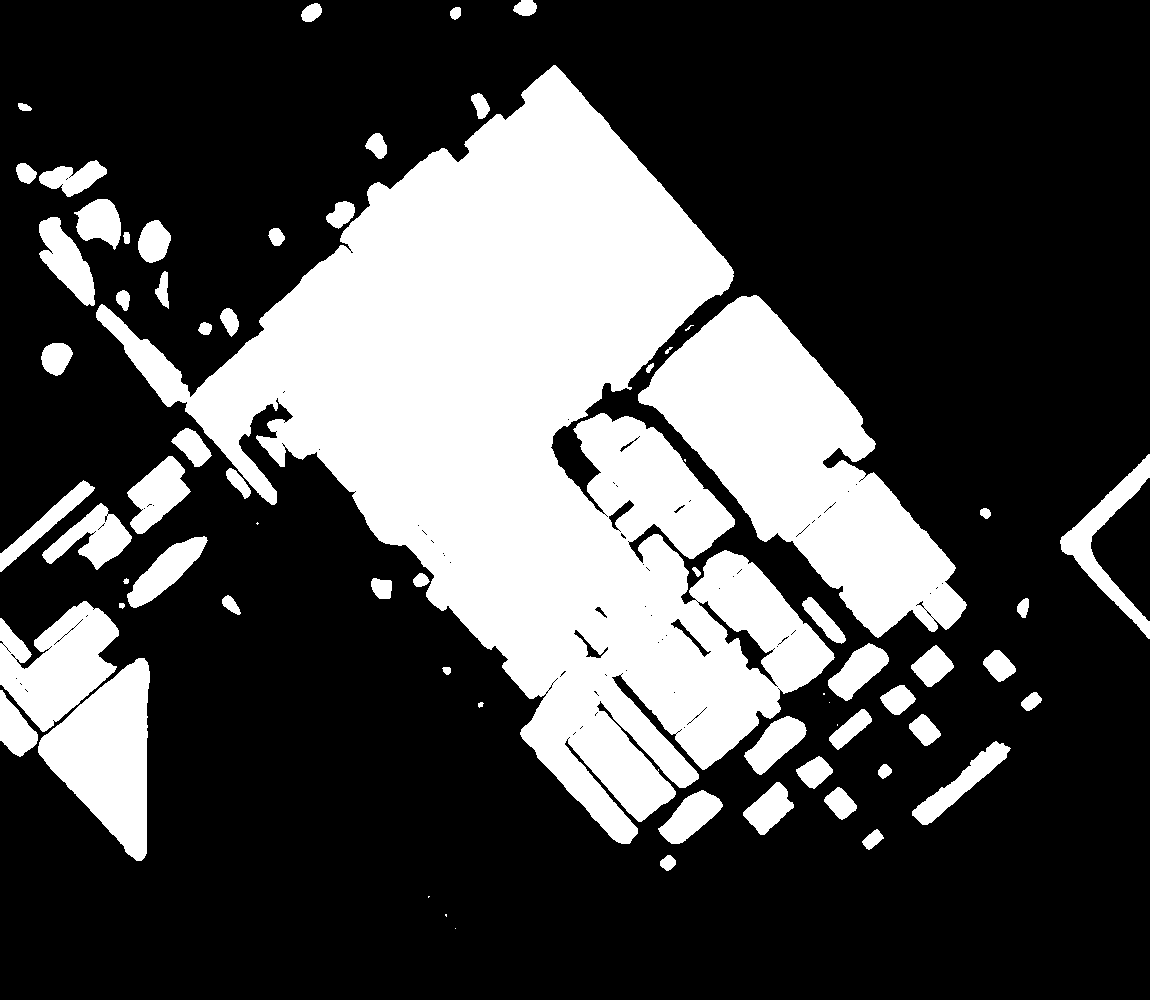} \\
\includegraphics[width=0.22\textwidth]{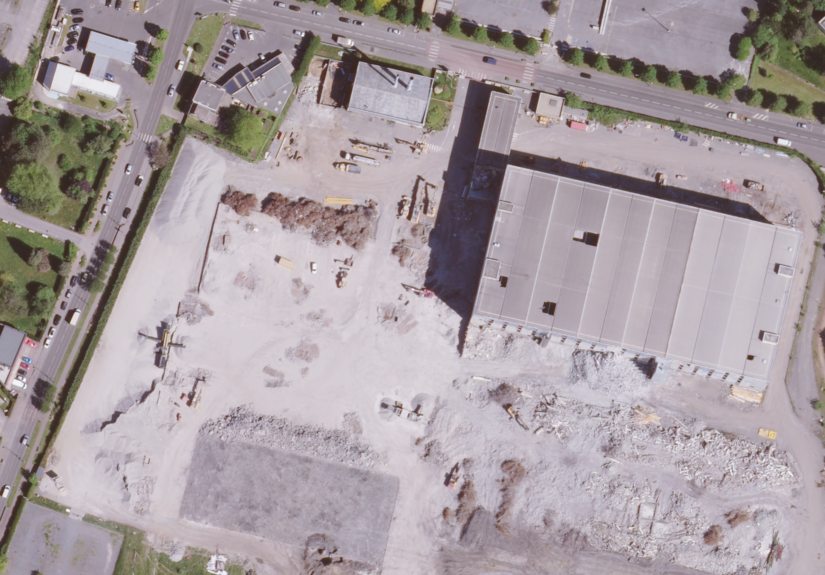} &
\includegraphics[width=0.22\textwidth]{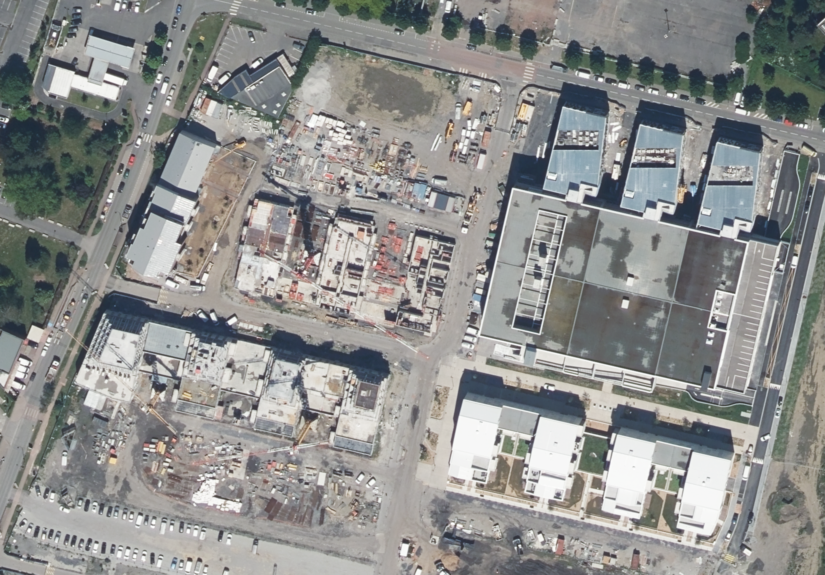} &
\includegraphics[width=0.22\textwidth]{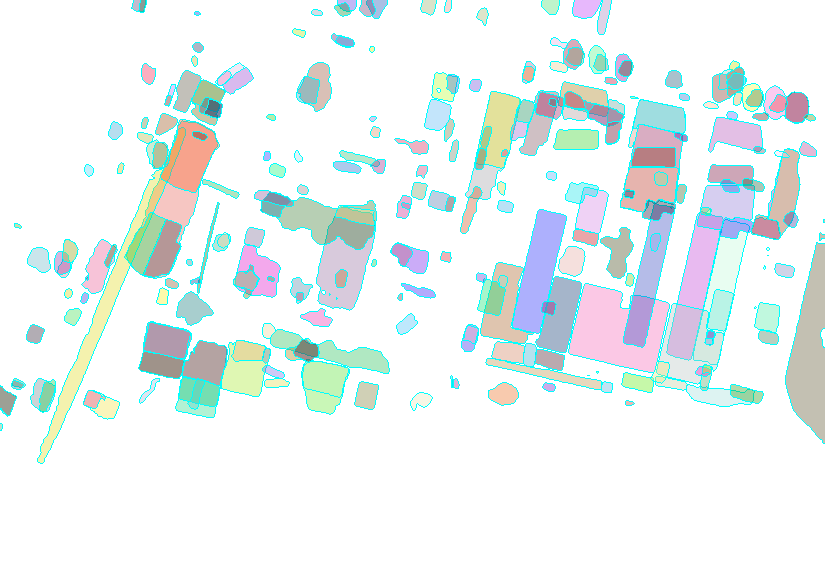} &
\includegraphics[width=0.22\textwidth]{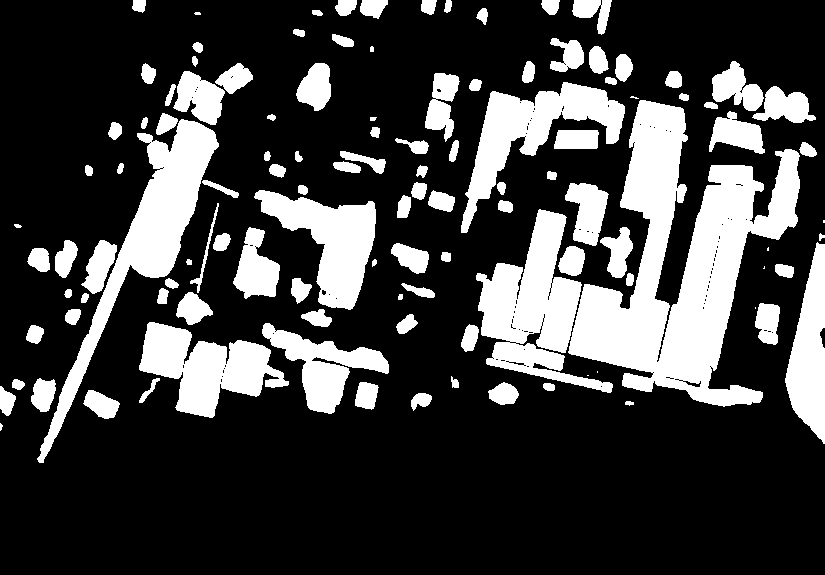} \\
\end{tabular}
}
\vspace{3pt}

\resizebox{\linewidth}{!}{
\begin{tabular}{ccccc}
Input t1 & Input t2 & Prediction & Binary Change Map & Ground Truth \\

\includegraphics[width=0.19\textwidth]{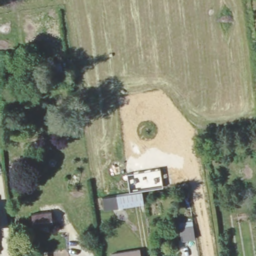} &
\includegraphics[width=0.19\textwidth]{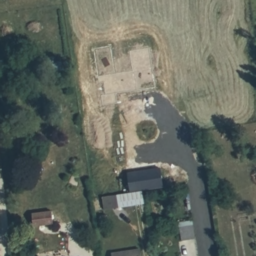} &
\includegraphics[width=0.19\textwidth]{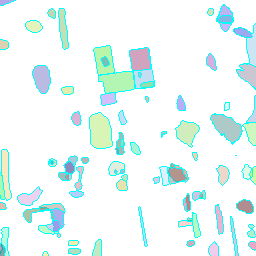} &
\includegraphics[width=0.19\textwidth]{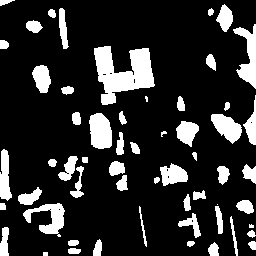} &
\includegraphics[width=0.19\textwidth]{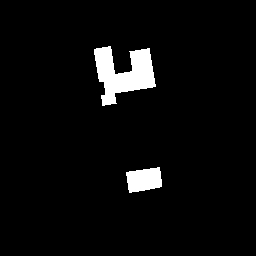} \\

\includegraphics[width=0.19\textwidth]{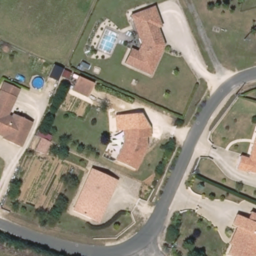} &
\includegraphics[width=0.19\textwidth]{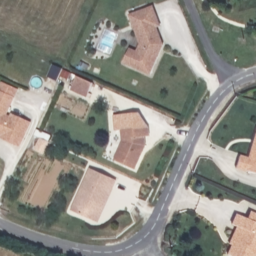} &
\includegraphics[width=0.19\textwidth]{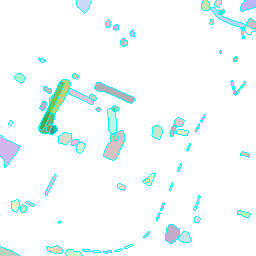} &
\includegraphics[width=0.19\textwidth]{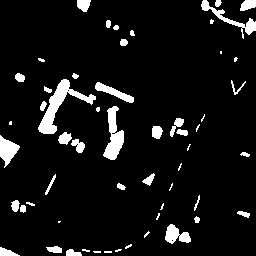} &
\includegraphics[width=0.19\textwidth]{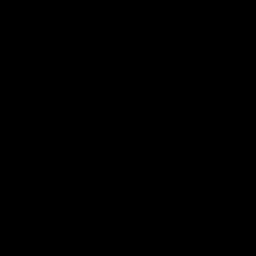} \\

\includegraphics[width=0.19\textwidth]{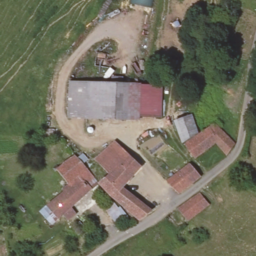} &
\includegraphics[width=0.19\textwidth]{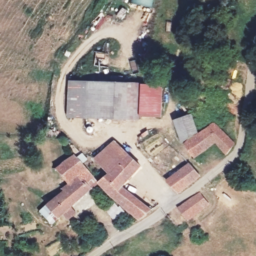} &
\includegraphics[width=0.19\textwidth]{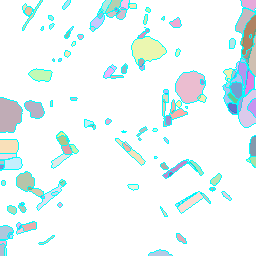} &
\includegraphics[width=0.19\textwidth]{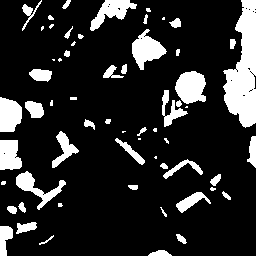} &
\includegraphics[width=0.19\textwidth]{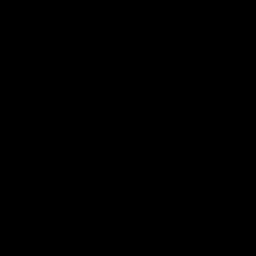} \\

\end{tabular}
}
\caption{\textbf{Qualitative examples of the Segment Any Change method with no prompt provided to indicate building semantics}. As shown, class-agnostic changes are predicted all over images, yielding a considerable number of false positives. Top three rows include images from the main text Figure 1, and bottom examples correspond to FLAIR-BCD samples.}
\label{fig:supp_segment_any_change_no_prompt}
\end{figure*}

Figure~\ref{fig:supp_segment_any_change_no_prompt} illustrates class-agnostic predictions produced by Segment Any Change on several examples. As observed, many detected instances do not correspond to buildings, leading to binary change maps with a large number of false positives. For examples from the FLAIR-BCD dataset, the discrepancy between the predicted changes and the ground-truth building masks illustrates the prevalence of these false detections.

\begin{figure*}[t]
\centering
\scriptsize
\setlength{\tabcolsep}{2pt}

\resizebox{\linewidth}{!}{
\begin{tabular}{cccc}
Input t1 & Input t2 & Prediction & Binary Change Map \\
\includegraphics[width=0.22\textwidth]{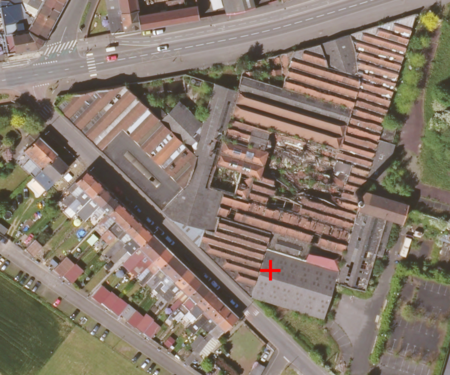} &
        \includegraphics[width=0.22\textwidth]{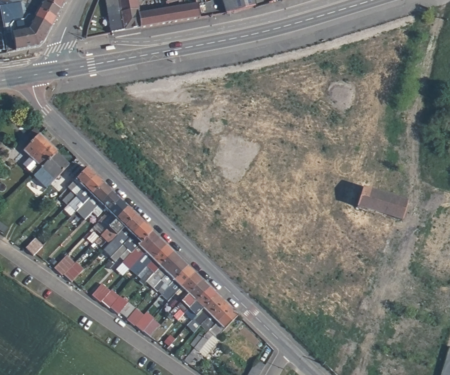} &
        \includegraphics[width=0.22\textwidth]{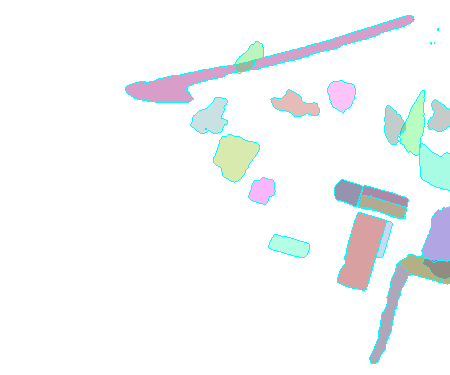} &
        \includegraphics[width=0.22\textwidth]{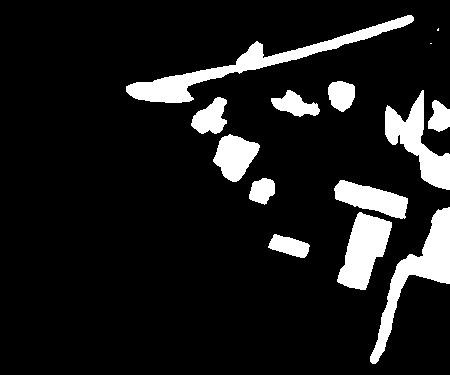} \\
        \includegraphics[width=0.22\textwidth]{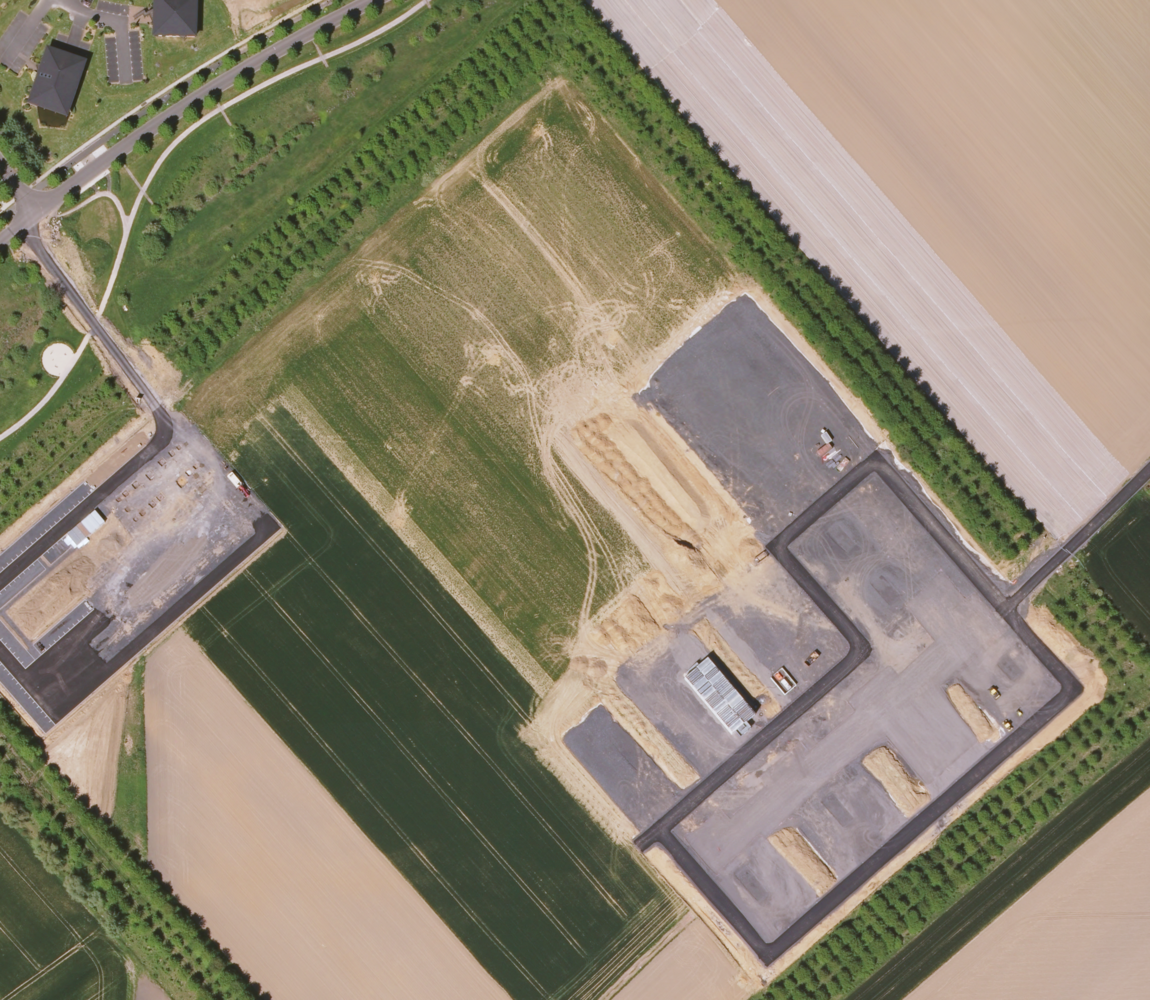} &
        \includegraphics[width=0.22\textwidth]{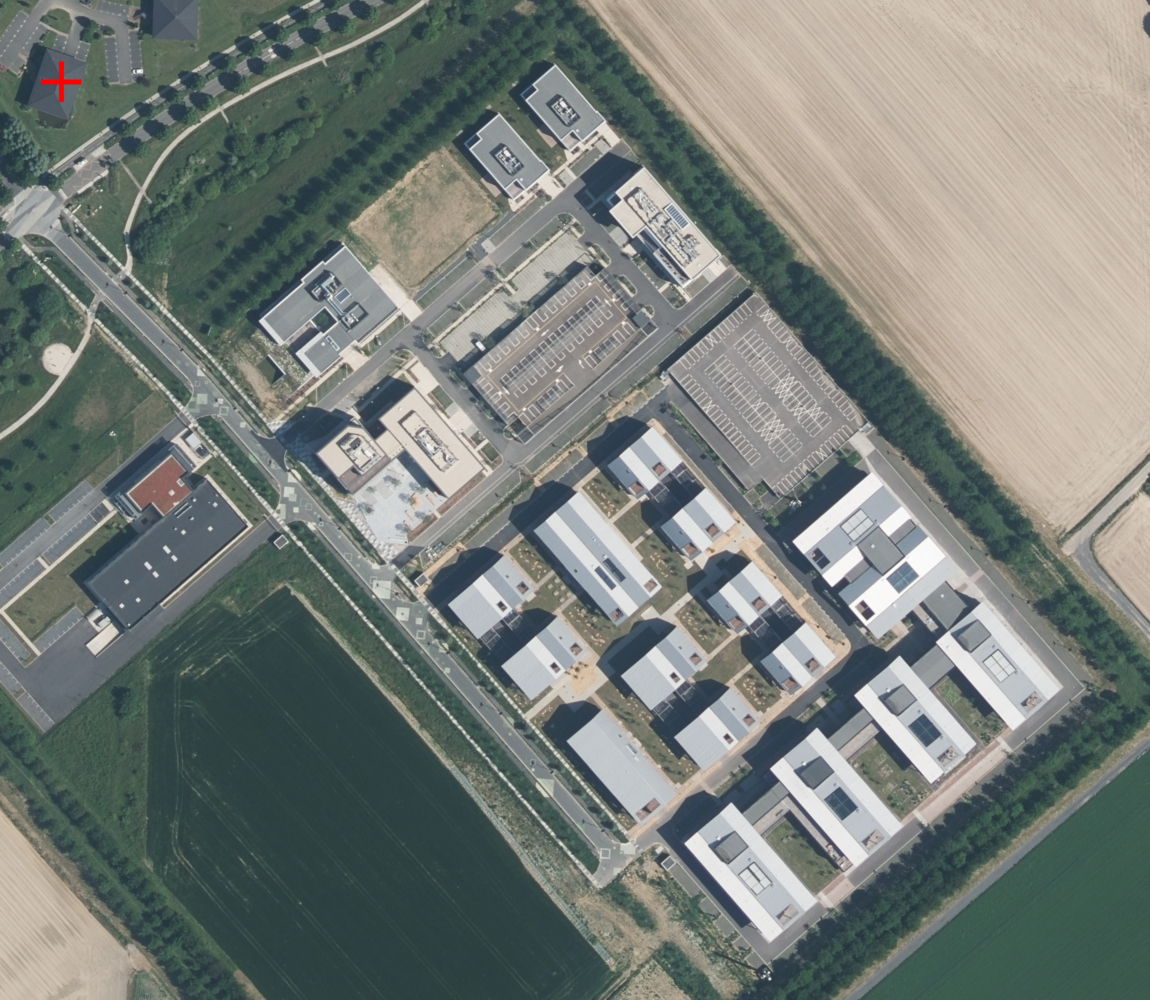} &
        \includegraphics[width=0.22\textwidth]{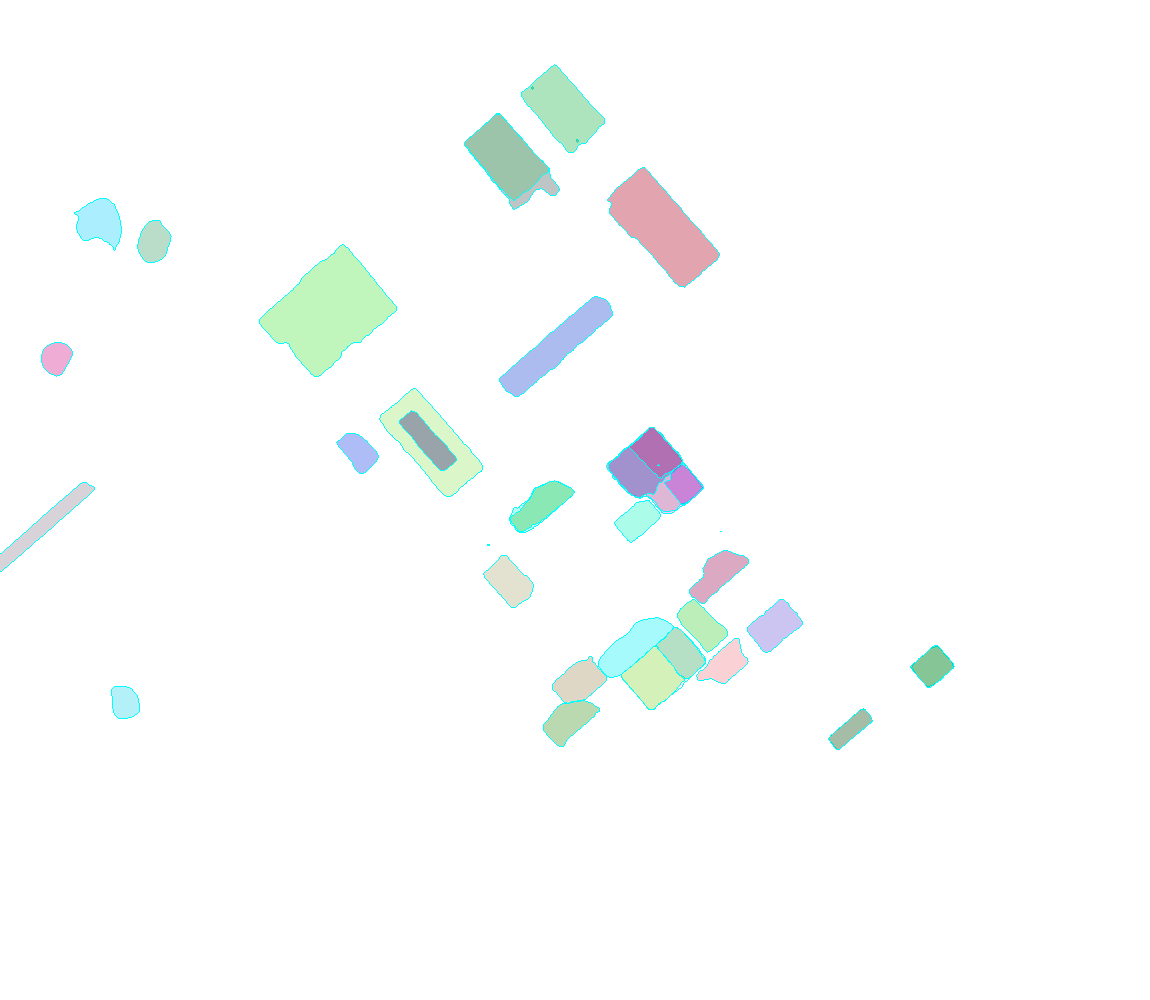} &
        \includegraphics[width=0.22\textwidth]{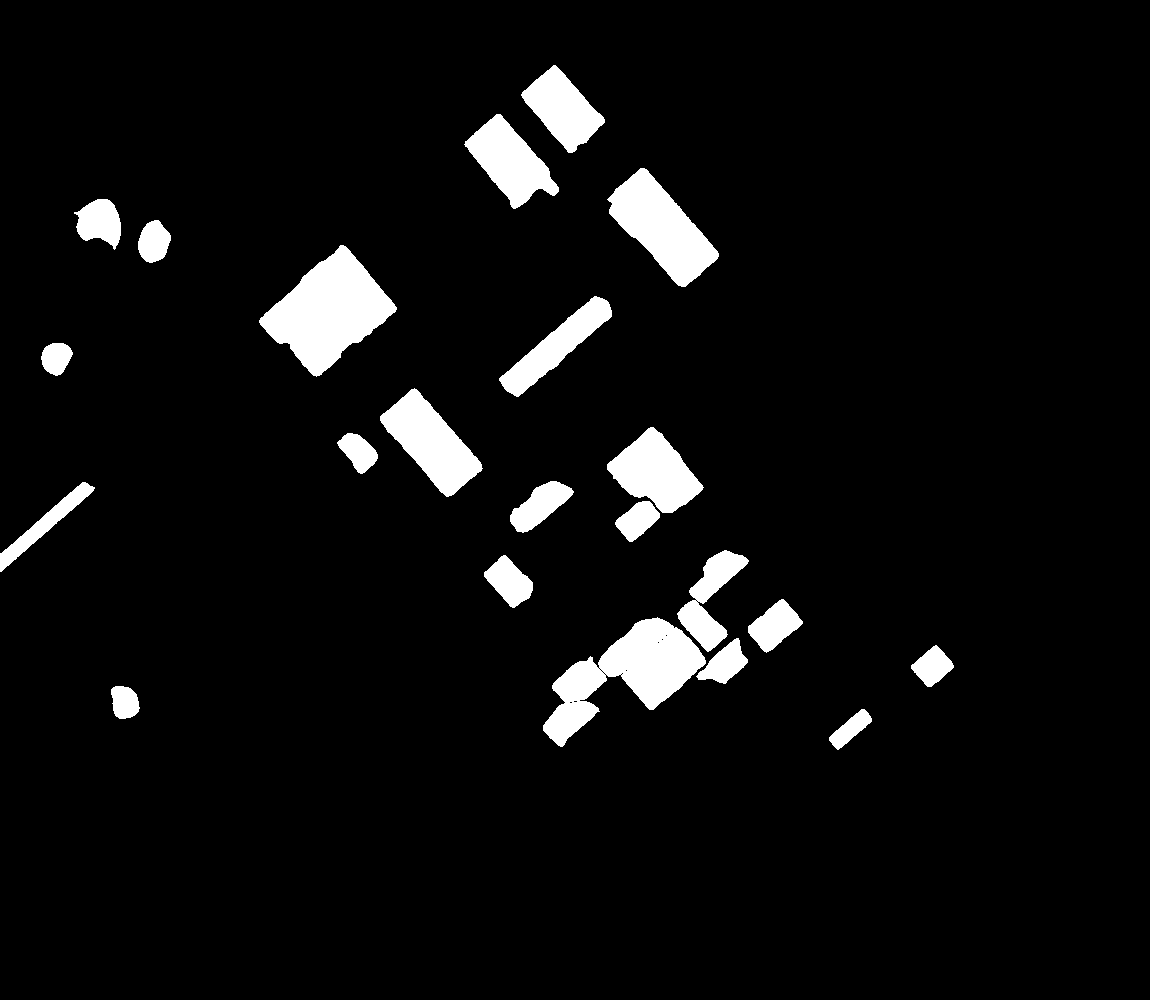} \\
        \includegraphics[width=0.22\textwidth]{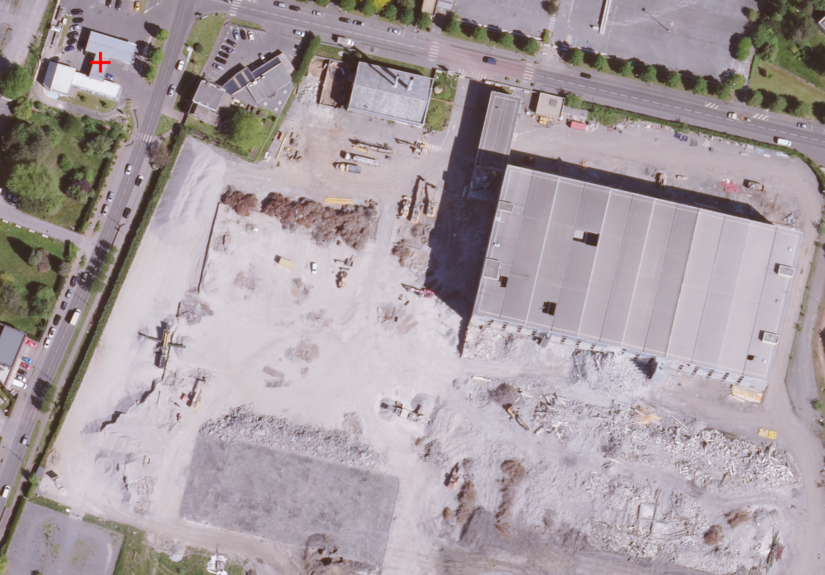} &
        \includegraphics[width=0.22\textwidth]{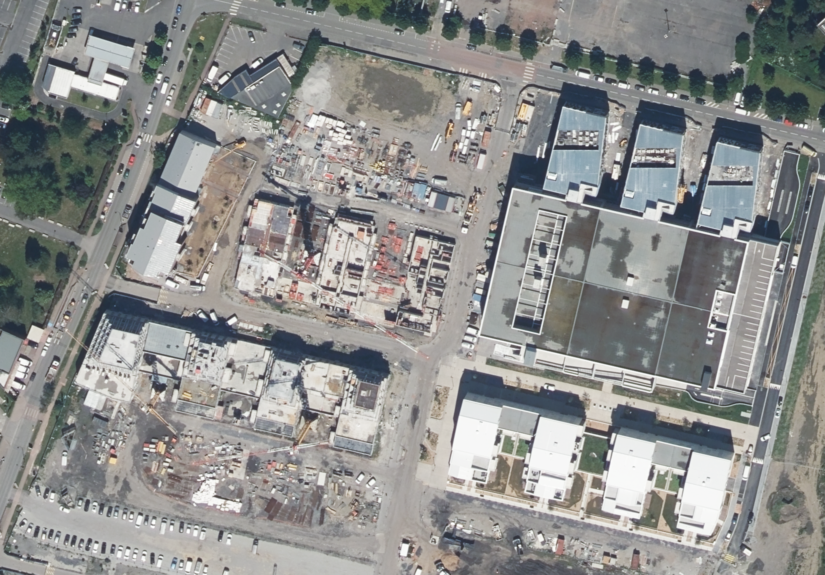} &
        \includegraphics[width=0.22\textwidth]{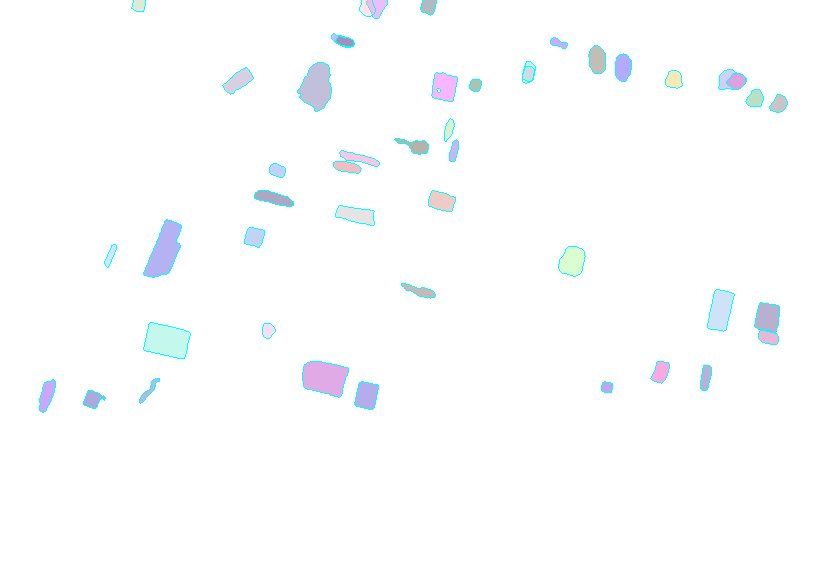} &
        \includegraphics[width=0.22\textwidth]{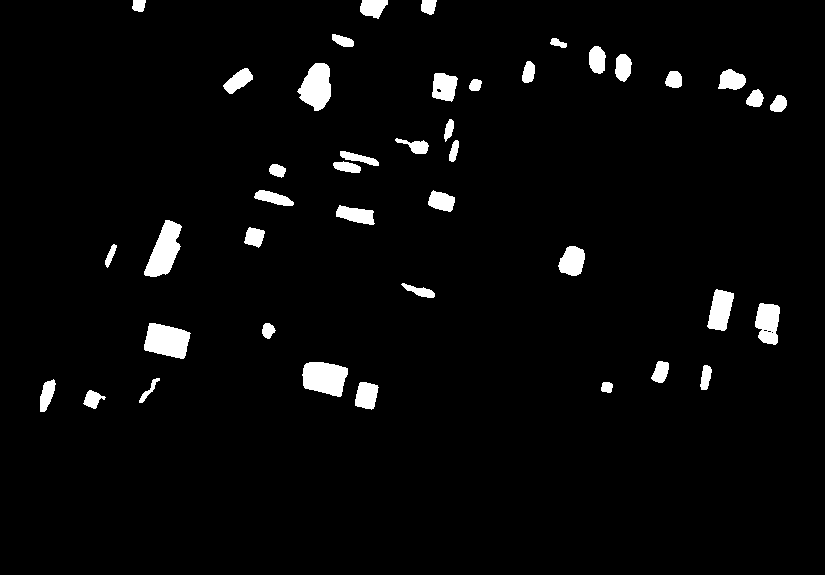} \\
        \end{tabular}
}
\vspace{3pt}

\resizebox{\linewidth}{!}{
\begin{tabular}{ccccc}
Input t1 & Input t2 & Prediction & Binary Change Map & Ground Truth \\

\includegraphics[width=0.19\textwidth]{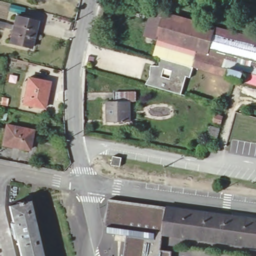} &
\includegraphics[width=0.19\textwidth]{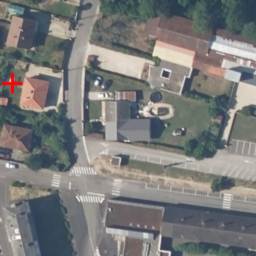} &
\includegraphics[width=0.19\textwidth]{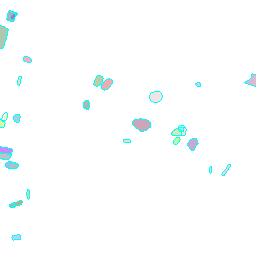} &
\includegraphics[width=0.19\textwidth]{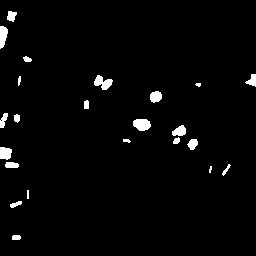} &
\includegraphics[width=0.19\textwidth]{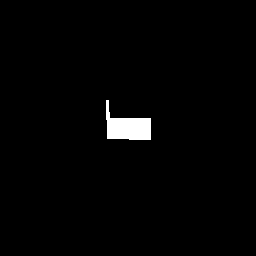} \\

\includegraphics[width=0.19\textwidth]{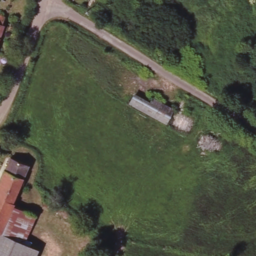} &
\includegraphics[width=0.19\textwidth]{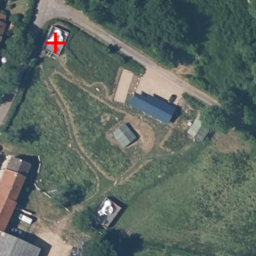} &
\includegraphics[width=0.19\textwidth]{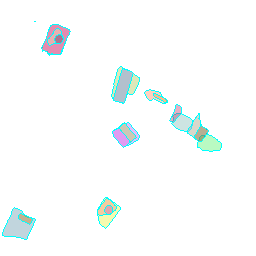} &
\includegraphics[width=0.19\textwidth]{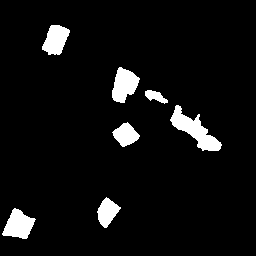} &
\includegraphics[width=0.19\textwidth]{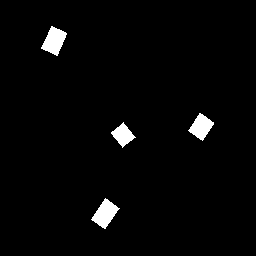} \\

\includegraphics[width=0.19\textwidth]{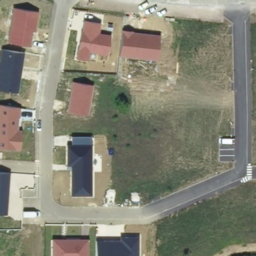} &
\includegraphics[width=0.19\textwidth]{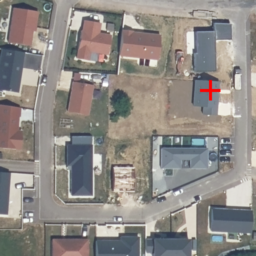} &
\includegraphics[width=0.19\textwidth]{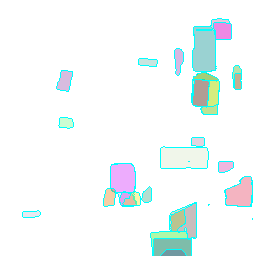} &
\includegraphics[width=0.19\textwidth]{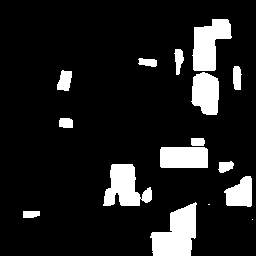} &
\includegraphics[width=0.19\textwidth]{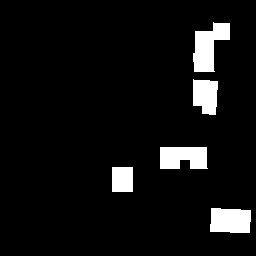} \\

\end{tabular}
}
\caption{\textbf{Prompt-guided qualitative examples of the Segment Any Change method.} A prompt, i.e. a localization point in one of the two input images, is given to indicate building semantics. As shown, prompt guiding does not provide competitive results. Top three rows include images from the main text Figure 1, and bottom examples correspond to FLAIR-BCD samples.}
    \label{fig:supp_segment_any_change_with_prompt}
\end{figure*}

\paragraph{Prompt-guided Segment Any Change for class-specific changes.}
The authors of Segment Any Change suggest that prompts can be used to guide the model toward semantically meaningful changes. In practice, this prompt consists of a pixel location provided in one of the input images. We therefore evaluate this setting on building change examples by selecting a pixel located on a building to guide the detection process. Figure~\ref{fig:supp_segment_any_change_with_prompt} presents the resulting predictions, where the prompt location is indicated by a red cross.

Despite the prompt guidance, the model fails to detect several building changes while still producing numerous false positives (see the first row of Figure~\ref{fig:supp_segment_any_change_with_prompt}). Moreover, although this approach enables some level of semantic control, it is impractical for real-world applications. Providing a prompt requires prior knowledge of building locations in one of the images, which is infeasible for large-scale change detection. Additionally, this strategy cannot characterize the semantics of scenes where no buildings are present (for building change detection).

\subsection{Qualitative results on Hi-UCD}
\begin{figure*}[t]
\centering
\setlength{\tabcolsep}{1pt}

\begin{tabular}{cccccccc}
$t_1$ & $t_2$ & CD-GT & Ours & STAR & PC+TA \\

\includegraphics[width=0.12\linewidth]{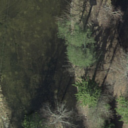} &
\includegraphics[width=0.12\linewidth]{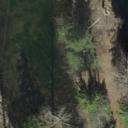} &
\includegraphics[width=0.12\linewidth]{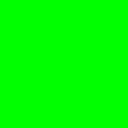} &
\includegraphics[width=0.12\linewidth]{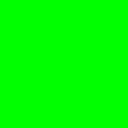} &
\includegraphics[width=0.12\linewidth]{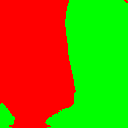} &
\includegraphics[width=0.12\linewidth]{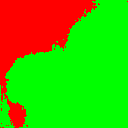} \\

\includegraphics[width=0.12\linewidth]{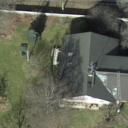} &
\includegraphics[width=0.12\linewidth]{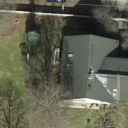} &
\includegraphics[width=0.12\linewidth]{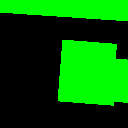} &
\includegraphics[width=0.12\linewidth]{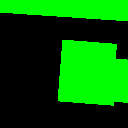} &
\includegraphics[width=0.12\linewidth]{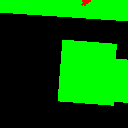} &
\includegraphics[width=0.12\linewidth]{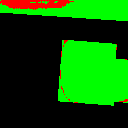} \\

\includegraphics[width=0.12\linewidth]{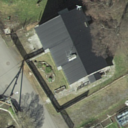} &
\includegraphics[width=0.12\linewidth]{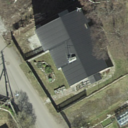} &
\includegraphics[width=0.12\linewidth]{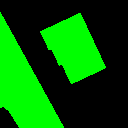} &
\includegraphics[width=0.12\linewidth]{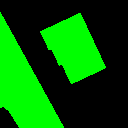} &
\includegraphics[width=0.12\linewidth]{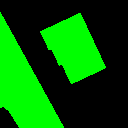} &
\includegraphics[width=0.12\linewidth]{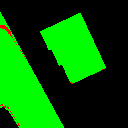} \\

\includegraphics[width=0.12\linewidth]{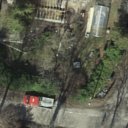} &
\includegraphics[width=0.12\linewidth]{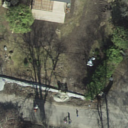} &
\includegraphics[width=0.12\linewidth]{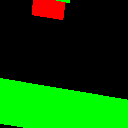} &
\includegraphics[width=0.12\linewidth]{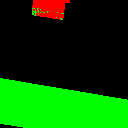} &
\includegraphics[width=0.12\linewidth]{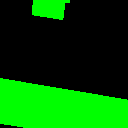} &
\includegraphics[width=0.12\linewidth]{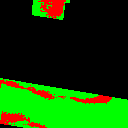} \\

\end{tabular}

\caption{\textbf{Qualitative Hi-UCD results on binary change detection.} We compare the change detection ground truth (CD-GT) to our approach, STAR and post-classification method with temporal augmentation (PC+TA). Green pixels indicate no change, red pixels indicate change, and black pixels are unlabeled. Complementary to quantitative results, our method seems to outperform baselines on binary change detection.}
\label{fig:qualitative_results_hiucd_bcd}
\end{figure*}

\begin{figure*}[t]
\centering
\setlength{\tabcolsep}{1pt}

\begin{tabular}{cccccccccc}
$t_1$ & GT$_{t1}$ & Ours$_{t1}$ & STAR$_{t1}$ & PC+TA$_{t1}$ &
$t_2$ & GT$_{t2}$ & Ours$_{t2}$ & STAR$_{t2}$ & PC+TA$_{t2}$ \\

\includegraphics[width=0.095\linewidth]{images/hiucd_qualitative/input_t1/sample_4_7_t1.png} &
\includegraphics[width=0.095\linewidth]{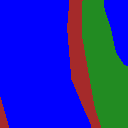} &
\includegraphics[width=0.095\linewidth]{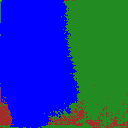} &
\includegraphics[width=0.095\linewidth]{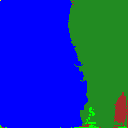} &
\includegraphics[width=0.095\linewidth]{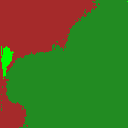} &
\includegraphics[width=0.095\linewidth]{images/hiucd_qualitative/input_t2/sample_4_7_t2.png} &
\includegraphics[width=0.095\linewidth]{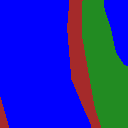} &
\includegraphics[width=0.095\linewidth]{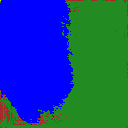} &
\includegraphics[width=0.095\linewidth]{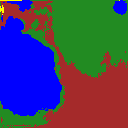} &
\includegraphics[width=0.095\linewidth]{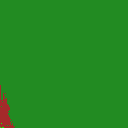} \\

\includegraphics[width=0.095\linewidth]{images/hiucd_qualitative/input_t1/sample_2_43_t1.png} &
\includegraphics[width=0.095\linewidth]{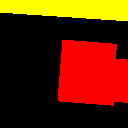} &
\includegraphics[width=0.095\linewidth]{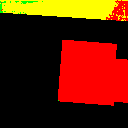} &
\includegraphics[width=0.095\linewidth]{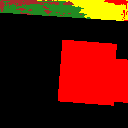} &
\includegraphics[width=0.095\linewidth]{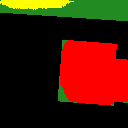} &
\includegraphics[width=0.095\linewidth]{images/hiucd_qualitative/input_t2/sample_2_43_t2.png} &
\includegraphics[width=0.095\linewidth]{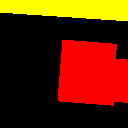} &
\includegraphics[width=0.095\linewidth]{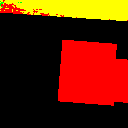} &
\includegraphics[width=0.095\linewidth]{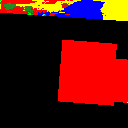} &
\includegraphics[width=0.095\linewidth]{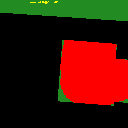} \\

\includegraphics[width=0.095\linewidth]{images/hiucd_qualitative/input_t1/sample_2_58_t1.png} &
\includegraphics[width=0.095\linewidth]{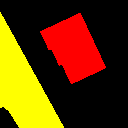} &
\includegraphics[width=0.095\linewidth]{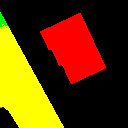} &
\includegraphics[width=0.095\linewidth]{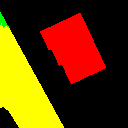} &
\includegraphics[width=0.095\linewidth]{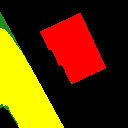} &
\includegraphics[width=0.095\linewidth]{images/hiucd_qualitative/input_t2/sample_2_58_t2.png} &
\includegraphics[width=0.095\linewidth]{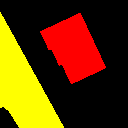} &
\includegraphics[width=0.095\linewidth]{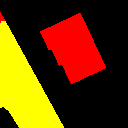} &
\includegraphics[width=0.095\linewidth]{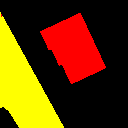} &
\includegraphics[width=0.095\linewidth]{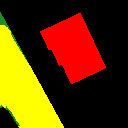} \\

\includegraphics[width=0.095\linewidth]{images/hiucd_qualitative/input_t1/sample_3_54_t1.png} &
\includegraphics[width=0.095\linewidth]{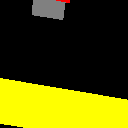} &
\includegraphics[width=0.095\linewidth]{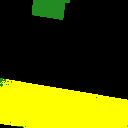} &
\includegraphics[width=0.095\linewidth]{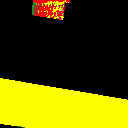} &
\includegraphics[width=0.095\linewidth]{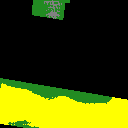} &
\includegraphics[width=0.095\linewidth]{images/hiucd_qualitative/input_t2/sample_3_54_t2.png} &
\includegraphics[width=0.095\linewidth]{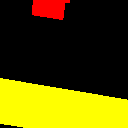} &
\includegraphics[width=0.095\linewidth]{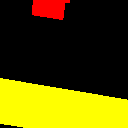} &
\includegraphics[width=0.095\linewidth]{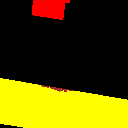} &
\includegraphics[width=0.095\linewidth]{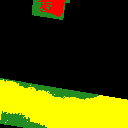} \\

\end{tabular}

\caption{\textbf{Qualitative Hi-UCD results on semantic segmentation.} We show segmentation predictions for timestamps $t_1$ and $t_2$. Our approach is compared with STAR and the post-classification method with temporal augmentation (PC+TA). Our method proved slightly below STAR on semantic segmentation.}
\label{fig:qualitative_results_hiucd_seg}
\end{figure*}

We provide examples of qualitative results in the multi-class setting for the models trained on the Hi-UCD dataset. To this end, Figure~\ref{fig:qualitative_results_hiucd_bcd} compares the predictions of our approach, STAR and post-classification with temporal augmentation for the binary change detection task. Moreover, Figure~\ref{fig:qualitative_results_hiucd_seg} shows the results on the semantic segmentation task for both input images.

\end{document}